\pdfoutput=1
\documentclass[letterpaper]{article} %
\usepackage[preprint]{aaai2027}  %
\usepackage[hyphens]{url}  %
\usepackage{graphicx} %
\usepackage{natbib}  %
\usepackage{caption} %
\usepackage{booktabs}

\usepackage{amsmath,bm}
\usepackage{amsfonts,amssymb}
\usepackage{mathrsfs}
\usepackage{pifont}

\usepackage{colortbl}
\usepackage{multirow}
\usepackage{arydshln}
\usepackage{tabularx}
\usepackage{threeparttable}
\usepackage{makecell}
\usepackage{rotating}

\usepackage{ifthen}
\usepackage{verbatim}
\usepackage{xcolor}
\usepackage{xspace}
\usepackage[most]{tcolorbox}
\tcbuselibrary{listings, breakable}
\usepackage{afterpage}
\usepackage{cuted}
\usepackage{capt-of}

\usepackage[capitalise]{cleveref}
\crefname{figure}{Fig.}{Figs.}
\Crefname{figure}{Figure}{Figures}
\crefname{table}{Tab.}{Tabs.}
\Crefname{table}{Table}{Tables}
\crefname{section}{Sec.}{Secs.}
\Crefname{section}{Section}{Sections}
\crefname{subsection}{Sec.}{Secs.}
\Crefname{subsection}{Section}{Sections}
\crefname{equation}{Eq.}{Eqs.}
\Crefname{equation}{Equation}{Equations}
\crefname{appendix}{App.}{Apps.}
\Crefname{appendix}{Appendix}{Appendices}

\makeatletter
\DeclareRobustCommand\onedot{\futurelet\@let@token\@onedot}
\def\@onedot{\ifx\@let@token.\else.\null\fi\xspace}
\makeatother
\newcommand{\eg}{\emph{e.g}\onedot}

\newcommand{\themethod}{\textsc{CalibAnyView}\xspace}

\definecolor{tabfirst}{rgb}{1, 0.7, 0.7}
\definecolor{tabsecond}{rgb}{1, 0.85, 0.7}
\newcommand{\cfirst}{\cellcolor{tabfirst}\bfseries}
\newcommand{\csecond}{\cellcolor{tabsecond}}
\newcommand{\cthird}{}

\newcommand{\beq}{\begin{equation}}
\newcommand{\eeq}{\end{equation}}

\newcommand{\0}{\phantom{0}}
\newcommand{\degree}{\ensuremath{^\circ}}
\newcommand{\lossone}[1]{intra-}
\newcommand{\losstwo}[1]{inter-}

\usepackage{listings}
\definecolor{lightgray}{gray}{0.95}
\newtcblisting{PromptBox}{
  colback=lightgray,
  colframe=black!30,
  listing only,
  listing options={
    basicstyle=\ttfamily\footnotesize,
    breaklines=true,
    columns=fullflexible,
    keepspaces=true,
    aboveskip=0pt,
    belowskip=0pt,
    lineskip=-1pt,
  },
  left=3pt,
  right=3pt,
  top=3pt,
  bottom=3pt,
  before=\setlength{\parskip}{0pt},
  before skip=6pt,
}

\newcommand{\suppref}[1]{\cref{#1}}

\title{CalibAnyView: Beyond Single-View Camera Calibration in the Wild}

\author{
    Boying Li\textsuperscript{\rm 1},
    Cheng Zhang\textsuperscript{\rm 1},
    Weirong Chen\textsuperscript{\rm 2},
    Guyuan Chen\textsuperscript{\rm 3},\\
    Daniel Cremers\textsuperscript{\rm 2},
    Jianfei Cai\textsuperscript{\rm 1},
    Ian Reid\textsuperscript{\rm 4},
    Hamid Rezatofighi\textsuperscript{\rm 1}
}
\affiliations{
    \textsuperscript{\rm 1}Monash University\\
    \textsuperscript{\rm 2}Technical University of Munich\\
    \textsuperscript{\rm 3}Zhejiang University\\
    \textsuperscript{\rm 4}Mohamed bin Zayed University of Artificial Intelligence (MBZUAI)
}

\begin{document}

\maketitle

\begin{strip}
    \centering
    \includegraphics[width=0.97\textwidth]{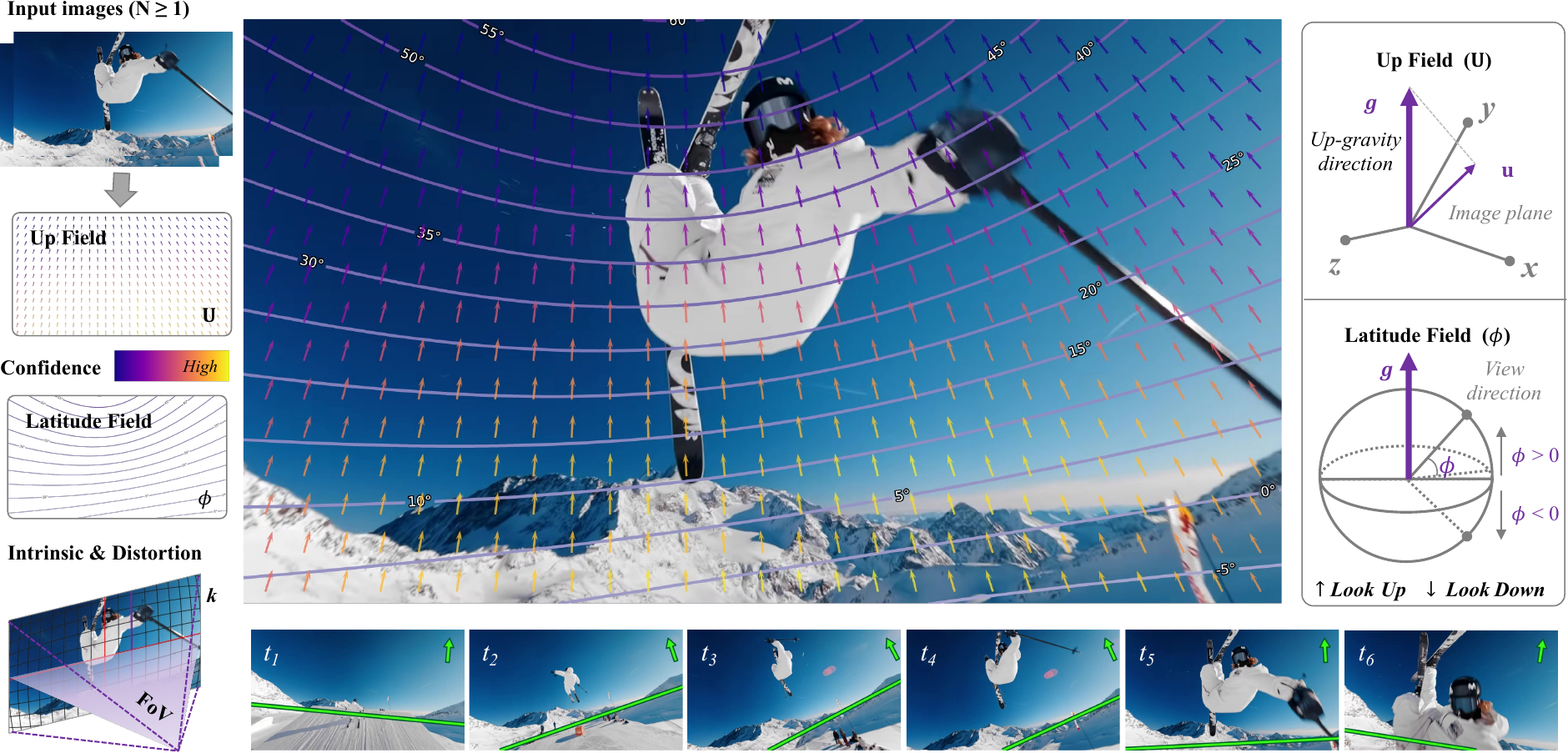}
    \captionof{figure}{
        \textbf{CalibAnyView calibrates cameras from image(s) in the wild.}
        Given a single image or a sparse multi-view sequence, our framework predicts the \textbf{Up Field} ($\mathbf{U}$) and \textbf{Latitude Field} ($\mathbf{\Phi}$), from which \textbf{camera intrinsics}, \textbf{distortion parameters}, and \textbf{gravity direction} are recovered by geometric optimization.
        The bottom row shows a sequential image clip with predicted gravity highlighted by green arrows and lines, while the middle panel presents a zoomed-in view with the gravity predictions.
    }\label{fig:teaser}
\end{strip}

\begin{abstract}

Camera calibration is fundamental to reliable geometric perception, yet classical approaches rely on dedicated targets, successful reconstruction, or dense view coverage, which casually captured imagery rarely satisfies.
Recent learning-based single-image methods lift these requirements by exploiting visual cues, and extend calibration beyond intrinsics to gravity estimation.
Yet in the common multi-view case, they predict each view independently and combine the estimates only afterwards, lacking full use of cross-view consistency.
We bridge this gap with CalibAnyView, a framework that unifies single- and sparse multi-view calibration by enforcing that consistency inside the network: a transformer with cross-view attention predicts camera-model-agnostic perspective fields, followed by a multi-view optimization that fuses them into shared intrinsics and per-view gravity directions, covering camera models from pinhole to severely distorted lenses.
To support this, we construct a large-scale in-the-wild multi-view video dataset spanning diverse camera models, dynamic scenes, realistic motion trajectories, and heterogeneous lens distortions.
Extensive experiments show that CalibAnyView outperforms state-of-the-art methods in both sparse multi-view and single-view settings spanning pinhole to distorted optics.

\end{abstract}

\section{Introduction}
\label{sec:intro}

Camera calibration is a fundamental prerequisite for reliable geometric perception, supporting applications such as 3D reconstruction~\citep{schoenberger2016sfm,agarwal2011building}, scene understanding~\citep{armeni2017joint}, and robotic navigation~\citep{shah2023lm}.
Classical methods recover the intrinsic parameters that define the projection from 3D rays to image pixels, including focal length, principal point, and lens distortion, by establishing precise 3D-to-2D correspondences from dedicated targets~\citep{zhang2000calibration,zhang1999flexible,lochman2021babelcalib}, through self-calibration within Structure-from-Motion (SfM)~\citep{schoenberger2016sfm,agarwal2011building,civera2009camera} and SLAM~\citep{engel2017direct,Hagemann2023droidcalib,carrera2011slam} pipelines, or from environmental cues such as lines and vanishing points~\citep{kovsecka2002video,coughlan1999manhattan,Pautrat2023uvp}.
Each of these presupposes some combination of dedicated targets, structured scenes, successful reconstruction, and dense view coverage, and degrades sharply once its prerequisites are violated~\citep{veicht2024geocalib}, the common case for casual imagery from smartphones, drones, and mobile robots.

Learning-based single-image methods lift these requirements by exploiting visual cues, and perform robustly in unconstrained environments~\citep{bogdan2018deepcalib,lopez2019deepcalib,hold2018perceptual,jin2023perspective,veicht2024geocalib,tirado2025anycalib}.
They also extend calibration beyond intrinsics to the \emph{gravity direction}, the absolute orientation of the camera with respect to the vertical axis of the world, by reading it off learned scene priors such as gravity-aligned surface normals, upright objects, and horizon lines~\citep{xian2019uprightnet,lee2021ctrl,jin2023perspective,veicht2024geocalib}.
This orientation is not determined by image correspondences: SfM and SLAM fix cameras only up to an arbitrary reference frame. Recovering it traditionally required an inertial sensor~\citep{qin2018vins} or strong structural assumptions such as a Manhattan world~\citep{kovsecka2002video,coughlan1999manhattan}.
Its value is increasingly evident downstream, where gravity-aware conditioning benefits controllable image~\citep{bernal2025precisecam} and video generation~\citep{fortier2025gimbaldiffusion,zhang2026unified}, and a shared vertical axis reduces the rotation relating two views to a single degree of freedom, improving incremental 3D reconstruction~\citep{kani2026g3t}.
Calibration from one image is nonetheless the harder regime: where reliable cues are scarce, different intrinsics produce nearly identical projections that an isolated view cannot always disambiguate.

In practice, casual captures rarely consist of an isolated image: handheld video clips, burst captures, and robot camera streams provide multiple views of the same physical camera, carrying complementary evidence about the scene geometry.
Existing single-image methods never exchange this evidence across views; they run a predictor on each view independently and only afterwards combine the per-image estimates through a shared-intrinsic optimization~\citep{veicht2024geocalib}.
Feed-forward 3D models do reason jointly across views: G3T~\citep{kani2026g3t}, the work closest to ours, builds on VGGT~\citep{wang2025vggt} to predict upright pointmaps and camera-to-gravity rotations for a whole sequence at once.
Like VGGT, however, it assumes a pinhole camera and therefore cannot account for the lens distortion that pervades casually captured imagery.

\begin{figure}[t]
    \centering
    \includegraphics[width=0.9\linewidth]{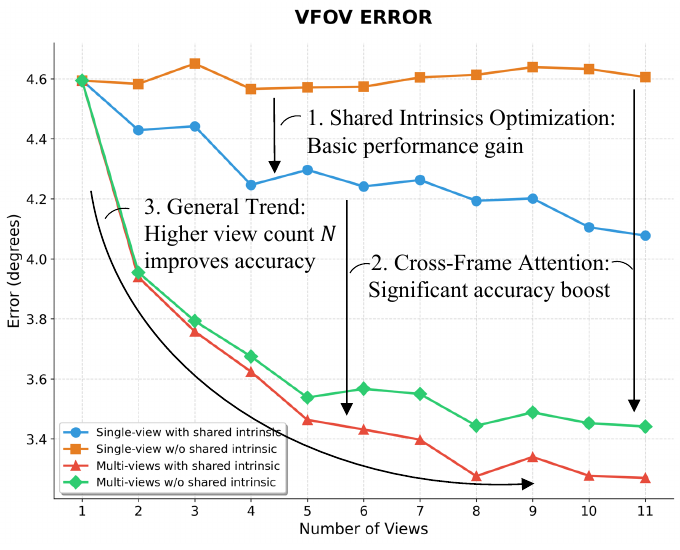}
    \caption{
        \textbf{Calibration error versus the number of input views.}
        vFoV error on the test split of our dataset, with each batch fed to the network jointly (Multi-views) or view by view (Single-view), and the solver either sharing the intrinsics across the batch or not.
        Aggregating views inside the network yields a larger gain than sharing intrinsics in the solver alone, as done in GeoCalib~\citep{veicht2024geocalib}.
    }\label{fig:vfov}
\end{figure}

This motivates our central question: instead of aggregating independent single-image predictions, can cross-view interaction be modeled \emph{within} the calibration model itself and accommodate different camera models?
We answer with \themethod (\cref{fig:teaser}), an \emph{any-view} calibrator that accepts any number of views from a single image to a sparse multi-view sequence ($N \geq 1$): a transformer with alternating intra-view and cross-view attention aggregates evidence across perspectives, and a compact dense-prediction head decodes it into camera-model-agnostic perspective fields.
A multi-view geometric optimization layer then fuses these fields into shared camera intrinsics and per-view gravity directions, covering camera models from pinhole to severely distorted lenses.
Cross-view consistency is thus enforced in both the learned representation and the geometric solver rather than deferred to post-processing: accuracy improves as views are added (\cref{fig:vfov}), while with a single input the same model reduces to robust single-image inference.
Training such a model requires realistic multi-view data that existing resources do not provide, as calibration datasets target single-image supervision~\citep{veicht2024geocalib,tirado2025anycalib,wallingford2024image} while multi-view datasets largely assume pinhole cameras~\citep{wang2020tartanair,liang2025influx}.
We therefore construct a large-scale multi-view video dataset spanning diverse camera models and lens distortions, realistic motion trajectories, dynamic scenes, and varied gravity. %

Our main contributions are summarized as follows:
\begin{itemize}
    \item We introduce \textbf{\themethod}, a unified framework that models cross-view interactions inside the network rather than in post-processing, coupling a cross-view transformer with a multi-view geometric optimization layer that recovers shared intrinsics and per-view gravity across diverse camera models, from a single image to sparse multi-view sequences ($N \geq 1$).
    \item We present a \textbf{large-scale in-the-wild multi-view video dataset} covering multiple camera models, realistic motion trajectories, dynamic scenes, and heterogeneous lens distortions.
    \item Experiments show that our approach \textbf{surpasses state-of-the-art methods} in the sparse multi-view setting and improves as views are added, while the same model remains state of the art on single images and robust in the wild, from pinhole to distorted optics.
\end{itemize}

\section{Related Work}
\label{sec:related}

\paragraph{Single-image geometric calibration.}
Classical approaches exploit projective geometry cues in a single image: parallel lines converge at vanishing points (VPs)~\citep{caprile1990using,cipolla1999camera}, unconstrained line clustering~\citep{Pautrat2023uvp,kluger2020consac}, and Manhattan-world~\citep{kovsecka2002video,coughlan1999manhattan}, while radial distortion can additionally be recovered from curved or covariant line segments~\citep{lochman2021minimal,pritts2020minimal}.
These methods are geometrically precise in structured environments but break down when relevant cues are sparse or absent, and they are restricted to the pinhole~\citep{Pautrat2023uvp,kovsecka2002video,coughlan1999manhattan} or division model~\citep{lochman2021minimal}.

\paragraph{Learning-based single-image calibration.}
Deep learning relaxes precisely these structural requirements by learning where the cues are.
\emph{Regression-based} methods use CNNs or ViTs to directly predict focal length, distortion, or horizon lines~\citep{bogdan2018deepcalib,lopez2019deepcalib,hold2018perceptual}, but introduce no geometric constraints and generalize poorly to unseen camera models~\citep{veicht2024geocalib,tirado2025anycalib}.
\emph{Hybrid methods} instead predict an intermediate geometric representation (\eg, gravity-aligned surface normals or classified line segments~\citep{xian2019uprightnet,lee2021ctrl}), and then recover the parameters analytically or via optimization.
More recent approaches represent the image as per-pixel geometric quantities: Perspective Fields~\citep{jin2023perspective} predict up-vectors and latitude per pixel, GeoCalib~\citep{veicht2024geocalib} refines intrinsics from these maps through a differentiable Levenberg--Marquardt solver, and AnyCalib~\citep{tirado2025anycalib} fits intrinsics in closed form from predicted per-pixel ray directions, covering perspective, fisheye, and edited images.
Single-image calibration nonetheless remains ambiguous wherever reliable cues are scarce, and treating each frame independently discards the geometric consistency present across views.
We address this with cross-view mechanisms that exchange geometric evidence inside a unified model handling both single images and multi-view sequences.

\paragraph{Multi-view calibration.}
Calibration from multiple images predates its single-image counterpart and has traditionally been folded into reconstruction, from target-based bundle adjustment with checkerboards or fiducial markers~\citep{zhang2000calibration} to targetless self-calibration from epipolar geometry~\citep{pollefeys1997stratified}.
Structure-from-Motion~\citep{schoenberger2016sfm} and SLAM~\citep{Hagemann2023droidcalib} pipelines recover poses and intrinsics jointly.
Learning-based variants extend this to self-supervised calibration from video, covering models from pinhole to catadioptric~\citep{gordon2019depth,vasiljevic2020neural,fang2022self}.
Others optimize flexible camera models together with scene geometry during neural reconstruction~\citep{wang2021neural,jeong2021self}, or embed a non-parametric model into SfM~\citep{wang2025structure}.
Across this family, intrinsics come as a by-product of a successful reconstruction and therefore inherit its prerequisites: sufficient parallax, feature overlap, and a reliable initialization.
Motivated by the feed-forward paradigm of recent 3D foundation models~\citep{wang2024dust3r,wang2025vggt},  our multi-view transformer instead calibrates reliably in sparse-view scenarios where reconstruction often fails, recovers the absolute gravity direction that reconstruction-based pipelines leave undetermined, and---unlike the closest concurrent work G3T~\citep{kani2026g3t}---is not confined to a pinhole projection.

\section{Method}
\label{sec:method}
\subsection{Preliminaries}
\label{sec:preliminaries}

\subsubsection{Camera Models and Gravity}
Camera calibration models the projection of a 3D point $\mathbf{P} = [X, Y, Z]^\top$ in the camera frame onto a 2D pixel $\mathbf{p} = [u, v]^\top$ of an $H \times W$ image.
All models we consider factor this map as $\mathbf{p} = f \cdot \mathbf{x} + \mathbf{c}$, where $\mathbf{x}$ is the projection of $\mathbf{P}$ onto the normalized image plane, $f$ is the focal length, which directly determines the Field of View (FoV), and $\mathbf{c}$ is the principal point.
Following common practice for images in the wild~\citep{veicht2024geocalib,jin2023perspective}, we assume a centered image crop and fix $\mathbf{c}$ to half the image dimensions, which reduces the intrinsics to a parameter set $\bm{\lambda} = \{f, d\}$, where $d$ encapsulates the lens distortion: empty for the pinhole model, for which $\mathbf{x} = [X/Z, Y/Z]^\top$; the coefficient $k_1$ for the Radial model~\citep{zhang2000calibration}; and the shape parameters $(\alpha, \beta)$ for the Enhanced Unified Camera Model (EUCM)~\citep{khomutenko2016enhanced}, which projects through an ellipsoidal surface, $\mathbf{x} = [X, Y]^\top / ( \alpha \sqrt{\beta (X^2 + Y^2) + Z^2} + (1 - \alpha) Z )$, and thus covers wide-angle to fisheye optics with a single closed-form model.
The gravity direction is a unit vector $\mathbf{g} = [g_x, g_y, g_z]^\top$ pointing towards the zenith in the camera frame, related to the camera's pitch $\theta$ and roll $\psi$ via $\mathbf{g} = [\sin \psi \cos \theta, -\sin \theta, \cos \psi \cos \theta]^\top$, and thus encapsulating its \textit{absolute orientation}.
We jointly recover $\bm{\lambda}$ and $\mathbf{g}$ across any number of views; the full projection equations of all three models are given in the supplementary material.

\subsubsection{Perspective Representation and Parameter Estimation}
We adopt perspective field~\citep{jin2023perspective, veicht2024geocalib} as the intermediate representation for camera calibration, which provides a dense, camera-model-agnostic geometric grounding by encoding the camera's relationship with the scene's vertical structure at every pixel.
It consists of two components:
\textit{i}) \textbf{Up-vector field} $\mathbf{U} \in \mathbb{R}^{H \times W \times 2}$, whose value $\mathbf{u}_{\mathbf{p}}$ at pixel $\mathbf{p}$ is a unit vector in the image plane pointing towards the projection of the zenith;
\textit{ii}) \textbf{Latitude field} $\Phi \in \mathbb{R}^{H \times W \times 1}$, whose value $\phi_{\mathbf{p}}$ is the angle between the horizon and the viewing ray $\mathbf{v}_{\mathbf{p}}$ back-projected from $\mathbf{p}$:
\begin{equation}
    \mathbf{u}_{\mathbf{p}} = \frac{\mathbf{J}_{\pi} (\mathbf{P}) \mathbf{g}}{\|\mathbf{J}_{\pi} (\mathbf{P}) \mathbf{g}\|}, \quad \phi_{\mathbf{p}} = \arcsin \left( \frac{\mathbf{v}_{\mathbf{p}}^\top \mathbf{g}}{\|\mathbf{v}_{\mathbf{p}}\|} \right),
\end{equation}
where $\mathbf{J}_{\pi}$ is the Jacobian of the projection function at the 3D point $\mathbf{P}$.
The perspective fields $(\mathbf{U}, \Phi)$ are thus determined by the camera intrinsics $\bm{\lambda}$ and the absolute orientation $\mathbf{g}$, so conversely the calibration can be recovered from the fields predicted by a network, denoted $(\hat{\mathbf{U}}, \hat{\Phi})$ with hats marking predicted quantities throughout, by a confidence-weighted non-linear least-squares fit~\citep{veicht2024geocalib}, which our framework extends to the multi-view setting.

\subsection{Unified Any-view Calibration Framework}
\label{sec:framework}

We propose \themethod, a unified transformer-based framework that calibrates a variable number of views ($N \ge 1$).
Specifically, it comprises the following components: alternating attention that propagates calibration cues across frames, a compact DPT head that decodes the latents into perspective fields, and a differentiable solver that enforces shared intrinsics under camera models from pinhole to strongly distorted optics.
Given input images $\mathcal{I} \in \mathbb{R}^{N \times H \times W \times 3}$, the model predicts dense perspective fields $(\mathbf{U}_i, \Phi_i)$ and a per-pixel confidence map $\sigma_i$ for each view $i$, from which the shared intrinsics $\bm{\lambda}$ and the per-view gravity $\mathbf{g}_i$ are jointly recovered via the solver.

\subsubsection{Geometric Latent Extraction.}
We first leverage DINOv2~\citep{oquab2023dinov2} to obtain dense patch-level representations from each input frame, which capture objects of canonical orientation and scene layouts that serve as implicit cues for gravity and focal length.
We adapt alternating attention~\citep{wang2025vggt} to calibration, capturing the image details that gravity and intrinsics estimation requires: \textit{intra-frame self-attention} encodes structural constraints such as gravity-aligned verticals and distorted perspective grids within each view, while \textit{global cross-frame attention} injects multi-view information, establishing correspondences that resolve orientation ambiguities and enforce shared intrinsic consistency across the sequence.

\subsubsection{Compact Camera Head.}
Following the extraction, we propose a compact camera head based on the Dense Prediction Transformer (DPT)~\citep{ranftl2021vision} to decode the latents into dense perspective fields.
We fuse tokens from the later extraction stages, which carry more consistent structural priors than shallower ones, and progressively upsample them into $(\mathbf{U}, \Phi)$ at $1/4$ of the input resolution: this suppresses the pixel-level noise that would destabilize the optimization, while retaining the global consistency on which gravity and intrinsics estimation depends.

\subsubsection{Multi-view Optimization covering Distorted Optics.}
Motivated by GeoCalib~\citep{veicht2024geocalib}, we recover the parameters from the predicted fields $(\hat{\mathbf{U}}_i, \hat{\Phi}_i)$ with a differentiable Levenberg--Marquardt (LM) solver that minimizes the field residual weighted by the predicted confidence $\sigma_i$.
For a sequence of $N$ views, we enforce a \textit{Shared Intrinsics} constraint, treating $\bm{\lambda}$ as global variables shared across the sequence while estimating a unique gravity direction $\mathbf{g}_i$ per view $i$.
However, this solver only covers pinhole and radial models, leaving out the strongly distorted lenses our dataset targets, so we extend it to EUCM.
This is not a matter of adding two variables: the shape parameters $(\alpha, \beta)$ are bounded and enter the projection under a square root, and the focal length trades against the ellipsoid shape with almost no change in the induced fields, so LM from a generic starting point settles into a local minimum whenever the field of view is wide.
We therefore propose a two-stage scheme for EUCM, in which a closed form obtained by linear least squares supplies a reliable initialization that LM then refines.
Concretely, since a radially symmetric projection preserves the pixel azimuth, the latitude along an iso-radius annulus is a sinusoid in that azimuth, so fitting it per annulus recovers $\mathbf{g}$ in closed form and turns every radius into a ray elevation, from which the focal length follows through a Kannala--Brandt proxy~\citep{kannala2006generic} and $(\alpha, \beta)$ through the EUCM identity, alternated since each is linear only once the other is fixed.
Being algebraic rather than geometric, the closed form requires no gravity prior ($\mathbf{g}$ is its output rather than its input), so it applies to every view independently, including the single-view case; the derivation is given in the supplementary material.

\begin{figure*}[t]
    \centering
    \includegraphics[width=0.9\textwidth]{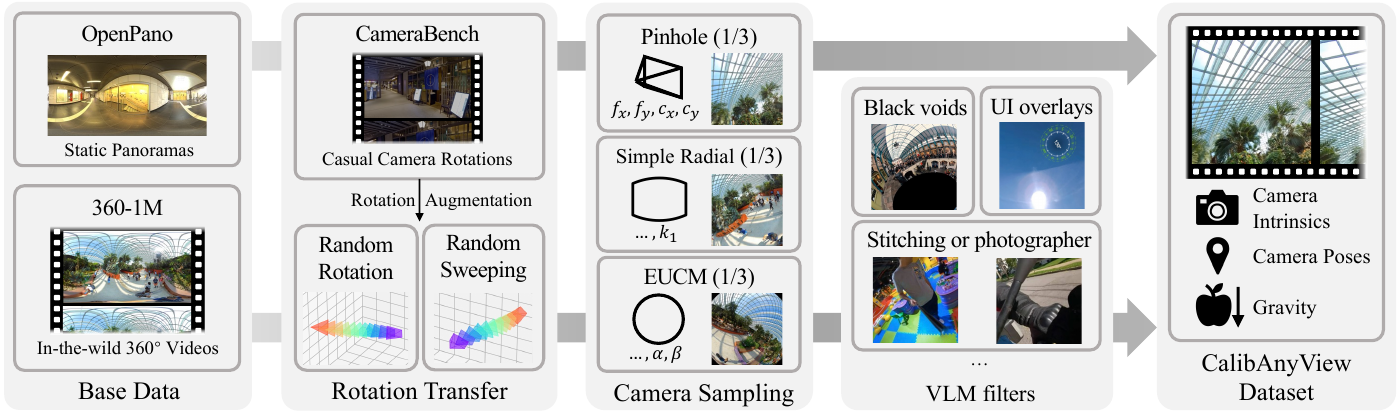}
    \caption{
        \textbf{Overview of the dataset construction pipeline.}
        Panoramic content from static OpenPano and in-the-wild 360-1M videos is projected with transferred camera rotations, each paired with an independently sampled virtual camera; the in-the-wild clips are further filtered by a VLM, yielding video sequences with calibration ground truth.
    }\label{fig:data}
\end{figure*}

\subsubsection{Loss Functions.}
The framework is trained end-to-end with a confidence-weighted dense field loss, which supervises the predicted perspective fields $(\mathbf{U}, \Phi)$ while simultaneously learning a per-pixel confidence map $\sigma$ that quantifies the geometric reliability of each region:
\begin{equation}
    \mathcal{L} = \sum_{i, \mathbf{p}} \left( \gamma \cdot \sigma_{i,\mathbf{p}} \| \mathbf{y}_{i,\mathbf{p}} - \hat{\mathbf{y}}_{i,\mathbf{p}} \|^2 - \eta \cdot \log(\sigma_{i,\mathbf{p}}) \right),
\end{equation}
where $\mathbf{y}_{i,\mathbf{p}}$ concatenates the ground-truth perspective field components $(\mathbf{U}, \Phi)$ for view $i$ at pixel $\mathbf{p}$, $\hat{\mathbf{y}}_{i,\mathbf{p}}$ is the corresponding prediction, and $\gamma, \eta$ are balancing hyperparameters. This likelihood-based weighting~\citep{kendall2017what} lets the network down-weight ambiguous regions such as textureless surfaces and occlusions (e.g., sky in \cref{fig:teaser}).

\subsection{Multi-view Calibration Dataset Construction}
\label{sec:dataset}

Recent advances in camera-aware video generation~\citep{zhang2026unified,fortier2025gimbaldiffusion} and camera pose estimation~\citep{huang2024360loc} highlight the necessity of modeling diverse real-world camera geometry, yet existing calibration datasets are severely limited to single view~\citep{veicht2024geocalib,tirado2025anycalib}, and most multi-view ones~\citep{liang2025influx,wang2020tartanair} lack diverse lens distortions that learning-based calibration requires.
We therefore build our dataset from panoramic content, which can be re-projected into virtual cameras of any model and field of view, and draw the viewing motion from two complementary sources (\cref{fig:data}): 
Static panoramas yield rotation-only sequences without parallax, whereas in-the-wild 360$^\circ$ videos carry the translation and dynamics of real capture.

\paragraph{Rotation-Only Sequences from Static Panoramas}
We place a virtual camera at the center of the static, gravity-aligned panoramas of OpenPano~\citep{veicht2024geocalib} and rotate it along real camera trajectories transferred from CameraBench~\citep{lintowards}, so that the views follow the rotational dynamics of an actual capture, further augmented with random orientation offsets and sweeps into several trajectories per panorama.
Every resulting trajectory is assigned a randomly sampled virtual camera, following the sampling protocol of AnyCalib~\citep{tirado2025anycalib}: Pinhole, Simple Radial, and EUCM in equal proportion, jointly covering vertical fields of view from $20^\circ$ to $180^\circ$; the exact parameter distributions are given in the supplementary material.
This subset totals 29,110 clips.

\paragraph{In-the-Wild Sequences from 360$^\circ$ Videos}
Panoramic videos lift the single-optical-center restriction, yet trajectories generated from a fixed view point~\citep{huang2024360loc} or by synthetic sweeps~\citep{fortier2025gimbaldiffusion} still poorly reflect the dynamics of physical cameras.
We therefore reuse the posed panoramic videos provided by UCPE~\citep{zhang2026unified}, which curates in-the-wild 360$^\circ$ videos from 360-1M~\citep{wallingford2024image,zhang2025panflow} and matches each clip to realistic camera rotations.
On these clips we apply the same augmentation and camera sampling as above, and finally filter the rendered clips with Qwen3-VL~\citep{bai2025qwen3} to discard watermarks, black voids from incomplete panoramas, artificial overlays, and stitching or camera-carrier artifacts, which retains 20,124 clips ($82.6\%$).
Every clip in both subsets spans 5 seconds (81 frames) at 16\,fps and $640 \times 640$ resolution, giving 49,234 multi-view clips and $\sim$4.0M annotated frames in total.

\section{Experiment}
\label{sec:experiment}

\begin{table}[!t]
\centering
\begin{threeparttable}
\small
\renewcommand{\arraystretch}{1.0}%
\setlength\tabcolsep{3.5pt}%
\begin{tabular}{lcccccc}
\toprule
\multirow{2}{*}[-.4em]{Approach}
& \multirow{2}{*}[-.4em]{$N$}
& \multirow{2}{*}{\makecell{AE\,$\downarrow$\\{}[\degree]}}
& \multirow{2}{*}{\makecell{RE\,$\downarrow$\\{}[pix]}}
& \multicolumn{2}{c}{FoV [degrees]}
& \multirow{2}{*}{\makecell{Succ.\\/100}}\\
\cmidrule(lr){5-6}
& & &
& mean\,$\downarrow$ & med.\,$\downarrow$
& \\
\midrule
DroidCalib~\citeyearpar{Hagemann2023droidcalib} & \multirow{5}{*}[-2pt]{8} & 2.89 & $\infty$ & 3.13 & \textbf{0.46} & 91 \\
COLMAP~\citeyearpar{schoenberger2016sfm} &  & 7.48 & 71.0 & 8.62 & 0.53 & 96 \\
GeoCalib$_\text{pin}$~\citeyearpar{veicht2024geocalib} &  & 7.25 & 49.6 & 19.35 & 18.90 & ALL \\
GeoCalib$_\text{dist}$~\citeyearpar{veicht2024geocalib} &  & 3.95 & 31.4 & 9.76 & 8.74 & ALL \\
\textbf{Ours} &  & \textbf{1.00} & \textbf{8.1} & \textbf{2.05} & 1.53 & ALL \\
\midrule
DroidCalib~\citeyearpar{Hagemann2023droidcalib} & \multirow{5}{*}[-2pt]{4} & \multicolumn{4}{c}{\emph{Failed}} & 0 \\
COLMAP~\citeyearpar{schoenberger2016sfm} &  & 22.72 & 356.1 & 30.72 & 16.52 & 80 \\
GeoCalib$_\text{pin}$~\citeyearpar{veicht2024geocalib} &  & 7.38 & 51.2 & 19.68 & 19.38 & ALL \\
GeoCalib$_\text{dist}$~\citeyearpar{veicht2024geocalib} &  & 4.19 & 34.1 & 10.39 & 8.28 & ALL \\
\textbf{Ours} &  & \textbf{1.16} & \textbf{8.9} & \textbf{2.40} & \textbf{1.72} & ALL \\
\midrule
DroidCalib~\citeyearpar{Hagemann2023droidcalib} & \multirow{5}{*}[-2pt]{2} & \multicolumn{4}{c}{\multirow{2}{*}{\emph{Failed}}} & 0 \\
COLMAP~\citeyearpar{schoenberger2016sfm} &  & \multicolumn{4}{c}{} & 0 \\
GeoCalib$_\text{pin}$~\citeyearpar{veicht2024geocalib} &  & 7.73 & 57.0 & 20.56 & 20.53 & ALL \\
GeoCalib$_\text{dist}$~\citeyearpar{veicht2024geocalib} &  & 4.57 & 39.0 & 11.56 & 9.10 & ALL \\
\textbf{Ours} &  & \textbf{1.36} & \textbf{10.2} & \textbf{2.73} & \textbf{2.15} & ALL \\
\bottomrule
\end{tabular}
\caption{
    \textbf{Multi-view calibration on ScanNet++ under a decreasing number of input views.}
    Each sequence keeps only the first $N$ of the 10 frames sampled in \cref{tbl:multiview_public}, with all methods run on the same sequences.
    The reconstruction-based baselines degrade sharply as views are removed, while ours improves steadily with $N$: at $N{=}2$ COLMAP fails on all 100 sequences, since recovering pose, focal length and distortion jointly is under-constrained from a two-view pair, and DroidCalib does not apply below $N{=}8$ by construction, as its SLAM front-end requires a warm-up of eight frames.
    The best of each column within an $N$ block is shown in bold.
}\label{tbl:multiview_views2_scannetpp}
\end{threeparttable}
\end{table}

\begin{table*}[t]
\centering
\begin{threeparttable}
\small
\renewcommand{\arraystretch}{1.05}%
\setlength\tabcolsep{1.4pt}%
\begin{tabular}{clcccccccccccccccccc}
\toprule
&\multirow{2}{*}[-.4em]{Approach}
& \multicolumn{5}{c}{Roll [degrees]}
& \multicolumn{5}{c}{Pitch [degrees]}
& \multirow{2}{*}{\makecell{AE\,$\downarrow$\\{}[\degree]}}
& \multirow{2}{*}{\makecell{RE\,$\downarrow$\\{}[pix]}}
& \multicolumn{5}{c}{FoV [degrees]}
& \multirow{2}{*}{\makecell{Succ.\\/100}}\\
\cmidrule(lr){3-7}
\cmidrule(lr){8-12}
\cmidrule(lr){15-19}
&& mean\,$\downarrow$ & med.\,$\downarrow$ & \multicolumn{3}{c}{AUC\,$\triangleright$\,1/5/10\degree\,$\uparrow$}
& mean\,$\downarrow$ & med.\,$\downarrow$ & \multicolumn{3}{c}{AUC\,$\triangleright$\,1/5/10\degree\,$\uparrow$}
& &
& mean\,$\downarrow$ & med.\,$\downarrow$ & \multicolumn{3}{c}{AUC\,$\triangleright$\,1/5/10\degree\,$\uparrow$}
& \\
\midrule
& DroidCalib~\citeyearpar{Hagemann2023droidcalib} & \multicolumn{10}{c}{\multirow{3}{*}{\emph{Does not estimate gravity}}} & 7.27 & $\infty$ & \cthird 13.81 & 6.49 & \csecond 21.0 & 27.4 & 32.6 & 72 \\
& COLMAP~\citeyearpar{schoenberger2016sfm} & \multicolumn{10}{c}{} & 17.17 & \cthird 263.8 & 26.13 & 6.58 & 14.0 & 23.3 & 31.0 & 75 \\
& VGGT~\citeyearpar{wang2025vggt} & \multicolumn{10}{c}{} & 6.99 & 415.7 & 19.23 & 6.17 & \cthird 17.4 & 32.4 & 43.4 & ALL \\
& G3T~\citeyearpar{kani2026g3t} & \csecond 3.93 & 1.22 & 51.4 & 69.4 & 79.5 & 5.04 & \cthird 2.01 & 35.7 & 55.3 & 69.4 & 7.61 & 438.3 & 20.75 & 9.09 & 14.5 & 24.9 & 34.8 & ALL \\
& GeoCalib$_\text{pin}$~\citeyearpar{veicht2024geocalib} & \cthird 4.66 & \csecond 0.62 & \cthird 73.9 & \cthird 86.7 & \cthird 90.2 & \cthird 4.89 & 2.03 & \cthird 37.2 & \cthird 57.5 & \cthird 71.9 & \cthird 6.01 & $\infty$ & 16.46 & \cthird 4.79 & 16.0 & \cthird 33.6 & \cthird 49.3 & ALL \\
& GeoCalib$_\text{dist}$~\citeyearpar{veicht2024geocalib} & 4.75 & \cthird 0.65 & \csecond 75.2 & \csecond 87.4 & \csecond 90.7 & \csecond 4.39 & \csecond 1.75 & \csecond 39.8 & \csecond 62.3 & \csecond 75.3 & \csecond 4.95 & \csecond 62.3 & \csecond 13.51 & \csecond 3.55 & 15.0 & \csecond 36.7 & \csecond 53.7 & ALL \\
\multirow{-7}{*}{\begin{sideways}\textbf{Stanford2D3D}\end{sideways}} %
& \textbf{Ours} & \cfirst 2.03 & \cfirst 0.60 & \cfirst 77.9 & \cfirst 91.8 & \cfirst 94.8 & \cfirst 2.49 & \cfirst 0.79 & \cfirst 58.4 & \cfirst 83.6 & \cfirst 90.8 & \cfirst 1.16 & \cfirst 11.7 & \cfirst 2.87 & \cfirst 1.83 & \cfirst 26.0 & \cfirst 55.7 & \cfirst 73.2 & ALL \\
\midrule
& DroidCalib~\citeyearpar{Hagemann2023droidcalib} & \multicolumn{10}{c}{\multirow{3}{*}{\emph{Does not estimate gravity}}} & \cthird 4.65 & \cthird 68.1 & \csecond 4.24 & \csecond 1.99 & \csecond 26.0 & \csecond 42.9 & \csecond 53.9 & 77 \\
& COLMAP~\citeyearpar{schoenberger2016sfm}       & \multicolumn{10}{c}{} & 15.84 & 284.8 & 19.60 & \cthird 4.48 & \cthird 19.0 & \cthird 36.6 & \cthird 43.9 & 95 \\
& VGGT~\citeyearpar{wang2025vggt}                            & \multicolumn{10}{c}{} & 10.22 & 281.1 & 26.03 & 26.29 & 0.0 & 0.0 & 0.0 & ALL \\
& G3T~\citeyearpar{kani2026g3t}                              & 0.84 & \cthird 0.56 & \csecond 73.3 & 89.3 & \cthird 94.6 & \csecond 1.40 & 1.27 & \cthird 50.0 & 76.7 & 88.2 & 10.77 & 299.2 & 27.24 & 27.23 & 0.0 & 0.0 & 0.0 & ALL \\
& GeoCalib$_\text{pin}$~\citeyearpar{veicht2024geocalib}                  & \csecond 0.81 & \csecond 0.55 & \cthird 72.9 & \csecond 90.3 & \csecond 95.0 & 1.66 & \csecond 1.04 & 47.0 & \cthird 78.1 & \cthird 88.4 & 5.53 & 79.7 & 13.53 & 12.98 & 0.0 & 0.0 & 4.2 & ALL \\
& GeoCalib$_\text{dist}$~\citeyearpar{veicht2024geocalib}    & \cthird 0.83 & 0.57 & 69.5 & \cthird 89.7 & \cthird 94.6 & \cthird 1.64 & \cthird 1.09 & \csecond 52.2 & \csecond 80.1 & \csecond 89.5 & \csecond 3.70 & \csecond 67.2 & \cthird 8.56 & 7.33 & 1.0 & 10.4 & 29.9 & ALL \\
\multirow{-7}{*}{\begin{sideways}\textbf{Aria-ADT-RGB}\end{sideways}} %
& \textbf{Ours}                                & \cfirst 0.50 & \cfirst 0.43 & \cfirst 89.7 & \cfirst 97.2 & \cfirst 98.6 & \cfirst 0.71 & \cfirst 0.58 & \cfirst 75.3 & \cfirst 92.2 & \cfirst 96.1 & \cfirst 2.34 & \cfirst 44.7 & \cfirst 1.86 & \cfirst 1.30 & \cfirst 36.0 & \cfirst 66.6 & \cfirst 83.1 & ALL \\
\midrule
& DroidCalib~\citeyearpar{Hagemann2023droidcalib} & \multicolumn{10}{c}{\multirow{3}{*}{\emph{Does not estimate gravity}}} & \csecond 4.25 & \csecond 37.7 & \csecond 6.64 & \cfirst 1.31 & \cfirst 23.0 & \csecond 32.2 & \csecond 37.5 & 53 \\
& COLMAP~\citeyearpar{schoenberger2016sfm}       & \multicolumn{10}{c}{} & 19.57 & 162.0 & 25.19 & \cthird 13.78 & \csecond 21.0 & \cthird 28.3 & \cthird 32.6 & 84 \\
& VGGT~\citeyearpar{wang2025vggt}                            & \multicolumn{10}{c}{} & 19.80 & 277.9 & 52.10 & 52.46 & 0.0 & 0.0 & 0.0 & ALL \\
& G3T~\citeyearpar{kani2026g3t}                              & 1.83 & \cthird 1.48 & 40.8 & 70.2 & 83.9 & 2.39 & 1.95 & 30.3 & 59.4 & 77.8 & 18.44 & 246.2 & 48.76 & 48.73 & 0.0 & 0.0 & 0.0 & ALL \\
& GeoCalib$_\text{pin}$~\citeyearpar{veicht2024geocalib}                  & \csecond 0.70 & \cfirst 0.39 & \cfirst 89.5 & \csecond 96.1 & \csecond 97.9 & \cthird 1.88 & \cthird 1.39 & \cthird 37.9 & \cthird 69.7 & \cthird 83.9 & 13.68 & 195.8 & 35.77 & 36.64 & 0.0 & 0.0 & 0.0 & ALL \\
& GeoCalib$_\text{dist}$~\citeyearpar{veicht2024geocalib}    & \cthird 0.71 & \cfirst 0.39 & \csecond 87.4 & \cthird 95.5 & \cthird 97.6 & \csecond 1.41 & \csecond 1.06 & \csecond 53.2 & \csecond 79.9 & \csecond 89.5 & \cthird 9.51 & \cthird 57.3 & \cthird 24.32 & 24.74 & 0.0 & 0.0 & 1.1 & ALL \\
\multirow{-7}{*}{\begin{sideways}\textbf{Aria-ADT-Gray}\end{sideways}} %
& \textbf{Ours}                                & \cfirst 0.53 & \csecond 0.46 & \cthird 86.9 & \cfirst 96.5 & \cfirst 98.2 & \cfirst 0.77 & \cfirst 0.61 & \cfirst 70.0 & \cfirst 90.5 & \cfirst 95.2 & \cfirst 1.63 & \cfirst 7.7 & \cfirst 3.78 & \csecond 3.22 & \cthird 15.0 & \cfirst 38.7 & \cfirst 63.3 & ALL \\
\midrule
& DroidCalib~\citeyearpar{Hagemann2023droidcalib} & \multicolumn{10}{c}{\multirow{3}{*}{\emph{Does not estimate gravity}}} & \csecond 1.84 & $\infty$ & \csecond 2.09 & \csecond 0.39 & \csecond 72.0 & \cfirst 81.3 & \cfirst 85.3 & 96 \\
& COLMAP~\citeyearpar{schoenberger2016sfm}       & \multicolumn{10}{c}{} & 6.84 & 66.6 & \cthird 5.70 & \cfirst 0.34 & \cfirst 73.0 & \csecond 78.5 & \cthird 80.7 & 97 \\
& VGGT~\citeyearpar{wang2025vggt}                            & \multicolumn{10}{c}{} & 10.85 & 153.5 & 29.48 & 29.55 & 0.0 & 0.0 & 0.0 & ALL \\
& G3T~\citeyearpar{kani2026g3t}                              & 1.33 & 1.21 & 51.3 & 78.5 & 89.0 & 2.98 & 2.75 & 22.8 & 50.6 & 71.6 & 11.71 & 169.3 & 31.54 & 31.15 & 0.0 & 0.0 & 0.0 & ALL \\
& GeoCalib$_\text{pin}$~\citeyearpar{veicht2024geocalib}                  & \cthird 0.89 & \cthird 0.54 & \cthird 77.0 & \cthird 90.2 & \cthird 94.5 & \cthird 1.65 & \cthird 1.40 & \cthird 43.2 & \cthird 74.1 & \cthird 86.5 & 7.35 & \cthird 50.3 & 19.60 & 19.57 & 0.0 & 0.1 & 1.0 & ALL \\
& GeoCalib$_\text{dist}$~\citeyearpar{veicht2024geocalib}    & \csecond 0.78 & \cfirst 0.48 & \csecond 81.0 & \csecond 91.5 & \csecond 95.3 & \csecond 1.60 & \csecond 1.33 & \csecond 48.5 & \csecond 75.5 & \csecond 87.0 & \cthird 4.02 & \csecond 31.7 & 9.90 & 9.05 & 1.0 & 9.6 & 24.8 & ALL \\
\multirow{-7}{*}{\begin{sideways}\textbf{ScanNet++}\end{sideways}} %
& \textbf{Ours}                                & \cfirst 0.56 & \csecond 0.52 & \cfirst 83.5 & \cfirst 95.4 & \cfirst 97.7 & \cfirst 1.04 & \cfirst 0.98 & \cfirst 60.3 & \cfirst 84.9 & \cfirst 92.4 & \cfirst 0.99 & \cfirst 7.9 & \cfirst 2.05 & \cthird 1.60 & \cthird 32.0 & \cthird 63.3 & \csecond 80.8 & ALL \\
\bottomrule
\end{tabular}
\caption{
    \textbf{Multi-view calibration on public benchmarks spanning perspective and distorted cameras.}
    All methods are run on the same sequences with $N{=}10$ views each, and the \colorbox{tabfirst}{best} and \colorbox{tabsecond}{second best} entries of each column are highlighted here and in \cref{tbl:generalization_s2d3d}.
    DroidCalib, COLMAP and VGGT recover camera poses only up to an arbitrary world frame, so roll and pitch relative to gravity are undefined for them by construction.
    \emph{Succ./100} counts the sequences for which a method returns a valid calibration and its metrics are averaged over that subset only, so rows with a lower count are not directly comparable; feed-forward methods always return a calibration and are marked \emph{ALL}.
    A cell is shown as $\infty$ when that mean exceeds $10^3$, which happens when the recovered camera is numerically singular on part of the sequences and the average is no longer meaningful.
}\label{tbl:multiview_public}
\end{threeparttable}
\end{table*}

\begin{table*}[!t]
\centering
\begin{threeparttable}
\small
\renewcommand{\arraystretch}{1.0}%
\setlength\tabcolsep{1.5pt}%
\begin{tabular}{clccccccccccccccc}
\toprule
&\multirow{2}{*}[-.4em]{Approach} 
& \multicolumn{5}{c}{Roll [degrees]} 
& \multicolumn{5}{c}{Pitch [degrees]} 
& \multicolumn{5}{c}{FoV [degrees]}\\
\cmidrule(lr){3-7}
\cmidrule(lr){8-12}
\cmidrule(lr){13-17}
&& mean\,$\downarrow$ & med.\,$\downarrow$ & \multicolumn{3}{c}{AUC\,$\triangleright$\,1/5/10\degree\,$\uparrow$}
& mean\,$\downarrow$ & med.\,$\downarrow$ & \multicolumn{3}{c}{AUC\,$\triangleright$\,1/5/10\degree\,$\uparrow$}
& mean\,$\downarrow$ & med.\,$\downarrow$ & \multicolumn{3}{c}{AUC\,$\triangleright$\,1/5/10\degree\,$\uparrow$} \\
\midrule
&DeepCalib~\citeyearpar{lopez2019deepcalib}      & - &          1.59 &          33.8 &          63.9 &   \cthird 79.2 & - &          2.58 &          21.6 &          46.9 &          \cthird 65.7 & - &          6.67 &          \08.1 &           20.6 &          37.6 \\
&Perceptual~\citeyearpar{perceptual}              & - &          2.08 &          26.8 &          53.8 &          70.7 & - &          3.17 &          21.5 &          41.8 &          57.8 & - &          11.35 &          \04.9 &          13.7 &          24.9 \\
&CTRL-C~\citeyearpar{lee2021ctrl}                       & - &          3.04 &          23.2 &          43.0 &          56.9 & - &          3.43 &          18.3 &          38.6 &          53.8 & - &          8.50 &          \07.7 &           18.2 &          31.5 \\
&MSCC~\citeyearpar{Song2024MSCC}                  & - &          3.43 &          13.5 &          36.8 &          57.3 & - &          2.64 &          22.6 &          45.0 &          60.5 & - &   5.81 &   9.6 &    23.8 &   41.6 \\
&ParamNet~\citeyearpar{jin2023perspective}  & \cthird \05.41 &          \02.51 &          20.5 &          48.5 &          68.1 & \cthird \08.23 &          \02.78 &          20.9 &          44.3 &          61.5 & \09.38 &          7.78 &          \07.4 &           18.0 &          33.2 \\
&SVA~\citeyearpar{lochman2021minimal}                            & - &              - &          21.7 &          24.6 &          25.8 & - &              - &          15.4 &          19.9 &          22.4 & - &              - &          \06.2 &           11.5 &          15.2 \\
&UVP~\citeyearpar{Pautrat2023uvp}    & 9.45 &           \cthird \00.86 &          \cthird 52.7 &          \cthird 64.6 &          71.4 & 10.51 &          \cthird \02.43 &          \cthird 35.6 &          \cthird 48.9 &          58.8 & 20.70 &          8.94 &          17.4 &           27.5 &          36.8 \\
& GeoCalib~\citeyearpar{veicht2024geocalib}& \csecond 1.60 & \cfirst \00.39 & \cfirst 83.1 &          \csecond 91.8 &          \csecond 94.8 & \csecond 2.41 & \csecond 0.93 &          \csecond 52.0 &          \csecond 74.7 &          \csecond 84.6 & 4.93 & 3.27 &          17.6 &           39.9 &          59.4 \\
& DUSt3R~\citeyearpar{wang2024dust3r} & \multicolumn{10}{c}{\multirow{4}{*}{\emph{Does not estimate gravity}}} &   8.12 &   5.34 &          10.6 &          26.1 &          43.8 \\
& VGGT~\citeyearpar{wang2025vggt}    & \multicolumn{10}{c}{} &   4.92 &   3.16 & 19.0 & 40.8 & 59.1 \\
& AnyCalib$_{p}$~\citeyearpar{tirado2025anycalib} & \multicolumn{10}{c}{} &   \csecond 4.22 &   \csecond 2.55 &          \csecond 21.4 &          \csecond 46.9 &          \csecond 64.7 \\
& AnyCalib$_{g}$~\citeyearpar{tirado2025anycalib} & \multicolumn{10}{c}{} &   \cthird 4.66 &   \cthird 2.95 &          \cthird 20.8 &          \cthird 43.6 &          \cthird 61.6 \\
\multirow{-13}{*}{\begin{sideways}\textbf{Stanford2D3D}~\citeyearpar{armeni2017joint}\end{sideways}} & \textbf{Ours} & \cfirst 1.07 & \csecond 0.45 & \csecond 82.9 & \cfirst 94.2 & \cfirst 96.7 & \cfirst 1.40 & \cfirst 0.68 & \cfirst 65.0 & \cfirst 84.6 & \cfirst 91.4 & \cfirst 3.64 & \cfirst 2.08 & \cfirst 28.4 & \cfirst 53.3 & \cfirst 69.4 \\
\bottomrule
\end{tabular}%

\caption{
    \textbf{Single-view comparison on Stanford2D3D.}
    Ours attains the lowest mean error on Roll, Pitch and FoV alike, whereas each baseline is competitive on at most one of gravity and field of view.
    }\label{tbl:generalization_s2d3d}
\end{threeparttable}
\end{table*}

\begin{figure}[t]
    \centering
    \includegraphics[width=\linewidth]{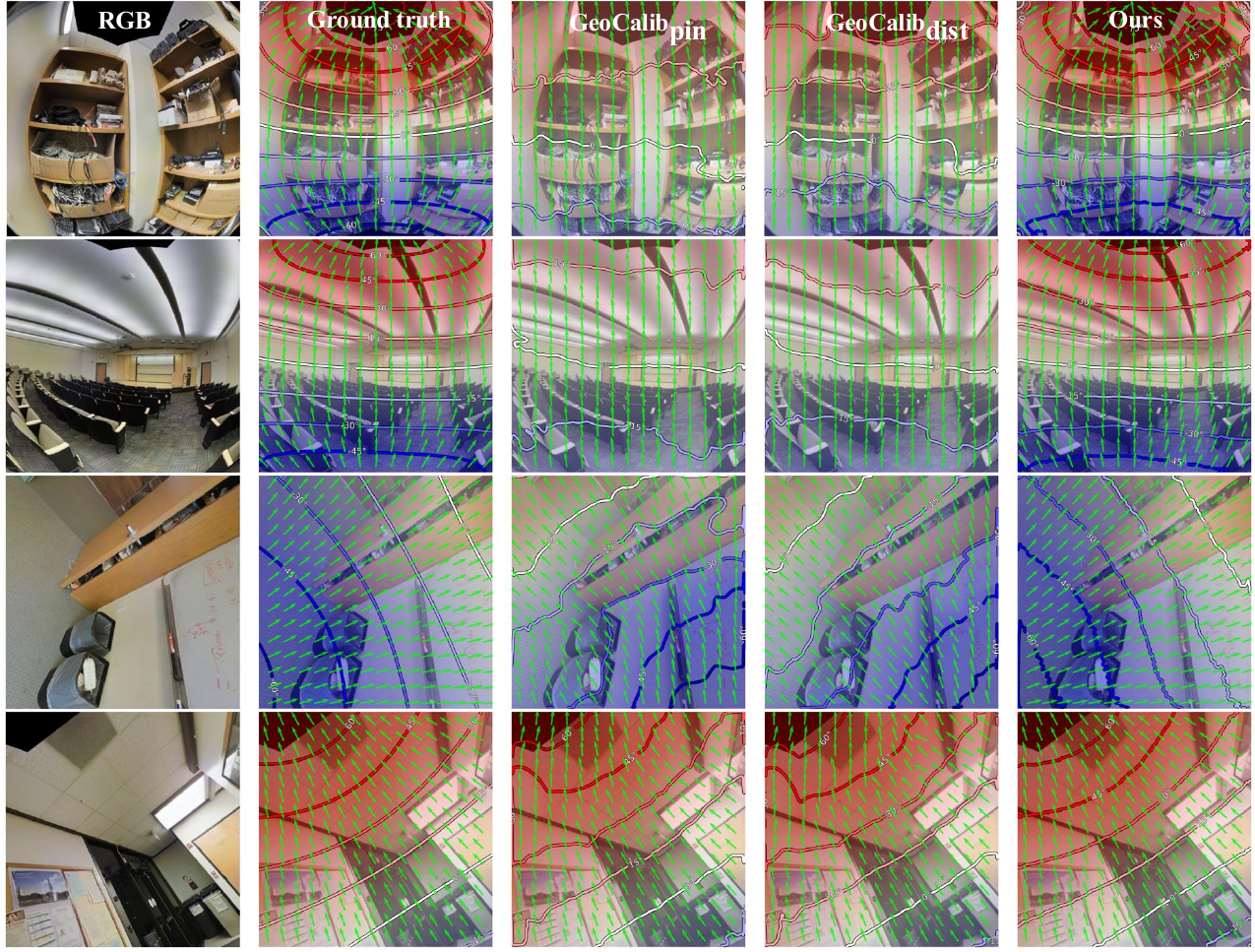}
    \caption{
        \textbf{Qualitative perspective field results on Stanford2D3D.}
        Rows span the benchmark's pinhole, radial, and EUCM cameras.
        As distortion grows, GeoCalib drifts from the ground truth, whereas ours stays faithful across all lens types.
    }\label{fig:multiview_public_qual}
\end{figure}

\subsection{Experimental Setup}
\label{sec:setup}
\subsubsection{Training Details.}
Our model is trained on a mixture of OpenPano and our proposed multi-view dataset (\cref{sec:dataset}) under a hybrid sampling strategy: OpenPano is fed as single images, while sequences of length $N \in [2, 24]$ are drawn from the latter.
We initialize the Geometric Latent Extraction module with pre-trained VGGT~\citep{wang2025vggt} to leverage established geometric priors.
 
\subsubsection{Evaluation Protocols.}
Our model operates on a variable number of views: we first quantify the gains from our geometric aggregation and multi-view data, then fall back to the single-view setting.
We evaluate gravity by the Roll and Pitch errors, intrinsics by the error on the vertical field of view (FoV), all in degrees.
For each quantity we report the median as the typical case, the Area Under the recall Curve (AUC) at $1^\circ$, $5^\circ$, $10^\circ$ as accuracy under increasing tolerances, and the mean, which unlike the other two exposes occasional catastrophic failures.
To further evaluate distortion, we report the mean angular error (AE) and reprojection error (RE) as defined in AnyCalib~\citep{tirado2025anycalib}, which are agnostic to the specific projection model.

\subsection{Multi-view Analysis}
\label{sec:multiview_analysis}

\subsubsection{Comparison on Public Benchmarks.}
Our method attains the lowest mean error on all four multi-view benchmarks in \cref{tbl:multiview_public}.
Notably, Stanford2D3D is the most challenging: it covers all camera models with parameters randomly sampled following AnyCalib, and is rendered from static panoramas with augmented rotations, giving zero parallax.
The other three feature fixed fisheye cameras with diverse trajectories, which favors reconstruction-based methods.
To favor traditional reconstruction methods, we use $N{=}10$ views and run COLMAP and DroidCalib with every camera model they support, reporting the best one (SIMPLE\_DIVISION for COLMAP and Mei for DroidCalib).
Even so, both fail to calibrate part of the sequences when the reconstruction fails, and elsewhere return numerically singular intrinsics that break the RE computation (marked as $\infty$ in the table).
Feed-forward reconstruction models (VGGT and G3T) instead assume a pinhole camera and break down on distorted images.
The dedicated calibration baseline GeoCalib is more robust, yet still falls behind our method.

\subsubsection{Effect of the Number of Views.}
Although traditional methods reach a lower median error once the views suffice for a successful reconstruction ($N  \ge  8$ in \cref{tbl:multiview_views2_scannetpp} and \cref{tbl:multiview_public}), they are not robust to fewer views.
We analyze this by keeping the first $N$ frames of the original 10-view ScanNet++ sequences, so that the view spacing stays unchanged.
Their accuracy degrades dramatically as views are removed and collapses entirely at $N{=}2$, whereas our method improves steadily with $N$ and, in AE and RE, already surpasses at two views what any baseline attains with all ten in \cref{tbl:multiview_public}.

\subsubsection{Ablation Study.}
We further ablate on the test split of our multi-view dataset (\cref{fig:vfov}) to analyze how our multi-view aggregation benefits from the number of views.
A $2 \times 2$ protocol separates the two sources of gain: feeding a batch jointly, activating cross-view attention (multi-views), rather than view by view (single-view), and sharing intrinsics across the batch in the solver (shared intrinsics).
Cross-view attention accounts for the larger share of the improvement and its margin grows with $N$, showing that the gain stems primarily from cross-frame reasoning inside the network.

\subsubsection{Qualitative Results.}
\cref{fig:multiview_public_qual} shows Stanford2D3D samples covering the pinhole, radial and EUCM cameras of the benchmark.
Although both share the same perspective field representation, our predictions stay faithful to the scene as distortion grows, whereas GeoCalib drifts from the ground truth.

\subsection{Single-view Fall Back}
\label{sec:sota_comparison}

We then fall back to the single-view setting ($N{=}1$) and compare against the state of the art on Stanford2D3D~\citep{armeni2017joint}, following GeoCalib's protocol~\citep{veicht2024geocalib}.
As reported in \cref{tbl:generalization_s2d3d}, ours attains the lowest mean error on all three quantities and the best FoV AUC at every threshold, with only its Roll median and AUC$\,\triangleright\,1^\circ$ marginally behind GeoCalib.
AnyCalib, which specializes in intrinsics calibration, stays above our FoV in both its variants, while the 3D foundation models DUSt3R and VGGT recover no gravity at all and remain less accurate in FoV.
Classical geometric estimators (SVA, UVP) and earlier regression networks (DeepCalib, Perceptual, CTRL-C, MSCC~\citep{Song2024MSCC}) trail by a large margin; we quote their numbers from GeoCalib, except Perceptual's FoV, for which we use the value corrected by AnyCalib~\citep{tirado2025anycalib}.
Our aggregator therefore does not trade single-view accuracy for multi-view capability: the same model is state of the art in both regimes.

\section{Conclusion}
\label{sec:conclusion}
We introduced \themethod, a unified framework that generalizes camera calibration to an any-view paradigm, together with a large-scale in-the-wild video dataset coupling realistic camera motion with diverse projection models.
By pairing a cross-view transformer with a geometric optimization layer, we resolve the ambiguities a single view leaves open: accuracy improves as views are added, while remaining state of the art on single images and robust to distortion.

\paragraph{Limitations and Future Work.}
We fix the principal point at the image center and assume shared intrinsics across the sequence, valid for most consumer cameras but not under zoom or asymmetric cropping.
Relaxing either assumption via ray-based parameterizations~\citep{tirado2025anycalib} or view-independent intrinsics remains future work.
Our gains also concentrate in the sparse regime; scaling to long sequences, where attention costs grow quadratically and bundle adjustment stays preferable, is a further challenge.

\bibliography{main}

\begin{thebibliography}{62}
\providecommand{\natexlab}[1]{#1}

\bibitem[{Agarwal et~al.(2011)Agarwal, Furukawa, Snavely, Simon, Curless,
  Seitz, and Szeliski}]{agarwal2011building}
Agarwal, S.; Furukawa, Y.; Snavely, N.; Simon, I.; Curless, B.; Seitz, S.~M.;
  and Szeliski, R. 2011.
\newblock Building rome in a day.
\newblock \emph{Communications of the ACM}, 54(10): 105--112.

\bibitem[{Armeni et~al.(2017)Armeni, Sax, Zamir, and
  Savarese}]{armeni2017joint}
Armeni, I.; Sax, S.; Zamir, A.~R.; and Savarese, S. 2017.
\newblock {Joint 2D-3D-Semantic Data for Indoor Scene Understanding}.
\newblock \emph{arXiv:1702.01105}.

\bibitem[{Bai et~al.(2025)Bai, Cai, Chen, Chen, Chen, Cheng, Deng, Ding, Gao,
  Ge et~al.}]{bai2025qwen3}
Bai, S.; Cai, Y.; Chen, R.; Chen, K.; Chen, X.; Cheng, Z.; Deng, L.; Ding, W.;
  Gao, C.; Ge, C.; et~al. 2025.
\newblock Qwen3-vl technical report.
\newblock \emph{arXiv preprint arXiv:2511.21631}.

\bibitem[{Bernal-Berdun et~al.(2025)Bernal-Berdun, Serrano, Masia, Gadelha,
  Hold-Geoffroy, Sun, and Gutierrez}]{bernal2025precisecam}
Bernal-Berdun, E.; Serrano, A.; Masia, B.; Gadelha, M.; Hold-Geoffroy, Y.; Sun,
  X.; and Gutierrez, D. 2025.
\newblock Precisecam: Precise camera control for text-to-image generation.
\newblock In \emph{Proceedings of the IEEE/CVF Conference on Computer Vision
  and Pattern Recognition}, 2724--2733.

\bibitem[{Bogdan et~al.(2018)Bogdan, Eckstein, Rameau, and
  Bazin}]{bogdan2018deepcalib}
Bogdan, O.; Eckstein, V.; Rameau, F.; and Bazin, J.-C. 2018.
\newblock DeepCalib: A deep learning approach for automatic intrinsic
  calibration of wide field-of-view cameras.
\newblock In \emph{CVMP}, 1--10.

\bibitem[{Caprile and Torre(1990)}]{caprile1990using}
Caprile, B.; and Torre, V. 1990.
\newblock Using vanishing points for camera calibration.
\newblock \emph{IJCV}, 4(2): 127--139.

\bibitem[{Carrera, Angeli, and Davison(2011)}]{carrera2011slam}
Carrera, G.; Angeli, A.; and Davison, A.~J. 2011.
\newblock SLAM-based automatic extrinsic calibration of a multi-camera rig.
\newblock In \emph{IEEE International Conference on Robotics and Automation},
  2652--2659. IEEE.

\bibitem[{Cipolla, Drummond, and Robertson(1999)}]{cipolla1999camera}
Cipolla, R.; Drummond, T.; and Robertson, D.~P. 1999.
\newblock Camera Calibration from Vanishing Points in Image of Architectural
  Scenes.
\newblock In \emph{BMVC}, 382--391.

\bibitem[{Civera et~al.(2009)Civera, Bueno, Davison, and
  Montiel}]{civera2009camera}
Civera, J.; Bueno, D.~R.; Davison, A.~J.; and Montiel, J. M.~M. 2009.
\newblock Camera self-calibration for sequential bayesian structure from
  motion.
\newblock In \emph{IEEE International Conference on Robotics and Automation},
  403--408. IEEE.

\bibitem[{Coughlan and Yuille(1999)}]{coughlan1999manhattan}
Coughlan, J.~M.; and Yuille, A.~L. 1999.
\newblock {Manhattan World: Compass Direction from a Bayesian Inference}.
\newblock In \emph{ICCV}.

\bibitem[{Engel, Koltun, and Cremers(2017)}]{engel2017direct}
Engel, J.; Koltun, V.; and Cremers, D. 2017.
\newblock Direct sparse odometry.
\newblock \emph{IEEE TPAMI}, 40(3): 611--625.

\bibitem[{Fang et~al.(2022)Fang, Vasiljevic, Guizilini, Ambrus, Shakhnarovich,
  Gaidon, and Walter}]{fang2022self}
Fang, J.; Vasiljevic, I.; Guizilini, V.; Ambrus, R.; Shakhnarovich, G.; Gaidon,
  A.; and Walter, M.~R. 2022.
\newblock Self-supervised camera self-calibration from video.
\newblock In \emph{ICRA}, 8468--8475. IEEE.

\bibitem[{Fortier-Chouinard et~al.(2025)Fortier-Chouinard, Hold-Geoffroy,
  Deschaintre, Gadelha, and Lalonde}]{fortier2025gimbaldiffusion}
Fortier-Chouinard, F.; Hold-Geoffroy, Y.; Deschaintre, V.; Gadelha, M.; and
  Lalonde, J.-F. 2025.
\newblock GimbalDiffusion: Gravity-Aware Camera Control for Video Generation.
\newblock \emph{arXiv preprint arXiv:2512.09112}.

\bibitem[{Gordon et~al.(2019)Gordon, Li, Jonschkowski, and
  Angelova}]{gordon2019depth}
Gordon, A.; Li, H.; Jonschkowski, R.; and Angelova, A. 2019.
\newblock Depth from videos in the wild: Unsupervised monocular depth learning
  from unknown cameras.
\newblock In \emph{ICCV}, 8977--8986.

\bibitem[{Hagemann, Knorr, and Stiller(2023)}]{Hagemann2023droidcalib}
Hagemann, A.; Knorr, M.; and Stiller, C. 2023.
\newblock Deep geometry-aware camera self-calibration from video.
\newblock In \emph{ICCV}, 3438--3448.

\bibitem[{Hold-Geoffroy et~al.(2022)Hold-Geoffroy, Piché-Meunier, Sunkavalli,
  Bazin, Rameau, and Lalonde}]{perceptual}
Hold-Geoffroy, Y.; Piché-Meunier, D.; Sunkavalli, K.; Bazin, J.-C.; Rameau,
  F.; and Lalonde, J.-F. 2022.
\newblock {A Deep Perceptual Measure for Lens and Camera Calibration}.
\newblock \emph{IEEE TPAMI}.

\bibitem[{Hold-Geoffroy et~al.(2018)Hold-Geoffroy, Sunkavalli, Eisenmann,
  Fisher, Gambaretto, Hadap, and Lalonde}]{hold2018perceptual}
Hold-Geoffroy, Y.; Sunkavalli, K.; Eisenmann, J.; Fisher, M.; Gambaretto, E.;
  Hadap, S.; and Lalonde, J.-F. 2018.
\newblock A perceptual measure for deep single image camera calibration.
\newblock In \emph{CVPR}, 2354--2363.

\bibitem[{Huang et~al.(2024)Huang, Liu, Zhu, Cheng, Braud, and
  Yeung}]{huang2024360loc}
Huang, H.; Liu, C.; Zhu, Y.; Cheng, H.; Braud, T.; and Yeung, S.-K. 2024.
\newblock 360loc: A dataset and benchmark for omnidirectional visual
  localization with cross-device queries.
\newblock In \emph{Proceedings of the IEEE/CVF conference on computer vision
  and pattern recognition}, 22314--22324.

\bibitem[{Huang et~al.(2025)Huang, Zhou, Rabeti, Korovko, Ling, Ren, Shen, Gao,
  Slepichev, Lin et~al.}]{huang2025vipe}
Huang, J.; Zhou, Q.; Rabeti, H.; Korovko, A.; Ling, H.; Ren, X.; Shen, T.; Gao,
  J.; Slepichev, D.; Lin, C.-H.; et~al. 2025.
\newblock Vipe: Video pose engine for 3d geometric perception.
\newblock \emph{arXiv preprint arXiv:2508.10934}.

\bibitem[{Jeong et~al.(2021)Jeong, Ahn, Choy, Anandkumar, Cho, and
  Park}]{jeong2021self}
Jeong, Y.; Ahn, S.; Choy, C.; Anandkumar, A.; Cho, M.; and Park, J. 2021.
\newblock Self-Calibrating Neural Radiance Fields.
\newblock In \emph{ICCV}.

\bibitem[{Jin et~al.(2023)Jin, Zhang, Hold-Geoffroy, Wang, Blackburn-Matzen,
  Sticha, and Fouhey}]{jin2023perspective}
Jin, L.; Zhang, J.; Hold-Geoffroy, Y.; Wang, O.; Blackburn-Matzen, K.; Sticha,
  M.; and Fouhey, D.~F. 2023.
\newblock Perspective fields for single image camera calibration.
\newblock In \emph{CVPR}, 17307--17316.

\bibitem[{Kani and Snavely(2026)}]{kani2026g3t}
Kani, B. R.~N.; and Snavely, N. 2026.
\newblock G3T Up! Gravity Aligned Coordinate Frames Simplify Pointmap
  Processing.
\newblock \emph{arXiv preprint arXiv:2605.27372}.

\bibitem[{Kannala and Brandt(2006)}]{kannala2006generic}
Kannala, J.; and Brandt, S.~S. 2006.
\newblock A generic camera model and calibration method for conventional,
  wide-angle, and fish-eye lenses.
\newblock \emph{IEEE TPAMI}, 28(8): 1335--1340.

\bibitem[{Kendall and Gal(2017)}]{kendall2017what}
Kendall, A.; and Gal, Y. 2017.
\newblock What uncertainties do we need in bayesian deep learning for computer
  vision?
\newblock \emph{NeurIPS}, 30.

\bibitem[{Khomutenko, Garcia, and Martinet(2016)}]{khomutenko2016enhanced}
Khomutenko, B.; Garcia, G.; and Martinet, P. 2016.
\newblock An enhanced unified camera model.
\newblock In \emph{IEEE International Conference on Robotics and Automation}.

\bibitem[{Kluger et~al.(2020)Kluger, Brachmann, Ackermann, Rother, Yang, and
  Rosenhahn}]{kluger2020consac}
Kluger, F.; Brachmann, E.; Ackermann, H.; Rother, C.; Yang, M.~Y.; and
  Rosenhahn, B. 2020.
\newblock {CONSAC: Robust Multi-Model Fitting by Conditional Sample Consensus}.
\newblock In \emph{CVPR}.

\bibitem[{Ko{\v{s}}eck{\'a} and Zhang(2002)}]{kovsecka2002video}
Ko{\v{s}}eck{\'a}, J.; and Zhang, W. 2002.
\newblock Video compass.
\newblock In \emph{ECCV}, 476--490. Springer.

\bibitem[{Lee et~al.(2021)Lee, Go, Lee, Cho, Sung, and Kim}]{lee2021ctrl}
Lee, J.; Go, H.; Lee, H.; Cho, S.; Sung, M.; and Kim, J. 2021.
\newblock Ctrl-c: Camera calibration transformer with line-classification.
\newblock In \emph{ICCV}.

\bibitem[{Li and Snavely(2018)}]{li2018megadepth}
Li, Z.; and Snavely, N. 2018.
\newblock Megadepth: Learning single-view depth prediction from internet
  photos.
\newblock In \emph{CVPR}, 2041--2050.

\bibitem[{Liang et~al.(2025)Liang, Bhattacharjee, Dey, Moschopoulos, Wang,
  Liao, Tan, Wang, Kayan, Alexandropoulos et~al.}]{liang2025influx}
Liang, E.; Bhattacharjee, R.; Dey, S.; Moschopoulos, R.; Wang, C.; Liao, M.;
  Tan, G.; Wang, A.; Kayan, K.; Alexandropoulos, S.; et~al. 2025.
\newblock InFlux: A Benchmark for Self-Calibration of Dynamic Intrinsics of
  Video Cameras.
\newblock \emph{arXiv preprint arXiv:2510.23589}.

\bibitem[{Lin et~al.(2025)Lin, Cen, Jiang, Karhade, Wang, Mitra, Ling, Huang,
  Zawar, Bai et~al.}]{lintowards}
Lin, Z.; Cen, S.; Jiang, D.; Karhade, J.; Wang, H.; Mitra, C.; Ling, Y. T.~T.;
  Huang, Y.; Zawar, R.; Bai, X.; et~al. 2025.
\newblock Towards Understanding Camera Motions in Any Video.
\newblock In \emph{The Thirty-ninth Annual Conference on Neural Information
  Processing Systems Datasets and Benchmarks Track}.

\bibitem[{Lochman et~al.(2021{\natexlab{a}})Lochman, Dobosevych, Hryniv, and
  Pritts}]{lochman2021minimal}
Lochman, Y.; Dobosevych, O.; Hryniv, R.; and Pritts, J. 2021{\natexlab{a}}.
\newblock {Minimal solvers for single-view lens-distorted camera
  auto-calibration}.
\newblock In \emph{WACV}, 2887--2896.

\bibitem[{Lochman et~al.(2021{\natexlab{b}})Lochman, Liepieshov, Chen, Perdoch,
  Zach, and Pritts}]{lochman2021babelcalib}
Lochman, Y.; Liepieshov, K.; Chen, J.; Perdoch, M.; Zach, C.; and Pritts, J.
  2021{\natexlab{b}}.
\newblock Babelcalib: A universal approach to calibrating central cameras.
\newblock In \emph{ICCV}, 15253--15262.

\bibitem[{Lopez et~al.(2019)Lopez, Mari, Gargallo, Kuang, Gonzalez-Jimenez, and
  Haro}]{lopez2019deepcalib}
Lopez, M.; Mari, R.; Gargallo, P.; Kuang, Y.; Gonzalez-Jimenez, J.; and Haro,
  G. 2019.
\newblock Deep single image camera calibration with radial distortion.
\newblock In \emph{CVPR}, 11817--11825.

\bibitem[{Mei and Rives(2007)}]{mei2007single}
Mei, C.; and Rives, P. 2007.
\newblock Single view point omnidirectional camera calibration from planar
  grids.
\newblock In \emph{ICRA}, 3945--3950. IEEE.

\bibitem[{Oquab et~al.(2023)Oquab, Darcet, Moutakanni, Vo, Szafraniec,
  Khalidov, Fernandez, Haziza, Massa, El-Nouby et~al.}]{oquab2023dinov2}
Oquab, M.; Darcet, T.; Moutakanni, T.; Vo, H.; Szafraniec, M.; Khalidov, V.;
  Fernandez, P.; Haziza, D.; Massa, F.; El-Nouby, A.; et~al. 2023.
\newblock Dinov2: Learning robust visual features without supervision.
\newblock \emph{arXiv preprint arXiv:2304.07193}.

\bibitem[{Pan et~al.(2023)Pan, Charron, Yang, Peters, Whelan, Kong, Parkhi,
  Newcombe, and Ren}]{pan2023aria}
Pan, X.; Charron, N.; Yang, Y.; Peters, S.; Whelan, T.; Kong, C.; Parkhi, O.;
  Newcombe, R.; and Ren, Y.~C. 2023.
\newblock {Aria Digital Twin}: A New Benchmark Dataset for Egocentric 3D
  Machine Perception.
\newblock In \emph{ICCV}.

\bibitem[{Pautrat et~al.(2023)Pautrat, Liu, Hruby, Pollefeys, and
  Barath}]{Pautrat2023uvp}
Pautrat, R.; Liu, S.; Hruby, P.; Pollefeys, M.; and Barath, D. 2023.
\newblock Vanishing point estimation in uncalibrated images with prior gravity
  direction.
\newblock In \emph{ICCV}, 14118--14127.

\bibitem[{Pollefeys and Van~Gool(1997)}]{pollefeys1997stratified}
Pollefeys, M.; and Van~Gool, L. 1997.
\newblock A stratified approach to metric self-calibration.
\newblock In \emph{CVPR}, 407--412. IEEE.

\bibitem[{Pritts et~al.(2020)Pritts, Kukelova, Larsson, Lochman, and
  Chum}]{pritts2020minimal}
Pritts, J.; Kukelova, Z.; Larsson, V.; Lochman, Y.; and Chum, O. 2020.
\newblock {Minimal Solvers for Rectifying from Radially-Distorted Conjugate
  Translations}.
\newblock \emph{IEEE TPAMI}.

\bibitem[{Qin, Li, and Shen(2018)}]{qin2018vins}
Qin, T.; Li, P.; and Shen, S. 2018.
\newblock {VINS-Mono}: A Robust and Versatile Monocular Visual-Inertial State
  Estimator.
\newblock \emph{IEEE Transactions on Robotics}, 34(4): 1004--1020.

\bibitem[{Ranftl, Bochkovskiy, and Koltun(2021)}]{ranftl2021vision}
Ranftl, R.; Bochkovskiy, A.; and Koltun, V. 2021.
\newblock Vision transformers for dense prediction.
\newblock In \emph{ICCV}, 12179--12188.

\bibitem[{Schonberger and Frahm(2016)}]{schoenberger2016sfm}
Schonberger, J.~L.; and Frahm, J.-M. 2016.
\newblock Structure-from-motion revisited.
\newblock In \emph{CVPR}, 4104--4113.

\bibitem[{Shah et~al.(2023)Shah, Osi{\'n}ski, Levine et~al.}]{shah2023lm}
Shah, D.; Osi{\'n}ski, B.; Levine, S.; et~al. 2023.
\newblock Lm-nav: Robotic navigation with large pre-trained models of language,
  vision, and action.
\newblock In \emph{Conference on robot learning}, 492--504. pmlr.

\bibitem[{Shi et~al.(2016)Shi, Caballero, Huszar, Totz, Aitken, Bishop,
  Rueckert, and Wang}]{Shi2016pixelshuffle}
Shi, W.; Caballero, J.; Huszar, F.; Totz, J.; Aitken, A.~P.; Bishop, R.;
  Rueckert, D.; and Wang, Z. 2016.
\newblock Real-Time Single Image and Video Super-Resolution Using an Efficient
  Sub-Pixel Convolutional Neural Network.
\newblock In \emph{Proceedings of the IEEE Conference on Computer Vision and
  Pattern Recognition}.

\bibitem[{Song et~al.(2024)Song, Kang, Moteki, Suzuki, Kobayashi, and
  Tan}]{Song2024MSCC}
Song, X.; Kang, H.; Moteki, A.; Suzuki, G.; Kobayashi, Y.; and Tan, Z. 2024.
\newblock {Mscc: Multi-scale transformers for camera calibration}.
\newblock In \emph{WACV}, 3262--3271.

\bibitem[{Tirado-Gar{\'\i}n and Civera(2025)}]{tirado2025anycalib}
Tirado-Gar{\'\i}n, J.; and Civera, J. 2025.
\newblock AnyCalib: On-manifold learning for model-agnostic single-view camera
  calibration.
\newblock In \emph{Proceedings of the IEEE/CVF International Conference on
  Computer Vision}, 8044--8055.

\bibitem[{Usenko, Demmel, and Cremers(2018)}]{usenko2018double}
Usenko, V.; Demmel, N.; and Cremers, D. 2018.
\newblock The Double Sphere Camera Model.
\newblock In \emph{3DV}, 552--560.

\bibitem[{Vasiljevic et~al.(2020)Vasiljevic, Guizilini, Ambrus, Pillai,
  Burgard, Shakhnarovich, and Gaidon}]{vasiljevic2020neural}
Vasiljevic, I.; Guizilini, V.; Ambrus, R.; Pillai, S.; Burgard, W.;
  Shakhnarovich, G.; and Gaidon, A. 2020.
\newblock Neural Ray Surfaces for Self-Supervised Learning of Depth and
  Ego-motion.
\newblock In \emph{Proceedings of the International Conference on 3D Vision
  (3DV)}.

\bibitem[{Veicht et~al.(2024)Veicht, Sarlin, Lindenberger, and
  Pollefeys}]{veicht2024geocalib}
Veicht, A.; Sarlin, P.-E.; Lindenberger, P.; and Pollefeys, M. 2024.
\newblock Geocalib: Learning single-image calibration with geometric
  optimization.
\newblock In \emph{European Conference on Computer Vision}, 1--20. Springer.

\bibitem[{Wallingford et~al.(2024)Wallingford, Bhattad, Kusupati, Ramanujan,
  Deitke, Kembhavi, Mottaghi, Ma, and Farhadi}]{wallingford2024image}
Wallingford, M.; Bhattad, A.; Kusupati, A.; Ramanujan, V.; Deitke, M.;
  Kembhavi, A.; Mottaghi, R.; Ma, W.-C.; and Farhadi, A. 2024.
\newblock From an image to a scene: Learning to imagine the world from a
  million 360 videos.
\newblock \emph{NeurIPS}, 37: 17743--17760.

\bibitem[{Wang et~al.(2025{\natexlab{a}})Wang, Chen, Karaev, Vedaldi,
  Rupprecht, and Novotny}]{wang2025vggt}
Wang, J.; Chen, M.; Karaev, N.; Vedaldi, A.; Rupprecht, C.; and Novotny, D.
  2025{\natexlab{a}}.
\newblock Vggt: Visual geometry grounded transformer.
\newblock In \emph{CVPR}, 5294--5306.

\bibitem[{Wang et~al.(2024)Wang, Leroy, Cabon, Chidlovskii, and
  Revaud}]{wang2024dust3r}
Wang, S.; Leroy, V.; Cabon, Y.; Chidlovskii, B.; and Revaud, J. 2024.
\newblock Dust3r: Geometric 3d vision made easy.
\newblock In \emph{CVPR}, 20697--20709.

\bibitem[{Wang et~al.(2020)Wang, Zhu, Wang, Hu, Qiu, Wang, Hu, Kapoor, and
  Scherer}]{wang2020tartanair}
Wang, W.; Zhu, D.; Wang, X.; Hu, Y.; Qiu, Y.; Wang, C.; Hu, Y.; Kapoor, A.; and
  Scherer, S. 2020.
\newblock Tartanair: A dataset to push the limits of visual slam.
\newblock In \emph{IROS}, 4909--4916. IEEE.

\bibitem[{Wang et~al.(2025{\natexlab{b}})Wang, Pan, Pollefeys, and
  Larsson}]{wang2025structure}
Wang, Y.; Pan, L.; Pollefeys, M.; and Larsson, V. 2025{\natexlab{b}}.
\newblock Structure-From-Motion with a Non-Parametric Camera Model.
\newblock In \emph{CVPR}, 1040--1049.

\bibitem[{Wang et~al.(2021)Wang, Wu, Xie, Chen, and
  Prisacariu}]{wang2021neural}
Wang, Z.; Wu, S.; Xie, W.; Chen, M.; and Prisacariu, V.~A. 2021.
\newblock Neural radiance fields without known camera parameters.
\newblock \emph{arXiv preprint arXiv:2102.07064}.

\bibitem[{Xian et~al.(2019)Xian, Li, Fisher, Eisenmann, Shechtman, and
  Snavely}]{xian2019uprightnet}
Xian, W.; Li, Z.; Fisher, M.; Eisenmann, J.; Shechtman, E.; and Snavely, N.
  2019.
\newblock {UprightNet: Geometry-Aware Camera Orientation Estimation from Single
  Images}.
\newblock In \emph{ICCV}.

\bibitem[{Yeshwanth et~al.(2023)Yeshwanth, Liu, Nie{\ss}ner, and
  Dai}]{yeshwanth2023scannetpp}
Yeshwanth, C.; Liu, Y.-C.; Nie{\ss}ner, M.; and Dai, A. 2023.
\newblock {ScanNet++}: A High-Fidelity Dataset of 3D Indoor Scenes.
\newblock In \emph{ICCV}.

\bibitem[{Zhang et~al.(2026{\natexlab{a}})Zhang, Li, Wei, Cao, Gambardella,
  Phung, and Cai}]{zhang2026unified}
Zhang, C.; Li, B.; Wei, M.; Cao, Y.-P.; Gambardella, C.~C.; Phung, D.; and Cai,
  J. 2026{\natexlab{a}}.
\newblock Unified Camera Positional Encoding for Controlled Video Generation.
\newblock In \emph{Proceedings of the Computer Vision and Pattern Recognition
  Conference}.

\bibitem[{Zhang et~al.(2026{\natexlab{b}})Zhang, Liang, Chen, Wu, Plataniotis,
  Gambardella, and Cai}]{zhang2025panflow}
Zhang, C.; Liang, H.; Chen, D.~Y.; Wu, Q.; Plataniotis, K.~N.; Gambardella,
  C.~C.; and Cai, J. 2026{\natexlab{b}}.
\newblock PanFlow: Decoupled Motion Control for Panoramic Video Generation.
\newblock In \emph{Proceedings of the AAAI Conference on Artificial
  Intelligence}.

\bibitem[{Zhang(1999)}]{zhang1999flexible}
Zhang, Z. 1999.
\newblock Flexible camera calibration by viewing a plane from unknown
  orientations.
\newblock In \emph{Proceedings of the seventh ieee international conference on
  computer vision}, volume~1, 666--673. Ieee.

\bibitem[{Zhang(2000)}]{zhang2000calibration}
Zhang, Z. 2000.
\newblock A flexible new technique for camera calibration.
\newblock \emph{IEEE TPAMI}, 22(11): 1330--1334.

\end{thebibliography}

\appendix
\twocolumn[%
  \begin{center}
    {\LARGE\bfseries Supplementary Material\par}
  \end{center}
  \vspace{1.5em}
]

\providecommand{\suppref}[1]{Supp-\cref{#1}}

\section{Framework Overview}
\label{sec:supp_overview}

\begin{figure}[t]
    \centering
    \includegraphics[width=\linewidth]{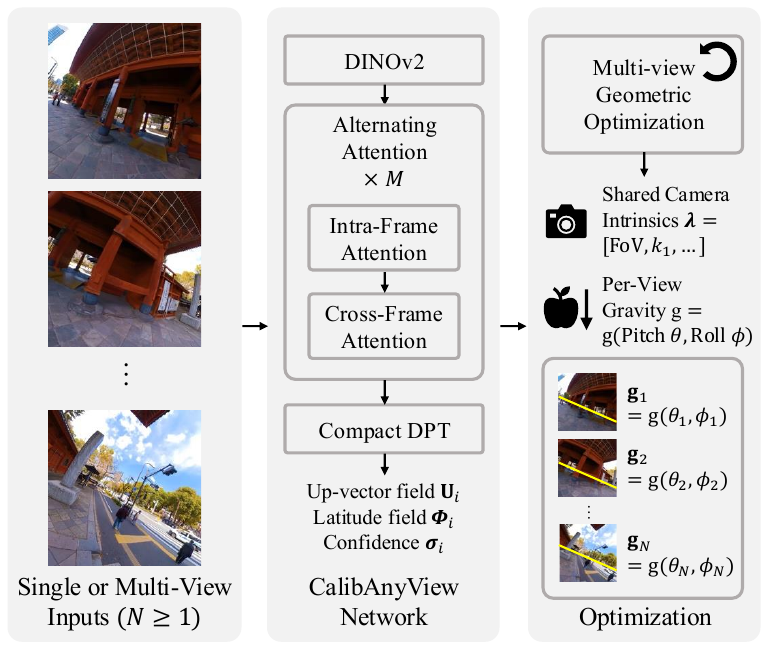}
    \caption{\textbf{Overview of the proposed \themethod framework.} Given single- or multi-view inputs ($N \geq 1$), the network uses DINOv2 and an alternating attention mechanism to extract geometric features. A DPT head then predicts dense perspective fields (up-vector and latitude) alongside confidence maps, which are fed into a multi-view geometric optimization that recovers the shared camera intrinsics and the per-view gravity.}
    \label{fig:supp_pipeline}
\end{figure}

\suppref{fig:supp_pipeline} shows the framework, which includes three stages, all three of which are introduced in \cref{sec:framework} of the main paper and expanded in the remainder of this document.

\textbf{Input.} The framework takes $N \geq 1$ views that are assumed to come from one physical camera, so a single intrinsic is shared across them while the gravity direction differs from view to view. The single-view case is not a separate mode but the same weights run with $N = 1$. 

\textbf{Network.} A DINOv2 backbone is followed by $M{=}24$ alternating attention blocks, each pairing intra-frame with cross-frame attention so that geometric evidence is exchanged between views before anything is decoded. A compact DPT head then reads intermediate tokens from the latter stages of the aggregator, specifically the outputs of layers $15$, $18$, $21$ and $23$, and fuses these multi-scale features with progressive upsampling to predict, for every view $i$, the up-vector field $\hat{\mathbf{U}}_i$, the latitude field $\hat{\Phi}_i$ and a per-pixel confidence map $\sigma_i$ at $1/4$ of the input resolution.

\textbf{Optimizer.} The predicted fields are passed to a differentiable Levenberg--Marquardt solver, which recovers the shared $\bm{\lambda}$ together with a per-view gravity $\mathbf{g}_i$ by minimising the confidence-weighted field residual. \suppref{sec:supp_eucm_solver} derives the closed-form EUCM initialisation the solver starts from.

\textbf{Training.} We initialise the aggregator from VGGT~\citep{wang2025vggt} and keep the DINOv2 backbone frozen throughout. The model is trained for 20 epochs with AdamW, a weight decay of $0.05$ and a peak learning rate of $5 \times 10^{-5}$, with each step packing at most $96$ images per GPU; the cap counts images rather than sequences, since the sequence length varies. Every training sample draws $N \in [2, 24]$ frames uniformly at random from the $81$ frames of a clip, so that the same weights see both nearly redundant and widely separated views. Supervision is the confidence-weighted field loss of \cref{sec:framework}, with $\gamma = 1$ and $\eta = 0.2$. We apply data augmentations across all datasets, including Gaussian noise, color jittering, motion blur, and photometric perturbations, to simulate the variability of real-world captures. The multi-view data that supplies this supervision is described in \cref{sec:dataset} of the main paper, with the construction details in \suppref{sec:supp_dataset}.

\section{Camera Models and Solver Details}
\label{sec:supp_camera_models}

\subsection{Projection Models}
This section gives in full the three projection models summarized in \cref{sec:preliminaries}, which map a 3D point $\mathbf{P} = [X, Y, Z]^\top$ in the camera frame to a pixel $\mathbf{p} = [u, v]^\top$.
All three share the same second stage, $\mathbf{p} = f \cdot \mathbf{x} + \mathbf{c}$, where $\mathbf{x}$ is a point on the normalized image plane, $f$ is the focal length and $\mathbf{c} = [c_x, c_y]^\top$ is the principal point, which we fix to half the image dimensions throughout.
They differ only in how $\mathbf{P}$ is mapped to $\mathbf{x}$:
\begin{itemize}
    \item \textbf{Pinhole}: a plain perspective division, $\mathbf{x} = [X/Z, Y/Z]^\top$, so that $f$ alone determines the field of view.
    \item \textbf{Simple Radial}: the pinhole point is displaced along the radius, $\mathbf{x} \mapsto (1 + k_1 \|\mathbf{x}\|^2)\, \mathbf{x}$, with a single coefficient $k_1$~\citep{zhang2000calibration} that captures the barrel or pincushion distortion of consumer lenses.
    \item \textbf{EUCM}~\citep{khomutenko2016enhanced}: a generalization of the Unified Camera Model~\citep{mei2007single} that projects through an ellipsoidal rather than a spherical surface,
    \begin{equation}
        \mathbf{x} = \frac{[X, Y]^\top}{\alpha \sqrt{\beta (X^2 + Y^2) + Z^2} + (1 - \alpha) Z},
    \end{equation}
    whose shape parameters $(\alpha, \beta)$ cover wide-angle through fisheye optics with a single closed-form model, and which degenerates to the pinhole case at $\alpha = 0$.
\end{itemize}
The intrinsic parameter set of \cref{sec:preliminaries} is therefore $\bm{\lambda} = \{f\}$, $\{f, k_1\}$ and $\{f, \alpha, \beta\}$ for the three models, respectively.

\subsection{EUCM Solver Details}
\label{sec:supp_eucm_solver}

This section expands the two-stage EUCM solver of \cref{sec:framework}. Unlike
methods that regress rays directly~\citep{tirado2025anycalib}, our observations
are the perspective fields themselves, from which the intrinsics cannot be
recovered by a linear system: rendering the up-vector field requires the
Jacobian of the projection, so the forward model is not linear in
$\bm{\lambda}$ under any reparameterization. The role of the closed-form stage
below is therefore to place the subsequent refinement inside the basin of the
correct minimum, without which that stage does not converge.

\subsubsection{Closed-Form Initialization from the Latitude Field}
\label{sec:supp_annulus}

The stage uses the latitude field only, leaving the up-vector field to the
refinement. For a pixel $\mathbf{p}$ we write $\rho$ for its distance to the
principal point, $\varphi$ for its polar image angle, and $\vartheta$ for the
polar angle of the corresponding ray; $r \mathrel{:=} \rho / f$ is the
normalized radius and $(R, Z) \mathrel{:=} (\sin\vartheta, \cos\vartheta)$ the
radial and axial components of the unit ray.

\paragraph{Annulus demodulation}
A radially symmetric camera preserves the pixel azimuth, so all pixels on an
annulus of radius $\rho$ share the same $\vartheta$, and the latitude restricted
to that annulus is a pure sinusoid in the polar image angle,
\begin{equation}
\label{eq:annulus_sinusoid}
    \sin \phi_{\mathbf{p}} = A(\rho) \cos(\varphi - \varphi_g) + B(\rho) ,
\end{equation}
whose three coefficients follow from a per-annulus linear fit.

\paragraph{Gravity}
With the gravity direction written as
$\mathbf{g} = [\,\rho_g \cos\varphi_g,~ \rho_g \sin\varphi_g,~ g_z\,]^\top$ and
$\rho_g^2 + g_z^2 = 1$, the amplitude and the offset of
\cref{eq:annulus_sinusoid} are $A = \rho_g \sin\vartheta$ and
$B = g_z \cos\vartheta$, while its phase is the azimuth $\varphi_g$ of the
gravity itself. Eliminating $\vartheta$ therefore leaves, for every annulus,
\begin{equation}
\label{eq:annulus_gravity}
    \frac{A^2}{\rho_g^2} + \frac{B^2}{g_z^2} = 1 ,
\end{equation}
an over-determined system in the single unknown $g_z^2$, from which roll and
pitch follow together with $\varphi_g$. Gravity is therefore recovered before
any intrinsic is known and, because $A$ and $B$ are global properties of an
annulus, without the branch ambiguity and the near-horizon degeneracy that a
per-pixel inversion of $\Phi$ would incur.

\paragraph{Radial profile and focal length}
Each annulus now yields a ray through
$\vartheta_i = \operatorname{atan2}(A_i / \rho_g,~ B_i / g_z)$, which reduces
the problem to a one-dimensional radius--elevation correspondence. Linearity in
the intrinsics is lost when $f$ is unknown, so we read it off a Kannala-Brandt
proxy~\citep{kannala2006generic},
\begin{equation}
\label{eq:kb_proxy}
    \rho = f\vartheta + \textstyle\sum_{n=1}^{3} (f k_n)\, \vartheta^{2n+1} ,
\end{equation}
which is linear in $\{f,~ f k_n\}$ and yields practically the same focal length
as the EUCM one~\citep{usenko2018double, lochman2021babelcalib}, the linear term
being the focal length of any radially symmetric camera as $\vartheta \to 0$.
The auxiliary coefficients $k_n$ are discarded afterwards.

\paragraph{Shape parameters}
With $f$ fixed, clearing the square root of the EUCM projection leaves
\begin{equation}
\label{eq:eucm_linear}
    r^2 R^2\, \mu ~+~ 2 r Z (r Z - R)\, \alpha ~=~ (R - r Z)^2 ,
    \qquad \mu \mathrel{:=} \alpha^2 \beta ,
\end{equation}
linear in $(\mu, \alpha)$ and solved subject to $\alpha \in [0, 1]$ and
$\mu \geq 0$. As $\alpha \to 0$ the projection becomes independent of $\beta$,
which is then unidentifiable and held at $1$.

\paragraph{Alternation}
Since $f$ given $(\alpha, \beta)$ and $(\alpha, \beta)$ given $f$ are both
exactly linear, we alternate the two solves for a few rounds, falling back to a
neutral guess if too few annuli survive or the recovered focal length is
degenerate.

\subsubsection{Field Residual and Refinement}
\label{sec:supp_eucm_lm}

Given the initialization, the refinement minimizes the discrepancy between the
predicted fields and those induced by the parameters,
\begin{equation}
\label{eq:eucm_residual}
    \mathbf{r}(\mathbf{s}) =
    \begin{bmatrix}
        \mathbf{U}(\mathbf{s}) - \hat{\mathbf{U}} \\[2pt]
        \sin \Phi(\mathbf{s}) - \sin \hat{\Phi}
    \end{bmatrix} ,
\end{equation}
with the two channels weighted equally and the latitude compared through its
sine, which removes the wrap-around of the raw angle.
A Levenberg-Marquardt solver then minimizes this residual to recover the gravity
and the intrinsics jointly. Because the residual of each view depends only on
the intrinsics shared across the sequence and on that view's own gravity, the
normal equations are block-arrow and we assemble them per view: this leaves the
solution unchanged while making the Jacobian cost grow linearly rather than
quadratically in the number of views.

\section{Evaluation Benchmarks}
\label{sec:supp_benchmarks}

\begin{table*}[t]
\centering
\small
\setlength\tabcolsep{8pt}
\renewcommand{\arraystretch}{1.15}
\begin{tabular}{llccccc}
\toprule
Benchmark & GT camera model & \makecell{Distinct\\cameras}
          & \makecell{vFoV\\{}[\degree]} & Resolution & Color & Parallax \\
\midrule
Stanford2D3D  & pinhole, radial, EUCM & 2794 & $20$--$180$ & $640^2$        & RGB  & $\times$   \\
Aria-ADT-RGB  & Fisheye624            & 3    & $102$       & ${\sim}1280^2$ & RGB  & \checkmark \\
Aria-ADT-Gray & Fisheye624            & 6    & $111$       & ${\sim}476^2$  & Gray & \checkmark \\
ScanNet++     & Kannala--Brandt       & 30   & $86$--$103$ & $640^2$        & RGB  & \checkmark \\
\bottomrule
\end{tabular}
\caption{\textbf{The four multi-view evaluation benchmarks, none of which is seen
during training.} All are evaluated on $100$ clips of $81$ frames each, from which
$N$ views are drawn uniformly. \emph{Distinct cameras} counts the independent sets
of intrinsic parameters: on Stanford2D3D a camera is sampled per clip, whereas on
the other three they are fixed by the device or by the scene, which is also why
the two Aria resolutions vary slightly from device to device. \emph{Parallax}
marks whether the camera translates at all: Stanford2D3D is rendered from
panoramas and therefore rotates in place.}
\label{tbl:supp_benchmarks}
\end{table*}

We evaluate on four benchmarks, whose quantitative results are reported in
\cref{tbl:multiview_public} of the main paper and whose properties are summarised
in \suppref{tbl:supp_benchmarks}: Stanford2D3D~\citep{armeni2017joint},
Aria-ADT~\citep{pan2023aria}, whose RGB and grayscale streams enter as two
separate benchmarks, and ScanNet++~\citep{yeshwanth2023scannetpp}.
\textbf{None of them is seen during training}, so the results in that table test the generalisation ability of the methods.
The whole evaluation benchmarks cover a wide variety of cameras, a wide range of distortion and field of view, and diverse and
demanding camera motion, rather than a single regime.
Each benchmark consists of $100$ clips of $81$ frames. The $N$ evaluation views
are drawn uniformly over the whole clip, and every method is run on the
same clips.

\paragraph{Stanford2D3D}
Clips are rendered from the panoramas of Stanford2D3D~\citep{armeni2017joint}
along augmented rotation trajectories, following the process of
GeoCalib~\citep{veicht2024geocalib} and AnyCalib~\citep{tirado2025anycalib}. What
makes this benchmark valuable is that the camera varies from clip to clip:
combining a vertical field of view that ranges from $20^\circ$ to $180^\circ$
with three projection models yields thousands of distinct cameras, together
spanning essentially the whole range of everyday optics. It is at the same time
the most demanding of the four, because rendering from a panorama keeps the
camera centre fixed: the sequences contain rotation but no translation at all.

\paragraph{Aria-ADT}
Egocentric recordings from Aria Digital Twin~\citep{pan2023aria}, captured with a
head-mounted device along real-life trajectories, which come with an accurate
gravity ground truth. We evaluate the \textbf{RGB stream} and the \textbf{grayscale streams}
as two separate benchmarks because they differ in almost every respect: different
capture sensors, different intrinsics, different distortion and different field of
view, and one is color while the other is grayscale. Our training data contains
color images only, so the Gray streams additionally test how the methods
generalize to grayscale input. Both are Fisheye624 cameras whose parameters are
fixed by the physical device, and both were cropped and rotated so that the
optical centre sits at the image centre and the image is upright with respect to
gravity. Being head-mounted, these captures look persistently downwards.

\paragraph{ScanNet++}
Handheld DSLR scans of indoor scenes~\citep{yeshwanth2023scannetpp}, with one
fisheye lens per scene and camera poses taken from the released reconstructions.
ScanNet++ registers its poses to the frame of a terrestrial laser scanner whose
dual-axis compensator levels every scan, and releases its scenes with the $+Z$
axis upright, so its world vertical is gravity by construction of the capture
hardware. We keep only the frames the dataset marks as good, discard the
defective ones, and group the remainder into clips. These clips carry the largest
parallax of the four and none of them is static, which is the favorable regime that
reconstruction-based methods are designed for. The distortion range is the
narrowest, however: the field of view stays close to $100^\circ$ throughout, so
this benchmark probes stability on real lenses rather than coverage of strong
distortion.

\paragraph{Qualitative results}
Perspective field visualizations are available for all four benchmarks: \cref{fig:multiview_public_qual} of the main paper shows Stanford2D3D, and \suppref{fig:supp_aria_rgb}, \suppref{fig:supp_aria_gray} and \suppref{fig:supp_scannetpp} below show the other three. Taken together, they show that our method attains higher accuracy than the baselines, stays robust across sensors as different as a color and a grayscale fisheye, and adapts to lenses whose distortion is far stronger than a near-pinhole geometry can absorb.

\section{Multi-view Analysis}
\label{sec:supp_multiview}

Two conventions recur in the main paper and in this document.
First, GeoCalib$_\text{pin}$ and GeoCalib$_\text{dist}$ denote the pinhole and
the distortion-aware variants of GeoCalib~\citep{veicht2024geocalib}, and
AnyCalib$_p$ and AnyCalib$_g$ the pinhole-only and the general variants of
AnyCalib~\citep{tirado2025anycalib}, in each case two separately released
checkpoints.
Second, a configuration is written as prediction $+$ optimization, with
\textbf{S} and \textbf{M} standing for single- and multi-view: \textbf{S$+$S}
predicts and optimizes every frame on its own, \textbf{S$+$M} still predicts
each frame independently but recovers a single intrinsic shared over the whole
sequence, and \textbf{M$+$M} additionally lets the network attend across views
before anything is decoded.
The GeoCalib rows of \cref{tbl:multiview_public} in the main paper are run in
\textbf{S$+$M}.

\begin{figure*}[t]
    \centering
    \includegraphics[width=0.98\textwidth]{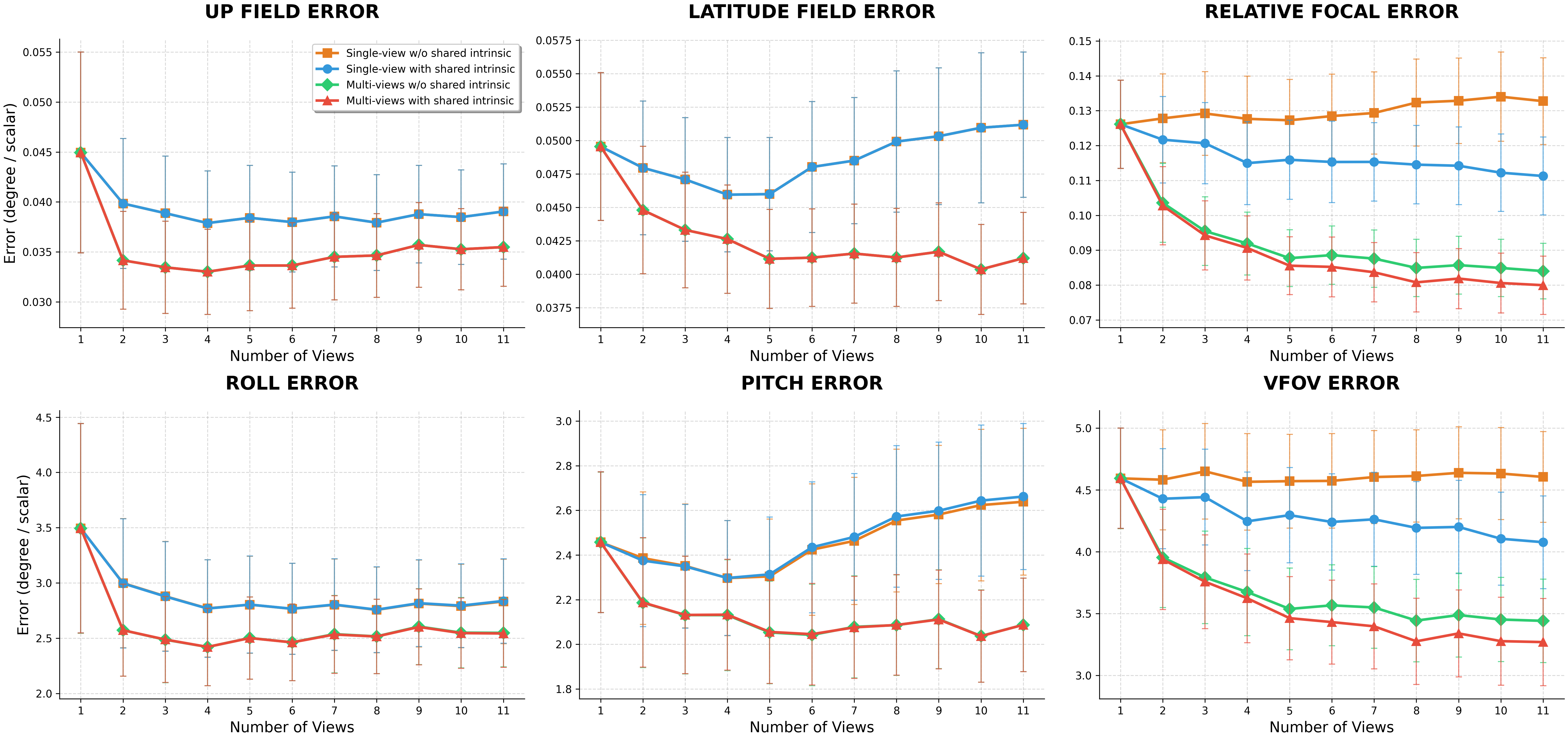}
    \caption{
        \textbf{Calibration performance vs.\ number of views.}
        We evaluate the impact of the number of input views on calibration error, reporting both the mean and standard deviation.
        To analyze the effect of cross-view reasoning, we compare two settings: (i) \textit{individual processing} (Single-view), where views are input and processed independently to estimate dense maps without cross-view interaction; and (ii) \textit{joint processing} (Multi-view), where all views are input simultaneously to enable cross-view attention.
        At the optimization stage, we further compare joint optimization (with shared intrinsics) against independent per-frame optimization.
        Joint processing with cross-view attention significantly outperforms the individual-view baseline by effectively aggregating cross-frame geometric cues.
        Furthermore, enforcing shared intrinsics during optimization improves focal length and vFoV estimation.
        Overall, the error drops as the number of views increases, demonstrating that our alternating attention mechanism successfully resolves single-view geometric ambiguities.
    }\label{fig:exp_multiview}
\end{figure*}

\begin{table*}[t]
\centering
\small
\renewcommand{\arraystretch}{1.1}
\setlength\tabcolsep{3.0pt}
\begin{tabular}{l c cc cc cc}
\toprule
\multirow{2}{*}{Method} & \multirow{2}{*}{Mode} & \multicolumn{2}{c}{Gravity [$^\circ$]} & \multicolumn{2}{c}{Intrinsics} & \multicolumn{2}{c}{Perspective Fields} \\
\cmidrule(lr){3-4} \cmidrule(lr){5-6} \cmidrule(lr){7-8}
& & Roll [$^\circ$] $\downarrow$ & Pitch [$^\circ$] $\downarrow$ & FoV [$^\circ$] $\downarrow$ & Focal [\%] $\downarrow$ & Up $\downarrow$ & Lat. $\downarrow$ \\
\midrule
AnyCalib$_{p}$~\citeyearpar{tirado2025anycalib} & S+S & - & - & 4.62 & 12.43 & - & - \\
AnyCalib$_{g}$~\citeyearpar{tirado2025anycalib}     & S+S & - & - & 4.98 & 12.53 & - & - \\
VGGT~\citeyearpar{wang2025vggt}                            & M & - & - & 7.09 & 17.36 & - & - \\
\midrule
GeoCalib~\citeyearpar{veicht2024geocalib}    & S+S & 6.22 & 5.28 & 6.10 & 15.01 & 0.17 & 0.11 \\
GeoCalib~\citeyearpar{veicht2024geocalib}    & S+M  & 6.19 & 5.39 & 5.09 & 12.19 & 0.17 & 0.11 \\
GeoCalib$_\text{dist}$~\citeyearpar{veicht2024geocalib}  & S+S & 5.84 & 5.42 & 6.23 & 15.23 & 0.13 & 0.11 \\
GeoCalib$_\text{dist}$~\citeyearpar{veicht2024geocalib}  & S+M  & 5.84 & 5.51 & 5.10 & \textbf{11.95} & 0.13 & 0.11 \\
GeoCalib$_{\text{finetuned}}$~\citeyearpar{veicht2024geocalib}  & S+S & 4.98 & 4.83 & 6.49 & 14.09 & \textbf{0.07} & 0.10 \\
GeoCalib$_{\text{finetuned}}$~\citeyearpar{veicht2024geocalib}  & S+M & 5.00 & 4.74 & 5.60 & 12.25 & \textbf{0.07} & 0.10 \\
\midrule
\textbf{CalibAnyView (Ours)} & S+S & \textbf{3.22} & \textbf{3.00} & \textbf{4.54} & 12.05 & \textbf{0.04} & \textbf{0.06} \\
\textbf{CalibAnyView (Ours)} & M+M & \textbf{2.62} & \textbf{2.43} & \textbf{3.88} & \textbf{9.94} & \textbf{0.04} & \textbf{0.04} \\
\bottomrule
\end{tabular}
\caption{\textbf{Quantitative Results on our Dataset Test Set.} We compare our multi-view framework against state-of-the-art single-view calibration baselines. ``Mode'' indicates the combination of prediction and optimization: \textbf{S}/\textbf{M} stands for \textbf{S}ingle or \textbf{M}ulti-view for prediction (first letter) and optimization (second letter). Errors are reported as mean values across the entire test set. The symbol ``-'' indicates the parameter is not supported for prediction by the model. \textbf{Bold} marks the best and the second-best entry of each column.}
\label{tbl:supp_test_results}
\end{table*}

\subsection{Multi-view Prediction and Optimization}
\label{sec:supp_multiview_protocol}

This section gives the full protocol and the per-parameter results behind the summary of \cref{sec:multiview_analysis}.
We evaluate the multi-view capability of our trained model on the test split of our proposed dataset, which contains 82 video sequences.

For each video, we uniformly sample 20 frames and group them into batches of size $N \in \{1, 2, \ldots, 11\}$. Our inference pipeline consists of two stages: (1)~the network predicts dense perspective fields from the input frames, and (2)~a geometric optimizer recovers the camera parameters from the predicted fields. To disentangle the contributions of the multi-view network and the multi-view optimizer, we design a $2 \times 2$ experimental protocol. On the network side, each batch is fed either as independent single-image inputs (Single-view) without cross-view interaction, or as a joint multi-view input (Multi-view) to activate cross-view attention, yielding the per-batch perspective fields. On the optimizer side, the resulting fields are then processed either with shared-intrinsic joint optimization (with shared intrinsics) or with independent per-frame optimization (without shared intrinsics). This yields four configurations whose results are visualized in \suppref{fig:exp_multiview}.

As shown in \suppref{fig:exp_multiview}, for the network-predicted (Up and Latitude fields), and gravity outputs (Roll and Pitch errors), the joint multi-view input consistently yields substantially lower errors than the single-view counterpart. This demonstrates that, through our alternating attention design and multi-view training regime, the network acquires strong cross-frame geometric reasoning capabilities.
Examining the Relative Focal error and vFoV error trends reveals that Multi-views consistently achieves lower errors than Single-view, regardless of which optimizer configuration is employed. Moreover, within the Single-view and Multi-view settings separately, shared-intrinsic optimization consistently outperforms independent per-frame optimization, with the gap widening as the number of views increases.
Overall, the calibration error drops sharply as the number of views grows, demonstrating that our framework effectively resolves the inherent geometric ambiguities of single-view calibration, confirming the benefit of our joint optimization strategy.

\subsection{Effectiveness of the Dataset and the Framework}
\label{sec:supp_data_vs_framework}

\suppref{tbl:supp_test_results} compares \themethod against the state of the art on the test split of our proposed dataset.
We report error of roll, pitch and vFoV in degrees, the focal length as a relative error in percent of its ground-truth value, and the last two columns as the errors of the predicted up-vector and latitude fields.
GeoCalib$_\text{finetuned}$ is the public GeoCalib architecture fine-tuned on our training data, so its distance from the original GeoCalib row isolates what the data alone contributes: the mean roll error drops by $20\%$ and the up-vector field error by $59\%$, demonstrating the effectiveness of our proposed dataset. Even after that fine-tuning, \themethod remains ahead in every column, at $3.22^\circ$ against $4.98^\circ$ in roll and $0.04$ against $0.07$ in the up-vector field from a single image and at $2.62^\circ$ and $0.04$ once the views are given jointly, so with the training data equalised the remaining gap is attributable to the architecture. Our multi-view configuration (M$+$M) attains the lowest error in every column, and the only place a baseline stays ahead of our single-image fallback is the relative focal error, where GeoCalib$_\text{dist}$~\citep{veicht2024geocalib} reaches $11.95\%$ against our $12.05\%$ while using multi-view optimization against our single image.

\section{Single-view Comparison}
\label{sec:supp_singleview}

The main paper reports the single-view comparison on Stanford2D3D in
\cref{tbl:generalization_s2d3d}; \suppref{tbl:generalization} extends it to
TartanAir~\citep{wang2020tartanair} and MegaDepth~\citep{li2018megadepth} under
the same protocol~\citep{veicht2024geocalib}. Ours
attains the lowest mean roll and pitch error on both benchmarks, ahead of the
strongest dedicated baseline by roughly two times, and the lowest mean
field-of-view error on MegaDepth. The single exception is the mean field of view
on TartanAir, where VGGT~\citep{wang2025vggt} leads by $0.44^\circ$; on the tighter
measures the ranking reverses, ours reaching a median of $1.72^\circ$ against
$1.88^\circ$ and an AUC at $1^\circ$ of $32.5$ against $22.4$, so the two are
separated only in the tail. VGGT also recovers no gravity at all.
Our model is designed for the sparse multi-view regime, where it attains the
lowest mean error on all four benchmarks of \cref{tbl:multiview_public}, and
reduces to single-image inference when a single view is given. The results above
show that this capability does not come at the expense of single-view accuracy:
the same model remains on par with the single-view state of the art.

\begin{table*}[t]
\centering
\begin{threeparttable}
\small
\renewcommand{\arraystretch}{1.0}%
\setlength\tabcolsep{1.5pt}%
\begin{tabular}{clccccccccccccccc}
\toprule
&\multirow{2}{*}[-.4em]{Approach} 
& \multicolumn{5}{c}{Roll [degrees]} 
& \multicolumn{5}{c}{Pitch [degrees]} 
& \multicolumn{5}{c}{FoV [degrees]}\\
\cmidrule(lr){3-7}
\cmidrule(lr){8-12}
\cmidrule(lr){13-17}
&& mean\,$\downarrow$ & med.\,$\downarrow$ & \multicolumn{3}{c}{AUC\,$\triangleright$\,1/5/10\degree\,$\uparrow$}
& mean\,$\downarrow$ & med.\,$\downarrow$ & \multicolumn{3}{c}{AUC\,$\triangleright$\,1/5/10\degree\,$\uparrow$}
& mean\,$\downarrow$ & med.\,$\downarrow$ & \multicolumn{3}{c}{AUC\,$\triangleright$\,1/5/10\degree\,$\uparrow$} \\
\midrule
&DeepCalib\tnote{*}~\citeyearpar{lopez2019deepcalib}      & - &          1.95 &          24.7 &          55.4 &          71.5 & - &          3.27 &          16.3 &          38.8 &          58.5 & - &          8.07 &          \01.5 &          \08.8 &          27.2 \\
&Perceptual~\citeyearpar{perceptual}              & - &          2.24 &          23.2 &          48.6 &          66.7 & - &          2.86 &          23.5 &          44.6 &          61.5 & - &           8.01 &          \04.8 &          28.0 &          39.4 \\
&CTRL-C~\citeyearpar{lee2021ctrl}                       & - &          1.68 &          32.8 &          59.1 &          74.1 & - &          2.39 &   24.6 &   48.6 &          65.2 & - &          5.64 &    10.7 &    25.4 &          43.5 \\
&MSCC~\citeyearpar{Song2024MSCC}                  & - &          3.50 &          15.0 &          37.2 &          57.7 & - &          3.48 &          18.8 &          38.6 &          54.3 & - &           11.18 &          \04.4 &           11.8 &          23.0 \\
&ParamNet~\citeyearpar{jin2023perspective}  & 3.40 &          \02.33 &          23.3 &          51.4 &          71.0 & \05.95 &          \02.87 &          19.9 &          43.8 &          62.9 & \07.46 &          \06.04 &          \08.6 &           22.6 &          40.9 \\
&SVA~\citeyearpar{lochman2021minimal}                            & - &          9.48 &          32.4 &          39.6 &          44.1 & - &           18.46 &          21.2 &          28.8 &          34.5 & - &           43.01 &          \08.8 &           16.1 &          21.6 \\
&UVP~\citeyearpar{Pautrat2023uvp}    & 9.45 &          \cthird 0.86 &          52.7 &          64.6 &          71.4 & 10.51 &          2.43 &          35.6 &          48.9 &          58.8 & 20.70 &          8.94 &          17.4 &           27.5 &          36.8 \\
& GeoCalib\tnote{*}~\citeyearpar{veicht2024geocalib}& \cthird 1.59 & \cfirst \00.43 &          \cthird 71.4 &          \cthird 83.7 &          \cthird 89.7 &          \cthird 3.48 &          \cthird 1.49 &          \cthird 38.2 &          \cthird 63.0 &          \cthird 76.6 &          7.40 &          4.90 &          14.2 &           30.5 &          47.7 \\
& DUSt3R~\citeyearpar{wang2024dust3r} & - & - & - & - & - & - & - & - & - & - &  13.59 &  12.37 &  \01.4 &  \03.8 &  \09.3 \\
& VGGT~\citeyearpar{wang2025vggt}    & - & - & - & - & - & - & - & - & - & - &   \cfirst 2.12 &   \cthird 1.88 & \cthird 22.4 & \cfirst 60.9 & \cfirst 80.3 \\
& AnyCalib$_{p}$\tnote{$\dagger$}~\citeyearpar{tirado2025anycalib}      & - & - & - & - & - & - & - & - & - & - &   5.40 &   3.63 &           15.6 &          36.4 &          \cthird 55.1 \\
& AnyCalib$_{g}$\tnote{$\dagger$}~\citeyearpar{tirado2025anycalib} & - & - & - & - & - & - & - & - & - & - &   5.72 &   4.06 &          13.6 &          33.4 &          52.8 \\
& \textbf{Ours\tnote{*}} & \csecond 0.79 & \csecond 0.46 & \csecond 79.4 & \csecond 91.5 & \csecond 95.4 & \csecond 1.57 & \cfirst 0.81 & \cfirst 56.9 & \cfirst 78.8 & \cfirst 88.5 & \cthird 2.58 & \cfirst 1.72 & \cfirst 32.5 & \csecond 59.8 & \csecond 76.5 \\
\multirow{-14}{*}{\begin{sideways}\textbf{TartanAir}~\citeyearpar{wang2020tartanair}\end{sideways}} & \textbf{Ours} & \cfirst 0.76 & \csecond 0.46 & \cfirst 79.8 & \cfirst 91.6 & \cfirst 95.5 & \cfirst 1.50 & \csecond 0.89 & \csecond 54.4 & \csecond 78.0 & \csecond 88.0 & \csecond 2.56 & \csecond 1.77 & \csecond 30.4 & \cthird 59.2 & \csecond 76.5 \\
\midrule
&DeepCalib\tnote{*}~\citeyearpar{lopez2019deepcalib}      & - &          1.41 &          34.6 &          65.4 &          79.4 & - &          5.19 &          11.9 &          27.8 &          44.8 & - &           11.14 &          \05.6 &           12.1 &          22.9 \\
&Perceptual~\citeyearpar{perceptual}              & - &          1.07 &          47.9 &          72.4 &          83.2 & - &          3.49 &   19.8 &          39.1 &   54.2 & - &           \06.21 &          \08.8 &          22.6 &          39.8 \\
&CTRL-C~\citeyearpar{lee2021ctrl}                       & - &   0.88 &   54.5 &   75.0 &   84.2 & - &          4.80 &          16.6 &          33.2 &          46.5 & - &           18.65 &          \02.0 &          \05.8 &          12.8 \\
&MSCC~\citeyearpar{Song2024MSCC}                  & - &          0.90 &          53.1 &          72.8 &          82.1 & - &          5.73 &          19.0 &          33.2 &          44.3 & - &   10.80 &          \06.0 &           14.6 &   26.2 \\
&ParamNet~\citeyearpar{jin2023perspective}  & 2.35 &          \01.46 &          37.0 &          66.4 &          80.8 & \07.07 &          \03.53 &          15.8 &          37.3 &          57.1 & 12.04 &          11.11 &          \05.2 &           12.7 &          23.8 \\
&SVA~\citeyearpar{lochman2021minimal}                            & - &              - &          31.9 &          35.0 &          36.2 & - &              - &          13.6 &          20.6 &          24.9 & - &              - &          9.4 &    16.1 &          21.1 \\
&UVP~\citeyearpar{Pautrat2023uvp}    & 3.72 & \cthird 0.50 &          68.3 &          \cthird 81.2 &          86.4 & 12.61 &          \04.78 &          20.9 &          35.9 &          47.3 & 22.93 &          10.63 &          \08.3 &           18.9 &          30.2 \\
&GeoCalib\tnote{*}~\citeyearpar{veicht2024geocalib}                                  & \cthird 1.08 & \cfirst 0.36 & \cthird 82.4 & \csecond 90.6 & \cthird 94.0 & \cthird \04.50 & \cthird 1.96 & \cthird 31.8 &          \cthird 53.1 & \cthird 67.4 & \07.75 & 4.45 & 13.9 &           31.5 &          48.0 \\
\rowcolor{gray!15} \cellcolor{white} & DUSt3R~\citeyearpar{wang2024dust3r} & - & - & - & - & - & - & - & - & - & - &   3.35 &   1.84 &          31.6 &          56.5 &          72.1 \\
\rowcolor{gray!15} \cellcolor{white} & VGGT~\citeyearpar{wang2025vggt}    & - & - & - & - & - & - & - & - & - & - &   1.57 &   0.87 & 55.6 & 78.5 & 87.9 \\
& AnyCalib$_{p}$\tnote{$\dagger$}~\citeyearpar{tirado2025anycalib} & - & - & - & - & - & - & - & - & - & - &   \cthird 5.05 & \cthird 3.14 & \cfirst 19.4 & \csecond 40.8 & \cthird 59.1 \\
& AnyCalib$_{g}$\tnote{$\dagger$}~\citeyearpar{tirado2025anycalib} & - & - & - & - & - & - & - & - & - & - &   5.57 & 3.57 & \cthird 14.8 & 36.6 &          55.7 \\
& \textbf{Ours\tnote{*}} & \csecond 0.61 & \csecond 0.38 & \cfirst 84.5 & \cfirst 94.3 & \csecond 97.0 & \csecond 2.92 & \cfirst 1.77 & \csecond 34.4 & \csecond 57.8 & \csecond 73.8 & \cfirst 3.59 & \cfirst 2.74 & \csecond 17.4 & \cfirst 45.1 & \cfirst 66.8 \\
\multirow{-14}{*}{\begin{sideways}\textbf{MegaDepth}~\citeyearpar{li2018megadepth}\end{sideways}} & \textbf{Ours} & \cfirst 0.60 & \csecond 0.38 & \csecond 84.4 & \cfirst 94.3 & \cfirst 97.1 & \cfirst 2.81 & \csecond 1.79 & \cfirst 34.8 & \cfirst 58.5 & \cfirst 74.6 & \csecond 3.93 & \csecond 3.08 & 14.7 & \cthird 40.7 & \csecond 63.7 \\
\bottomrule
\end{tabular}%

\begin{tablenotes}
\item[*] Trained on the \texttt{OpenPano} dataset.
\item[$\dagger$] Trained on the \texttt{extended OpenPano} dataset.
\end{tablenotes}
\caption{
    \textbf{Single-view comparison on TartanAir and MegaDepth ($N=1$).}
    All methods calibrate each image independently, following the protocol of
    GeoCalib~\citep{veicht2024geocalib}.
    \textbf{Ours*} is trained on \texttt{OpenPano} alone and isolates the
    contribution of the architecture, while \textbf{Ours} uses our full training
    mixture.
    \textbf{Gray} backgrounds indicate models that may have been exposed to the
    test data during training; they are reported for completeness but excluded
    from the ranking.
}\label{tbl:generalization}
\end{threeparttable}
\end{table*}

\section{Ablation Study}
\label{sec:supp_ablation}
We conduct multiple ablation studies to validate our architectural choices, including the choice of the head architecture and the design of the dense prediction (DPT) head. Every variant is evaluated on TartanAir~\citep{wang2020tartanair} under the singleview evaluation, with $N{=}10$ views per sequence, so the comparison between rows isolates the architectural change alone. Overall results are summarized in \suppref{tbl:supp_ablation}.

\begin{table*}[t]
\centering
\small
\renewcommand{\arraystretch}{1.1}
\setlength\tabcolsep{4.5pt}

\begin{tabular}{ll c cc cc cc}
\toprule
\multirow{2}{*}{Ablation Setting} & \multirow{2}{*}{Variant} & \multirow{2}{*}{\#Params} & \multicolumn{2}{c}{Roll [$^\circ$]} & \multicolumn{2}{c}{Pitch [$^\circ$]} & \multicolumn{2}{c}{FoV [$^\circ$]} \\
\cmidrule(lr){4-5} \cmidrule(lr){6-7} \cmidrule(lr){8-9}
& & & error\,$\downarrow$ & AUC@5$^\circ$$\uparrow$ & error\,$\downarrow$ & AUC@5$^\circ$$\uparrow$ & error\,$\downarrow$ & AUC@5$^\circ$$\uparrow$ \\
\midrule
\rowcolor{gray!10} \multicolumn{9}{l}{\textit{Head Architecture}} \\
Head Type        & MLP                      & 0.6M  & 0.76               & 91.9               & 1.47               & 79.4               & \textbf{3.77}      & \textbf{46.5}      \\
                 & \textbf{DPT}             & 5.9M  & \textbf{0.73}               & \textbf{92.6}      & \textbf{1.34}      & \textbf{82.7}      & 4.30               & 43.0               \\
\midrule
\rowcolor{gray!10} \multicolumn{9}{l}{\textit{DPT Sampling Ratio ($down\_ratio$)}} \\
Sampling Ratio   & 1/1 ($dr=1$)             & 31.2M & 0.77               & 92.1               & 1.49               & 81.9               & 4.93               & 34.4               \\
                 & 1/2 ($dr=2$)             & 8.9M  & 0.74               & 92.1               & 1.70               & 75.2               & \textbf{4.08}               & \textbf{45.8}               \\
                 & \textbf{1/4} ($dr=4$)   & 5.9M  & 0.73               & \textbf{92.6}      & \textbf{1.34}      & \textbf{82.7}      & 4.30               & 43.0               \\
                 & 1/7 ($dr=7$)             & 1.9M  & \textbf{0.72}      & 92.2               & 1.70               & 75.7               & 4.16               & 45.1               \\
\midrule
\rowcolor{gray!10} \multicolumn{9}{l}{\textit{Transformer Layer Selection}} \\
Layer Indices    & [4, 11, 17, 23]          & 5.9M  & 0.82               & 91.0               & 1.64               & 78.5               & \textbf{3.88}               & \textbf{45.5}               \\
                 & \textbf{[15, 18, 21, 23]} & 5.9M  & \textbf{0.73}               & \textbf{92.6}      & \textbf{1.34}      & \textbf{82.7}      & 4.30               & 43.0               \\
\bottomrule
\end{tabular}
\caption{\textbf{Ablation Study on Architectural Components.} We evaluate the impact of different head architectures, DPT sampling ratios, and backbone layer selections. All metrics (Roll, Pitch, and FoV) are reported as mean errors (denoted as error) and Area Under the Curve (AUC) at 5$^\circ$ threshold on TartanAir. Params denotes the number of parameters in the head. Bold indicates the default configuration.}
\label{tbl:supp_ablation}
\end{table*}

To verify the choice of the Transformer-based DPT architecture, we compare it against a variant in which the DPT is replaced by an MLP head.
The MLP head performs feature upsampling using linear layers and pixel shuffling~\citep{Shi2016pixelshuffle}. 
From \suppref{tbl:supp_ablation}, we can see that the DPT head outperforms the MLP baseline in terms of gravity estimation. This performance gap highlights the effectiveness of the DPT's multi-scale feature fusion for high-fidelity geometric field prediction.

The DPT head is responsible for generating high-resolution perspective fields from the Transformer's latent representations. We investigate the impact of the downsampling ratio, expressed relative to the full training resolution, together with the feature dimension that goes with it. We compare four configurations:
\begin{itemize}
    \item 1/1 Sampling: A high-capacity setting with down\_ratio=1 and features=256, directly predicting fields at the patch resolution.
    \item 1/2 Sampling: A medium-capacity setting with down\_ratio=2 and features=128.
    \item 1/4 Sampling: A compact version with down\_ratio=4 and features=64.
    \item 1/7 Sampling: A lightweight setting with down\_ratio=7 and features=32.
\end{itemize}
Our results show that a higher-resolution head does not pay off: at 31.2M parameters the 1/1 setting is by far the largest of the four, yet it attains the worst roll and the worst vFoV error among them. The remaining ratios are close in accuracy, with 1/4 clearly ahead on pitch, so we adopt it as the trade-off between geometric precision and computational efficiency.

Furthermore, we evaluate the selection of intermediate layers from the 24-layer Transformer backbone used for DPT feature fusion. 
We test two configurations: an evenly distributed selection (layers [4, 11, 17, 23]) and a selection concentrated in the deeper blocks (layers [15, 18, 21, 23]).
As shown in \suppref{tbl:supp_ablation}, the deeper combination is the better choice for gravity, lowering the roll error from $0.82^\circ$ to $0.73^\circ$ and the pitch error from $1.64^\circ$ to $1.34^\circ$, at the cost of a slightly higher vFoV error. This suggests that the orientation cues camera calibration relies on are carried by the high-level geometric consistency and global structure of the deep layers rather than by the local texture patterns typically found in earlier ones.

\section{Additional Qualitative Results}
\label{sec:supp_qualitative}
Due to space constraints in the main paper, we present more visualization results to demonstrate the performance of our proposed method.

\subsection{Visualizations on Multi-view Benchmarks}
\label{sec:supp_qual_multiview}
\begin{figure}[t]
    \centering
    \includegraphics[width=\linewidth]{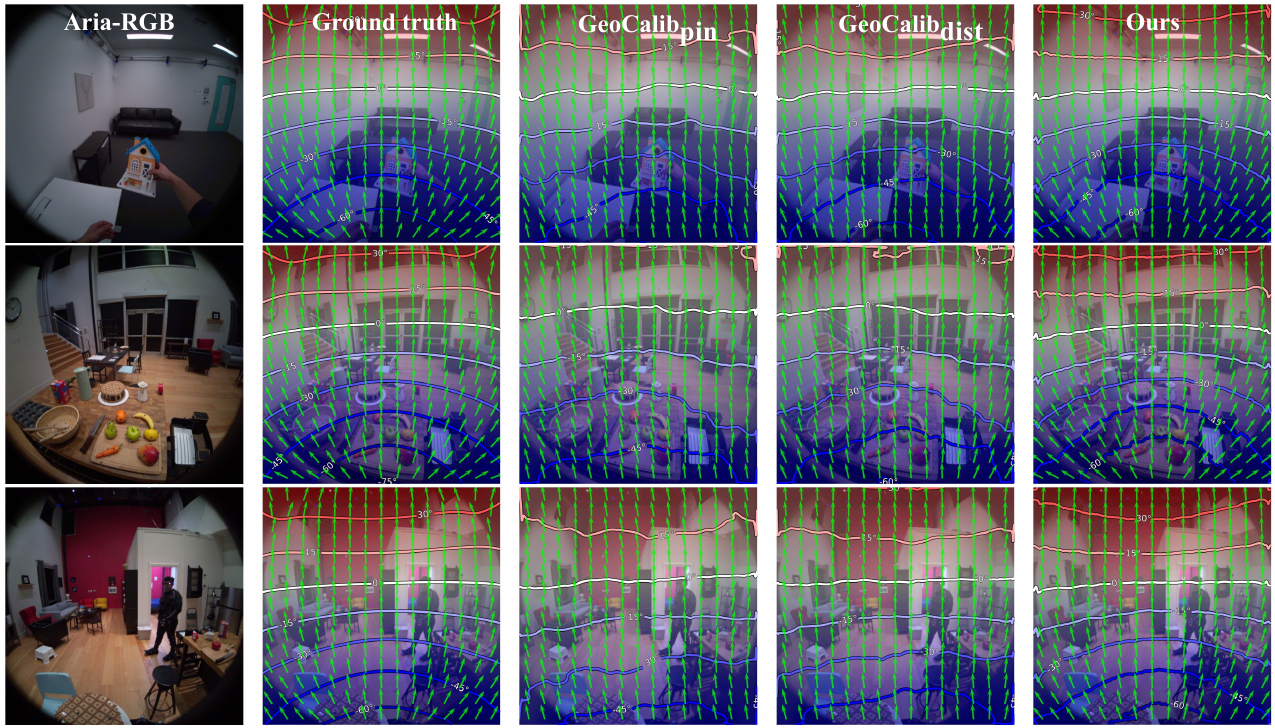}
    \caption{\textbf{Qualitative perspective field results on Aria-ADT-RGB.} Three sequences from the RGB stream of Project Aria. Columns, from left to right: the input image, the ground-truth perspective field, GeoCalib$_\text{pin}$, GeoCalib$_\text{dist}$, and ours.}
    \label{fig:supp_aria_rgb}
\end{figure}

\begin{figure}[t]
    \centering
    \includegraphics[width=\linewidth]{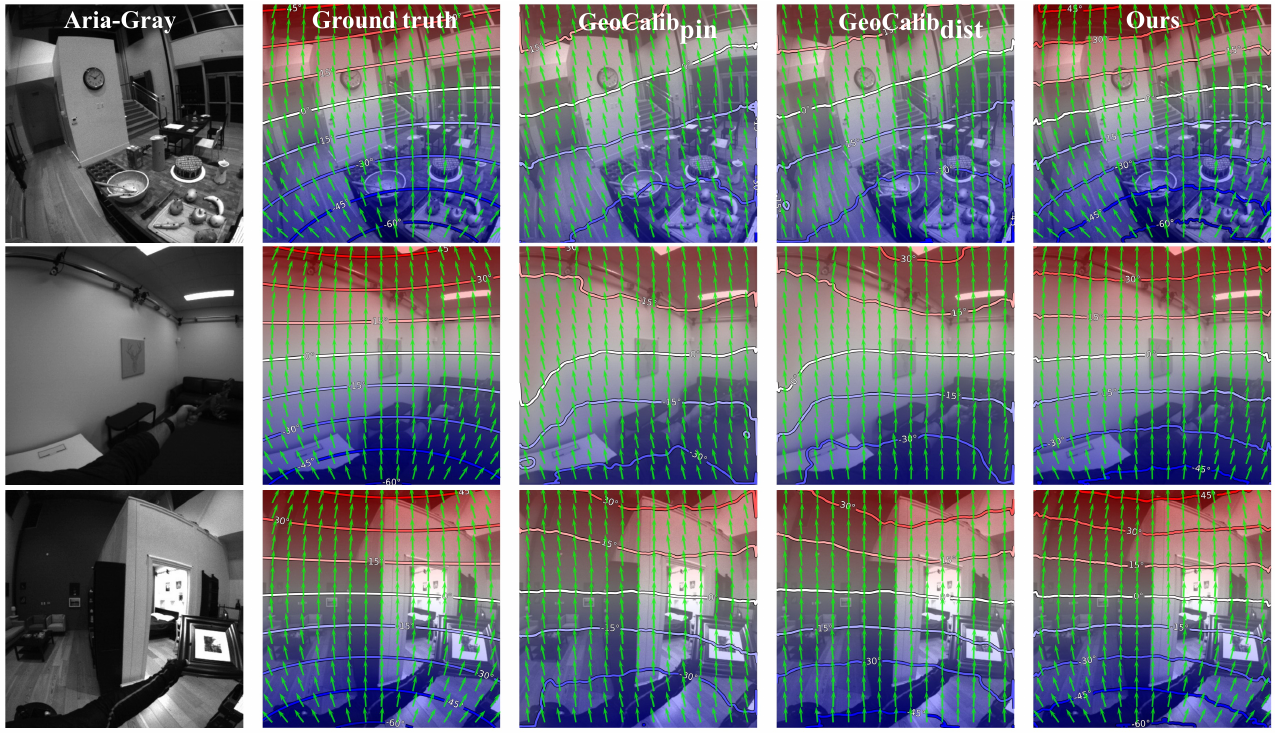}
    \caption{\textbf{Qualitative perspective field results on Aria-ADT-Gray.} Three sequences from the grayscale side cameras of Project Aria, whose wider field of view makes the latitude contours curve more sharply than in the RGB stream. Columns as in \suppref{fig:supp_aria_rgb}.}
    \label{fig:supp_aria_gray}
\end{figure}

\begin{figure}[t]
    \centering
    \includegraphics[width=\linewidth]{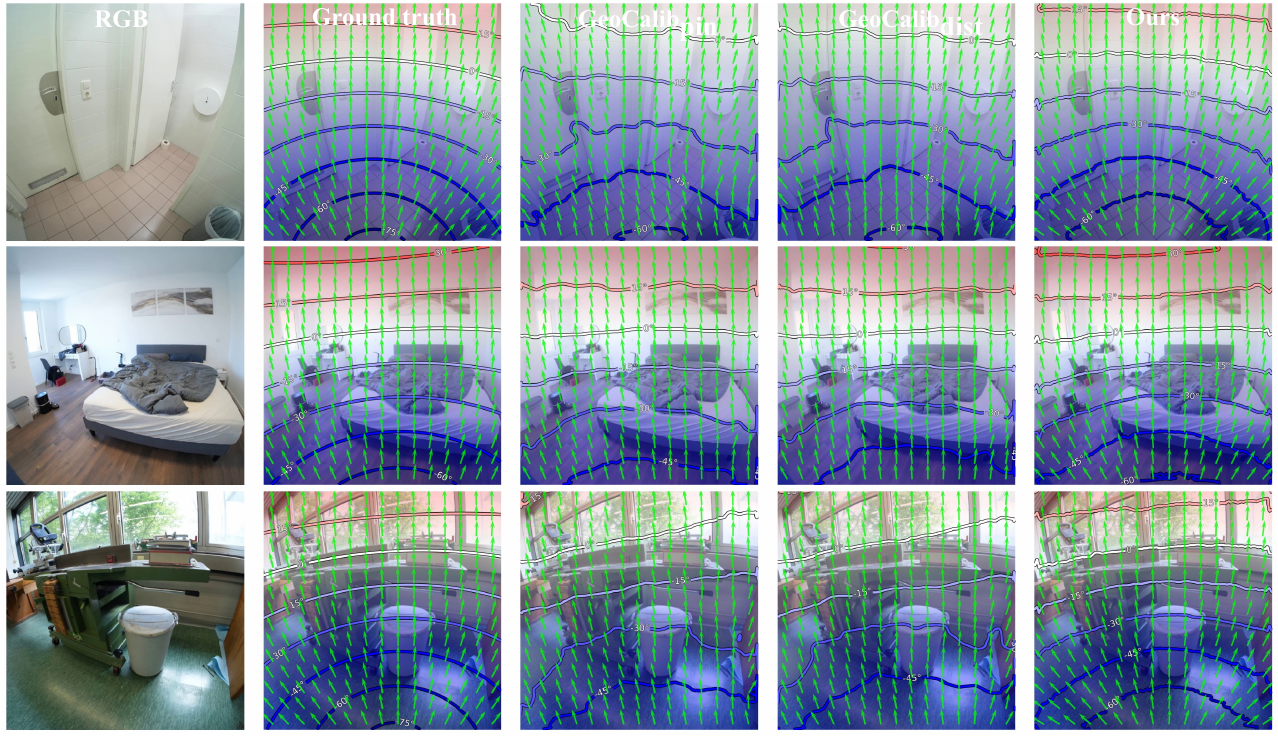}
    \caption{\textbf{Qualitative perspective field results on ScanNet++.} Three indoor scenes captured with the DSLR fisheye of ScanNet++. Columns as in \suppref{fig:supp_aria_rgb}.}
    \label{fig:supp_scannetpp}
\end{figure}

\suppref{fig:supp_aria_rgb,fig:supp_aria_gray,fig:supp_scannetpp} show the three benchmarks of \cref{tbl:multiview_public} that are captured by physical fisheye cameras, completing the qualitative coverage begun in \cref{fig:multiview_public_qual} of the main paper. In every figure the columns are, from left to right, the input image, the ground-truth perspective field, GeoCalib$_\text{pin}$, GeoCalib$_\text{dist}$~\citep{veicht2024geocalib}, and our \themethod; green arrows denote the up-vector field, coloured contours the latitude field, and the white curve the horizon.

Across all three benchmarks the two GeoCalib variants flatten the latitude contours and push the horizon away from its true position, which is the signature of fitting a nearly pinhole geometry to a strongly curved lens. Our predictions instead reproduce the curvature of the ground-truth contours, and the effect is most visible towards the image periphery, where these lenses bend the rays hardest. None of these cameras or scenes is seen during training, so the agreement is obtained zero-shot.

\subsection{Visualizations on Single-view Benchmarks}

\begin{figure}[t]
    \centering
    \includegraphics[width=\linewidth]{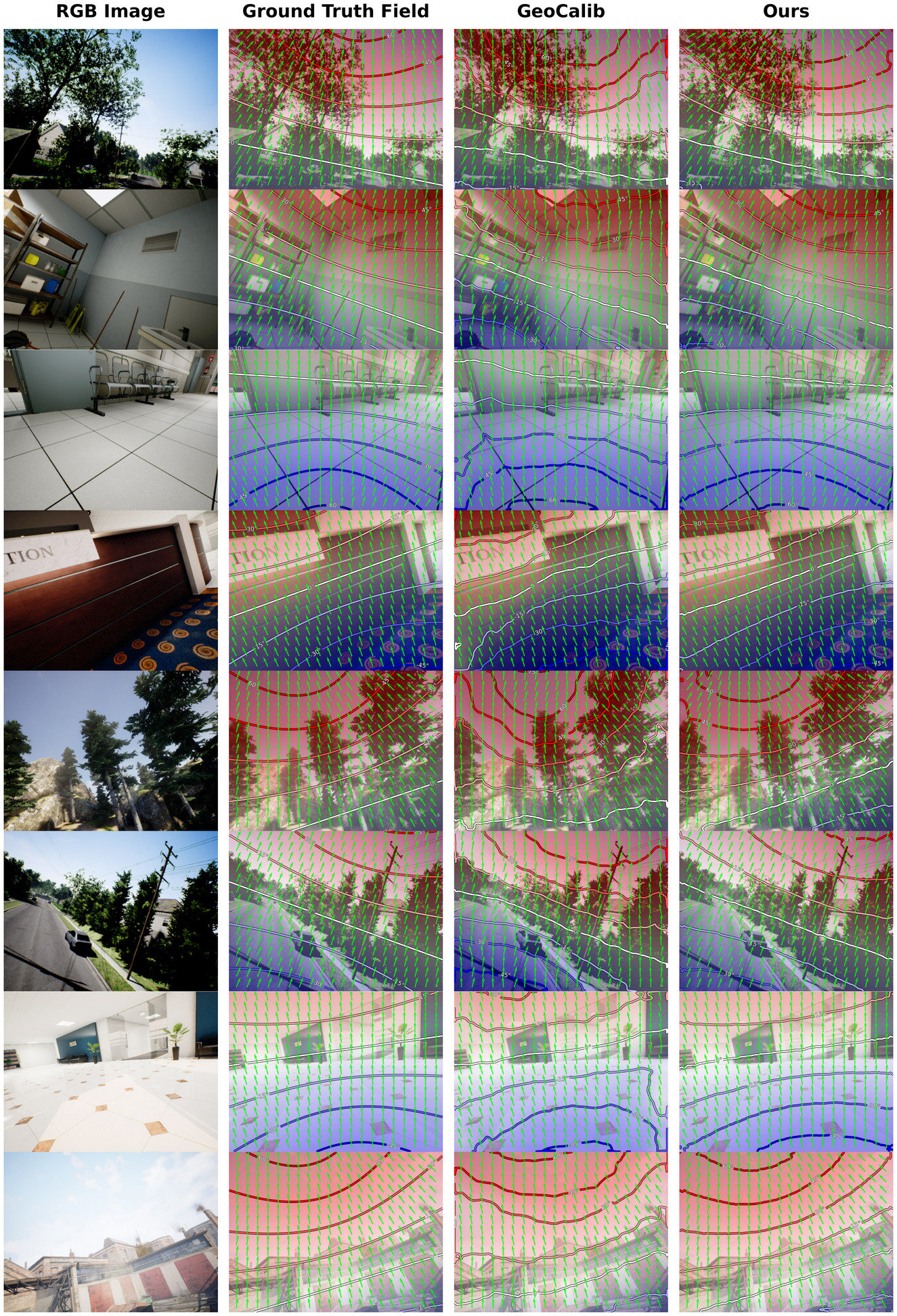}
    \caption{\textbf{Qualitative results on the TartanAir dataset.} We present a comparison of single-view calibration on eight randomly sampled images from the TartanAir benchmark. The columns show (from left to right): the input RGB image, the ground-truth perspective fields, visualizations of GeoCalib~\citep{veicht2024geocalib}, and our \themethod. It is clearly visible that our method generates perspective fields that are more consistent with the ground truth.}
    \label{fig:supp_tartanair_qualitative}
\end{figure}

\begin{figure}[t]
    \centering
    \includegraphics[width=\linewidth]{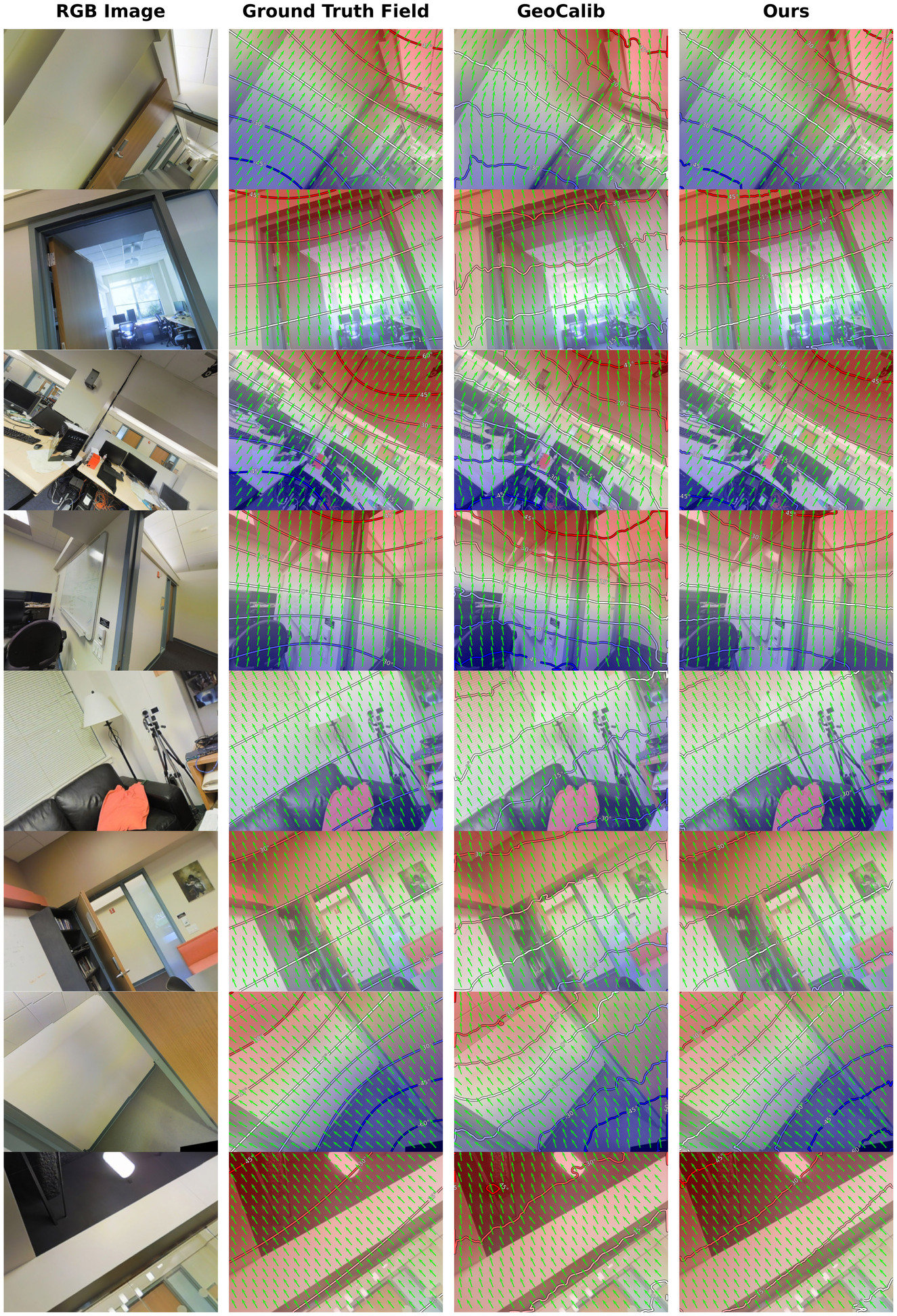}
    \caption{\textbf{Qualitative results on the Stanford2D3D dataset.} We visualize the single-view calibration performance on eight random samples from the Stanford2D3D benchmark. The columns represent (from left to right): the input RGB image, the ground-truth perspective fields, the results predicted by GeoCalib~\citep{veicht2024geocalib}, and our \themethod. Our framework exhibits superior capability in recovering accurate perspective fields across diverse indoor environments, closely aligning with the ground truth.}
    \label{fig:supp_stanford2d3d_qualitative}
\end{figure}

\begin{figure}[t]
    \centering
    \includegraphics[width=\linewidth]{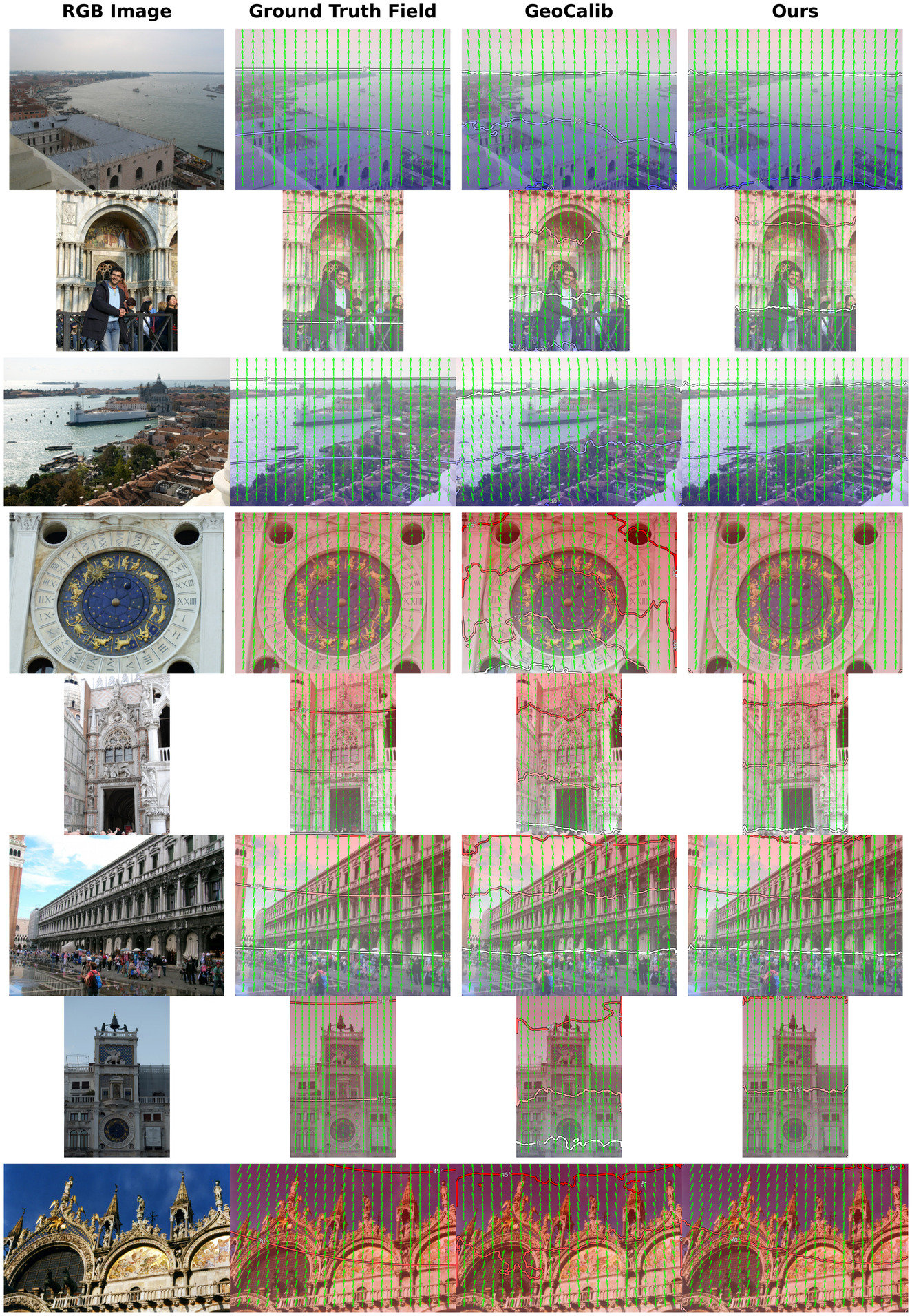}
    \caption{\textbf{Qualitative results on the MegaDepth dataset.} We compare our single-view calibration results against GeoCalib~\citep{veicht2024geocalib} on eight images from the MegaDepth dataset, featuring diverse outdoor scenes. The columns show (left to right): input RGB image, ground-truth perspective fields, GeoCalib's prediction, and our \themethod. Our approach demonstrates high robustness in handling complex global structures and varying depths.}
    \label{fig:supp_megadepth_qualitative}
\end{figure}

The quantitative comparison of \cref{tbl:generalization_s2d3d} and \suppref{tbl:generalization} is complemented here by qualitative results on the same three benchmarks. As shown in \suppref{fig:supp_tartanair_qualitative,fig:supp_stanford2d3d_qualitative,fig:supp_megadepth_qualitative}, our \themethod generates perspective fields that are significantly more consistent with the ground-truth geometric structures compared to the baselines. 

Specifically, our method produces much more accurate latitude field estimations, leading to more precise alignment with the scene's horizontal structures. Furthermore, the up-vector predictions exhibit superior orientation accuracy, which is particularly noticeable in challenging cases such as the 2nd, 5th, and 6th rows of \suppref{fig:supp_tartanair_qualitative} and the 1st and 3rd rows of \suppref{fig:supp_stanford2d3d_qualitative}. These visual comparisons reinforce the robust geometric reasoning and high-fidelity estimation capabilities of our framework across indoor, synthetic, and diverse outdoor environments.

\subsection{Visualizations on Proposed Dataset}

\begin{figure*}[t]
    \centering
    \includegraphics[width=0.88\linewidth]{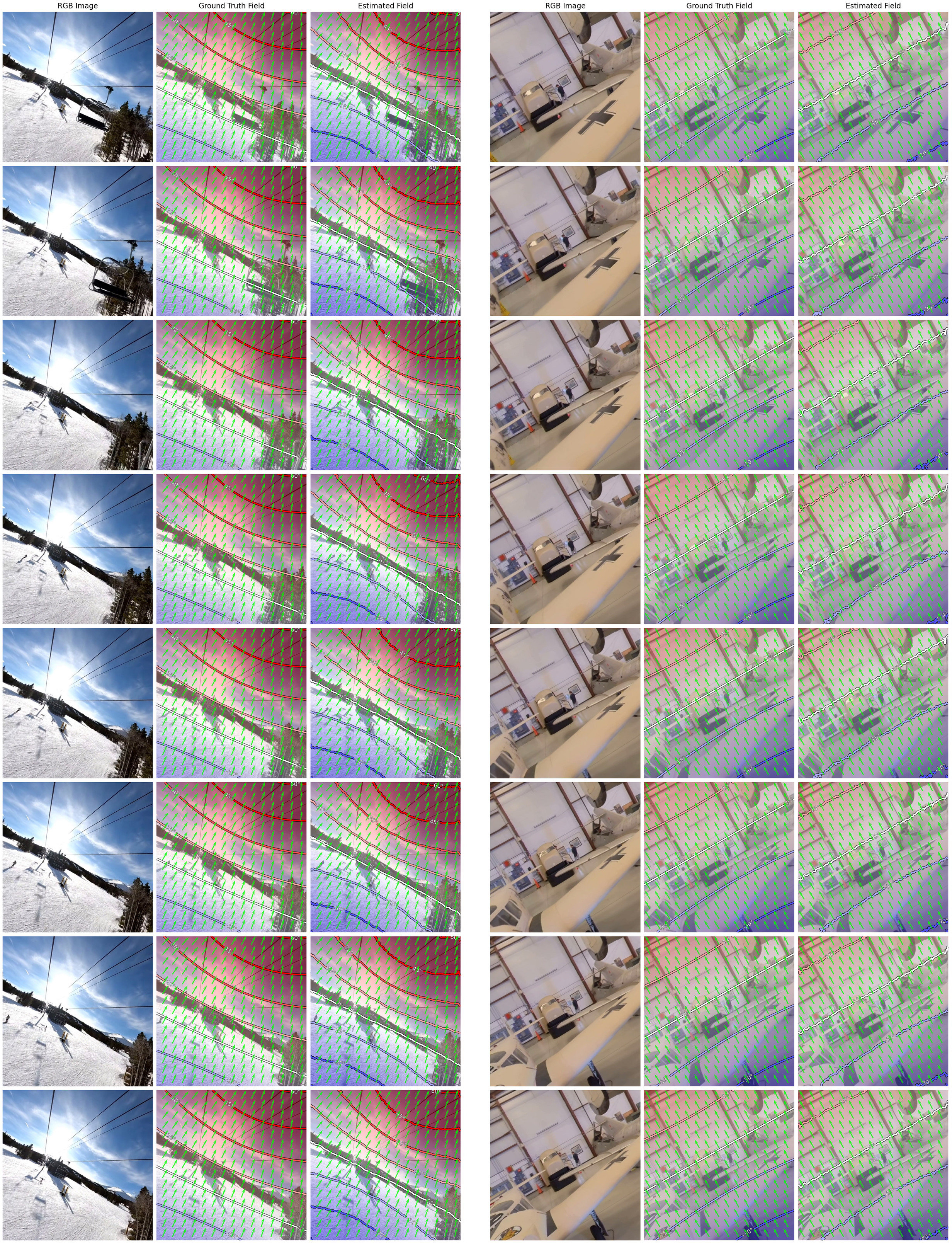}
    \caption{\textbf{Qualitative results on the test set of our proposed dataset.} Two sequences are shown side by side, eight frames each; within each half the columns are input RGB, ground-truth perspective fields, and the fields estimated by our \themethod. Left: Simple Radial, $72.0^\circ$ vFoV. Right: Simple Radial, $51.4^\circ$ vFoV.}
    \label{fig:supp_ours_test_pair1}
\end{figure*}

\begin{figure*}[t]
    \centering
    \includegraphics[width=0.88\linewidth]{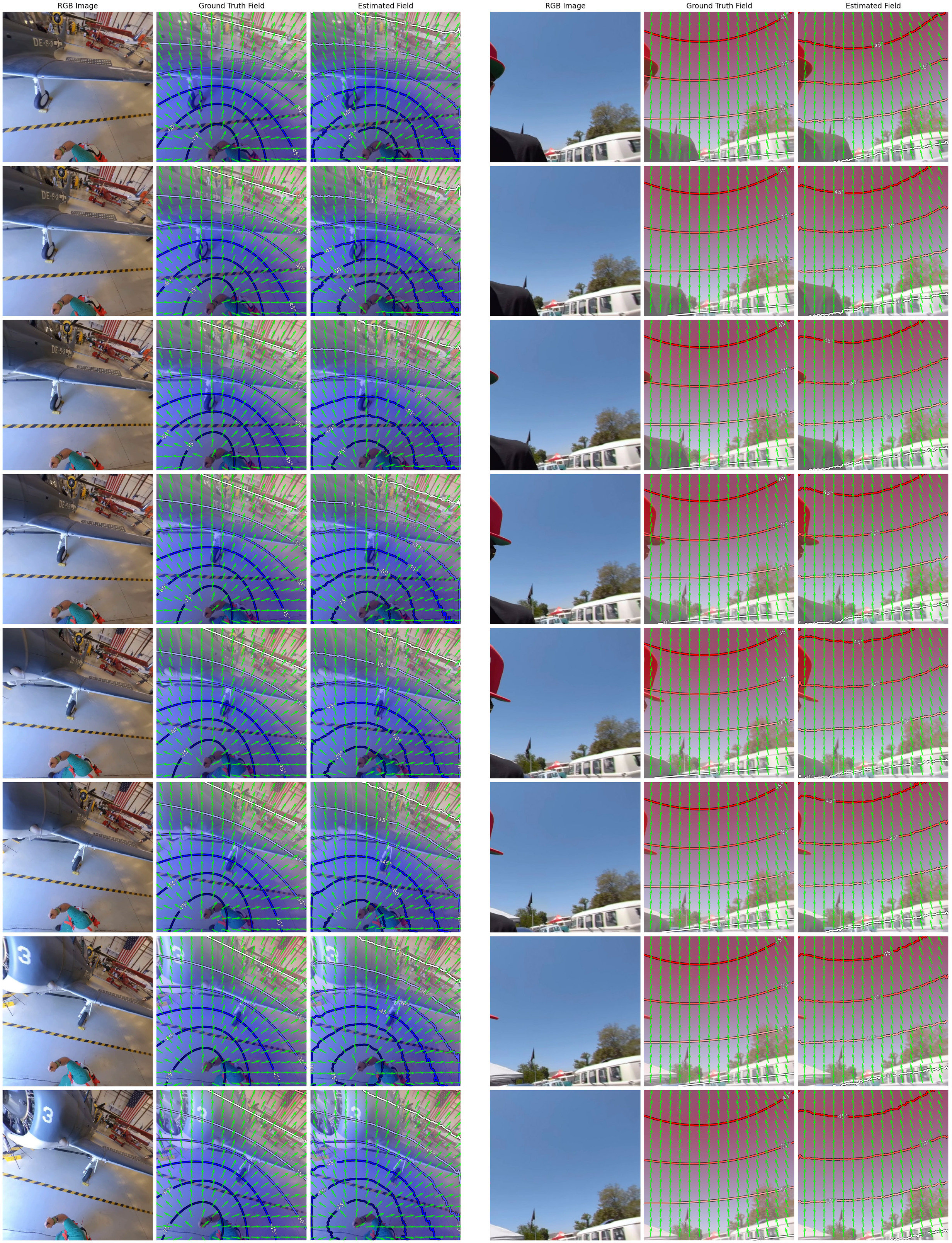}
    \caption{\textbf{Qualitative results on the test set of our proposed dataset.} Two sequences are shown side by side, eight frames each; within each half the columns are input RGB, ground-truth perspective fields, and the fields estimated by our \themethod. Left: Pinhole, $98.8^\circ$ vFoV. Right: Simple Radial, $56.2^\circ$ vFoV.}
    \label{fig:supp_ours_test_pair2}
\end{figure*}

\begin{figure*}[t]
    \centering
    \includegraphics[width=0.88\linewidth]{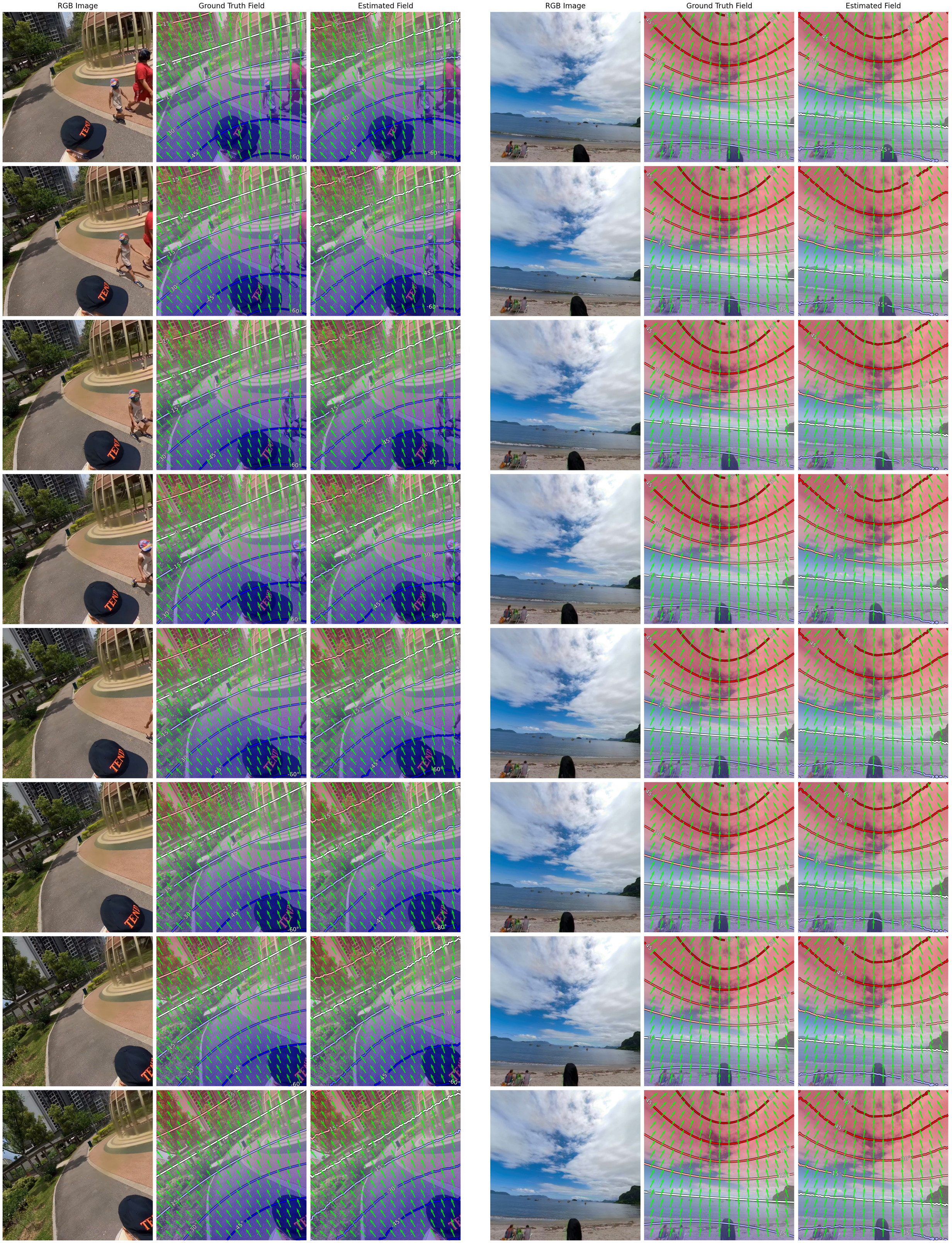}
    \caption{\textbf{Qualitative results on the test set of our proposed dataset.} Two sequences are shown side by side, eight frames each; within each half the columns are input RGB, ground-truth perspective fields, and the fields estimated by our \themethod. Left: Pinhole, $75.3^\circ$ vFoV. Right: Simple Radial, $98.0^\circ$ vFoV.}
    \label{fig:supp_ours_test_pair3}
\end{figure*}

\begin{figure*}[t]
    \centering
    \includegraphics[width=0.88\linewidth]{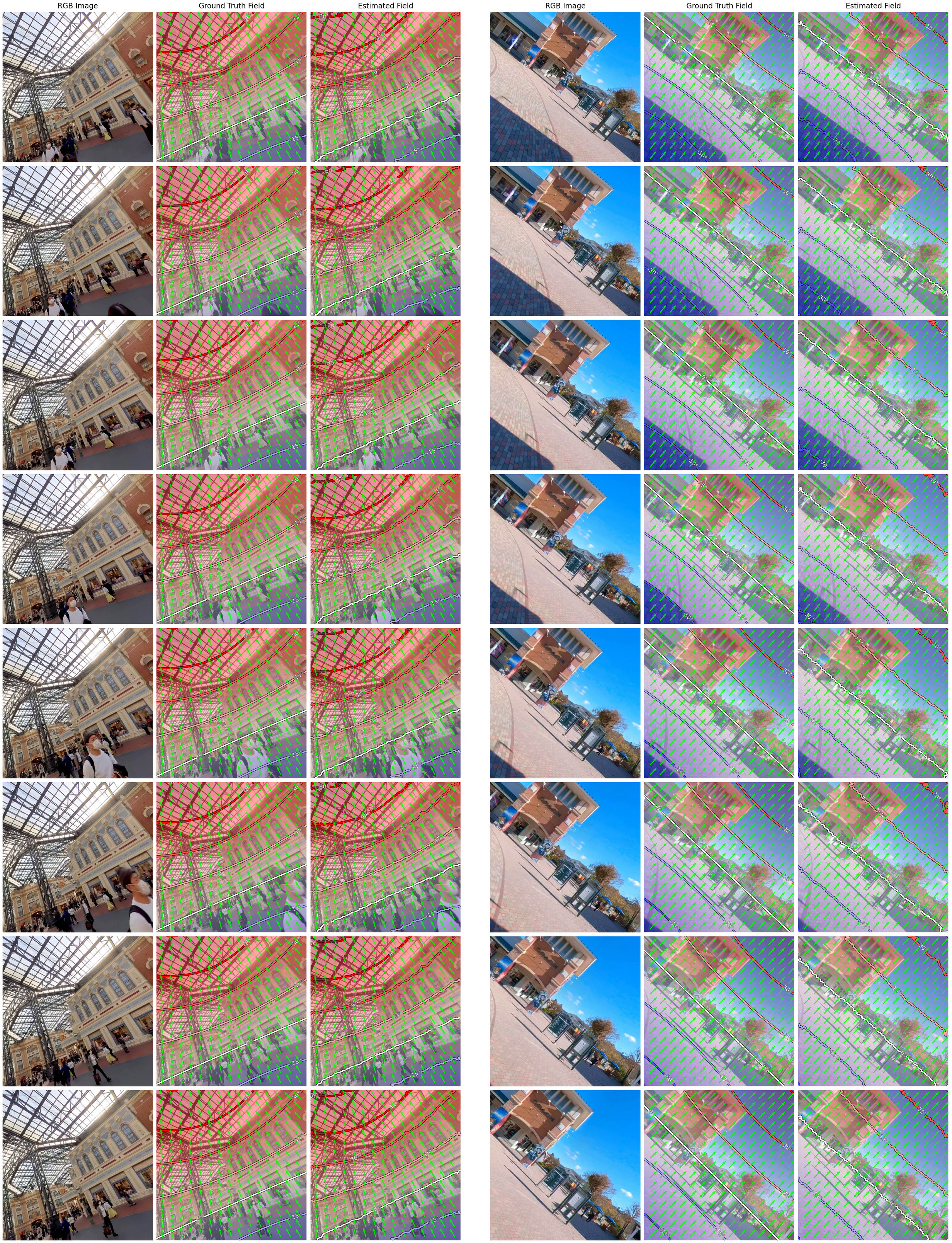}
    \caption{\textbf{Qualitative results on the test set of our proposed dataset.} Two sequences are shown side by side, eight frames each; within each half the columns are input RGB, ground-truth perspective fields, and the fields estimated by our \themethod. Left: Simple Radial, $73.7^\circ$ vFoV. Right: Simple Radial, $62.4^\circ$ vFoV.}
    \label{fig:supp_ours_test_pair4}
\end{figure*}

\begin{figure*}[t]
    \centering
    \includegraphics[width=0.88\linewidth]{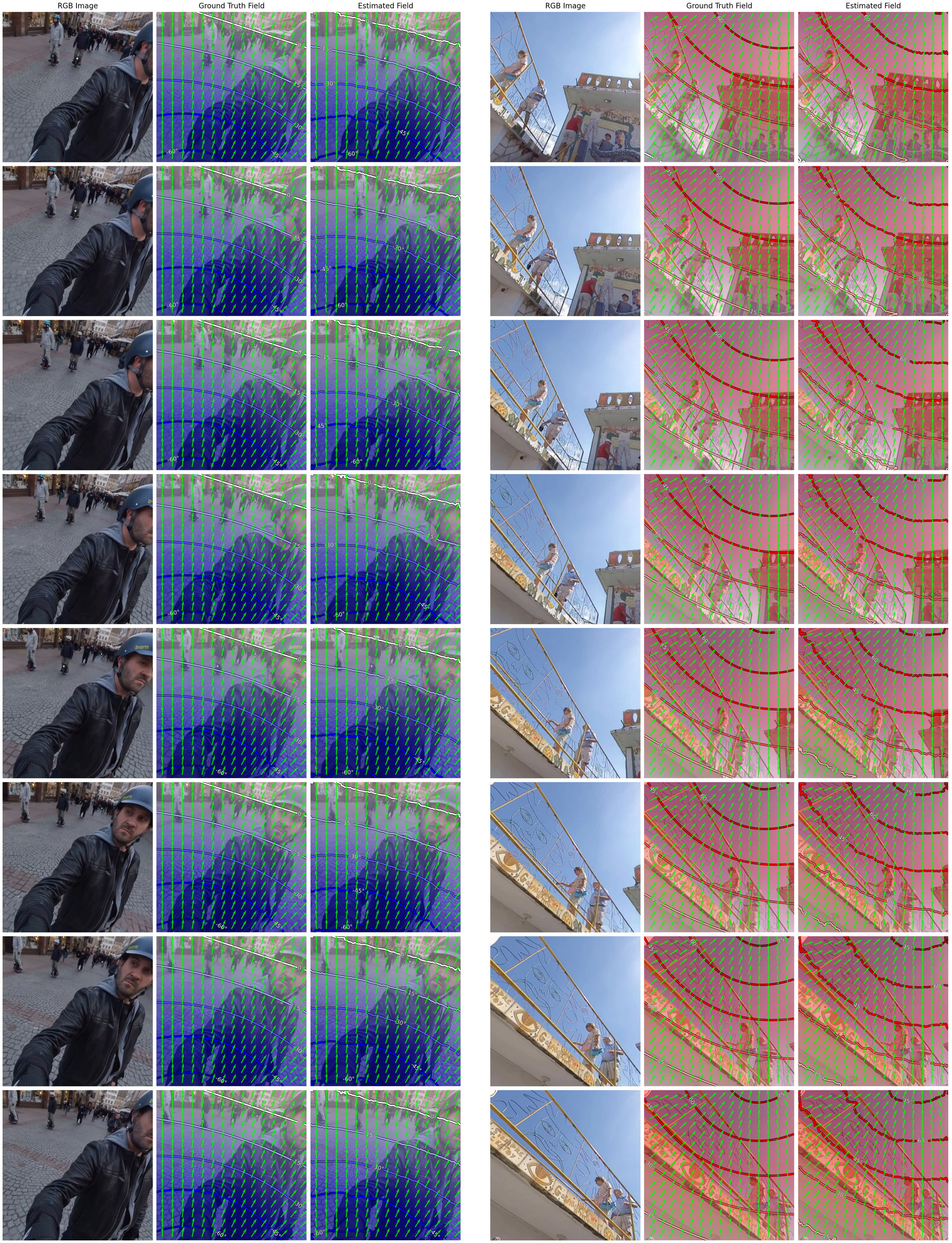}
    \caption{\textbf{Qualitative results on the test set of our proposed dataset.} Two sequences are shown side by side, eight frames each; within each half the columns are input RGB, ground-truth perspective fields, and the fields estimated by our \themethod. Left: Pinhole, $67.3^\circ$ vFoV. Right: Pinhole, $62.1^\circ$ vFoV.}
    \label{fig:supp_ours_test_pair5}
\end{figure*}

\begin{figure*}[t]
    \centering
    \includegraphics[width=0.88\linewidth]{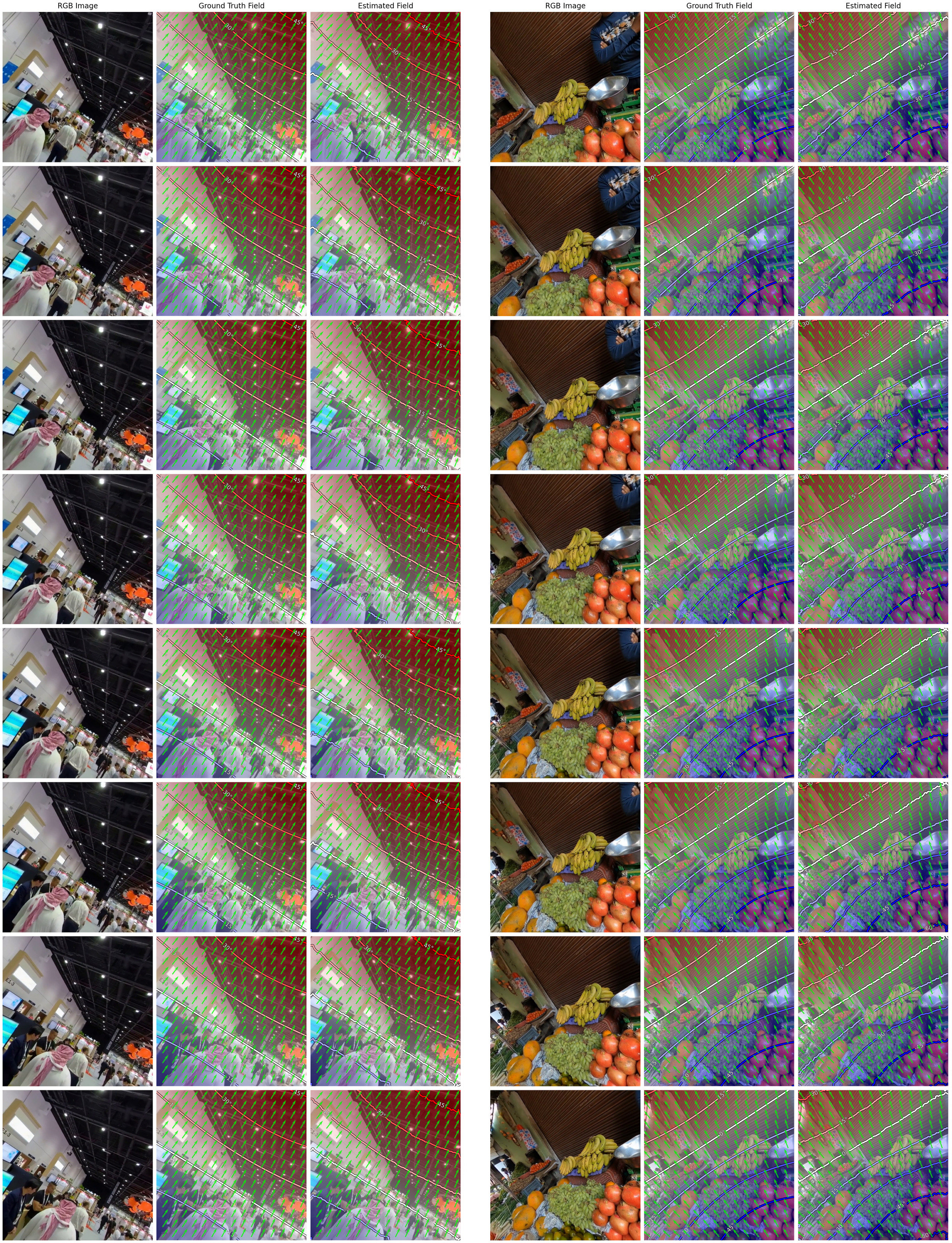}
    \caption{\textbf{Qualitative results on the test set of our proposed dataset.} Two sequences are shown side by side, eight frames each; within each half the columns are input RGB, ground-truth perspective fields, and the fields estimated by our \themethod. Left: Simple Radial, $61.8^\circ$ vFoV. Right: Simple Radial, $76.7^\circ$ vFoV.}
    \label{fig:supp_ours_test_pair6}
\end{figure*}
To further demonstrate the robustness of \themethod, we present additional qualitative results on the test set of our proposed calibration dataset. 
Our dataset is constructed from a vast collection of ``in-the-wild'' video clips, which introduce significant real-world challenges for camera calibration. These include scenes with dense dynamic pedestrians (\eg, \suppref{fig:supp_ours_test_pair1,fig:supp_ours_test_pair2,fig:supp_ours_test_pair3,fig:supp_ours_test_pair4}), texture-less surfaces or large expanses of sky and ground (\eg, \suppref{fig:supp_ours_test_pair5,fig:supp_ours_test_pair6}), and scenarios with ambiguous geometric orientations that remain potentially confusing even for human observers (\eg, \suppref{fig:supp_ours_test_pair5}).
These samples accurately reflect the complexities encounterable in daily life while presenting formidable obstacles for accurate geometric estimation. Despite these challenges, our method successfully produces high-fidelity perspective fields that are closely aligned with the ground truth, as visualized across these diverse examples. For comprehensive quantitative validation on this same test split, please refer to \suppref{sec:supp_multiview}.

\section{Dataset Construction Details}
\label{sec:supp_dataset}
This section details the two subsets of \cref{sec:dataset} in the same order, together with the virtual-camera stage they share.

\subsection{Rotation-Only Sequences from Static Panoramas}
The panorama pool is the gravity-aligned OpenPano collection of GeoCalib~\citep{veicht2024geocalib}: 2,616 / 147 / 148 panoramas for training, validation, and testing.
The rotation pool holds 619 trajectories of 81 frames, recovered by UCPE~\citep{zhang2026unified} with ViPE~\citep{huang2025vipe} from CameraBench videos~\citep{lintowards}, of which we draw five uniformly at random per panorama.
Each trajectory lives in the arbitrary world frame of the run that produced it, so we compute its mean rotation by singular value decomposition and left-multiply the sequence by the transpose of that mean.
This places the trajectory around the identity of the gravity-aligned panorama frame and leaves the relative inter-frame rotations---the actual camera motion---untouched.
Two augmentations then diversify each rectified sequence, with ranges following the extrinsic augmentation of AnyCalib~\citep{tirado2025anycalib}: both add a constant orientation offset drawn uniformly up to $180^\circ$ in yaw and $45^\circ$ in pitch and roll, and the second additionally ramps linearly towards a second random rotation from the same ranges.
The result is ten clips per panorama, i.e. 26,160 / 1,470 / 1,480 clips, whose ground-truth pose is a pure rotation with zero translation at every frame.

\subsection{Multi-Camera Synthesis and EUCM Parameterization}
Both subsets pass through the same virtual-camera stage, so their intrinsic distributions are identical and only the source of the trajectories differs.
Every augmented trajectory receives an independently sampled camera, following the synthesis protocol of AnyCalib~\citep{tirado2025anycalib}: Pinhole, Simple Radial, and EUCM, each with probability $1/3$.

\paragraph{Camera Sampling}
Pinhole and Simple Radial draw a vertical field of view $vFoV \sim \mathcal{U}[20^\circ, 105^\circ]$; the radial coefficient of the latter is parameterized as $k_1 = \hat{k}_1 \cdot f / H$ following GeoCalib~\citep{veicht2024geocalib}, with $\hat{k}_1$ from a normal distribution of standard deviation $0.07$ truncated at $\pm 0.3$.
EUCM draws $vFoV \sim \mathcal{U}[50^\circ, 180^\circ]$, $\alpha \sim \mathcal{U}[0.5, 0.8]$ and $\beta \sim \mathcal{U}[0.5, 2.0]$, spanning wide-angle optics through the ultra-wide fisheye lenses of action cameras.
Not every sampled triplet admits a valid projection over the full image, so we clamp the focal length to the range in which the model stays invertible across the sensor and recompute the realized $vFoV$, which is stored as the ground-truth annotation.
Each panorama is finally downscaled according to the sampled $vFoV$ before resampling, keeping the rendered $640 \times 640$ frames free of aliasing.

\subsection{The In-the-Wild Subset and Its Quality Control}
This subset starts from 1.1K in-the-wild panoramic source videos, from which UCPE~\citep{zhang2026unified} yields 2,479 clips, each matched by translation similarity to trajectories of the pool above and thereby anchored to a gravity-aligned world reference.
Keeping up to five matches per clip and applying the same augmentations and camera sampling gives ten synthesized clips per panoramic clip, 24,360 in total.

Unlike the curated OpenPano panoramas, in-the-wild footage carries visual defects that survive reprojection, so this subset is filtered with Qwen3-VL-4B~\citep{bai2025qwen3} using the prompt in \suppref{fig:filter}.
Since the synthesized frames deliberately exhibit strong barrel and fisheye curvature, the prompt states that lens distortion is an intended property of the virtual camera and must never by itself trigger a flag.
Given the middle frame of a clip, the model reports five binary flags: text overlays or watermarks, black voids from incomplete panoramas, artificial UI overlays, stitching seams or a large-area view of the camera carrier, and low visual quality.
Any of the first four rejects the clip; low quality is recorded but excluded from the rule, as blur or noise in the source footage does not corrupt the projective geometry that supervises calibration.
We likewise keep CGI and game-engine content, whose projection geometry remains valid for training.
Rejection is applied per clip rather than per source, so one flawed viewing direction does not discard the remaining usable views of a panorama.
This leaves 20,124 of the 24,360 clips (82.6\%).

\begin{figure*}[t]
\centering
\begin{PromptBox}
You are a video understanding assistant analyzing frames from a synthetic camera dataset.
Each frame was rendered from a real-world 360 video through a virtual camera that may be
a normal perspective camera, a camera with radial distortion, or a wide-angle fisheye
camera (field of view ranges from 20 up to ~180 degrees).

**IMPORTANT - lens distortion is intentional**: strong barrel / fisheye curvature (bent
straight lines, stretched or compressed periphery, circular appearance) is an expected
property of the virtual camera. It is NOT an image defect and NOT synthetic content.
Never flag a frame only because of lens distortion.

Given one such frame, your tasks are:

1. **Filtering**: Identify if the video should be filtered out.
Output boolean flags for the following conditions (true if the issue exists, false otherwise):

- has_subtitle_or_watermark: The frame contains **text overlays, subtitles, logos, or watermarks**. They may appear warped or bent by the lens distortion, and may sit anywhere in the frame (source-video watermarks near the 360 poles often end up mid-frame after reprojection). If such elements are present and not part of the real scene, set this to true.

- black_void: The frame shows large unnatural solid black areas or missing pixels. This typically happens when the source 360 video had raw black borders or missing regions that bleed into the rendered view. If substantial black void patches are visible anywhere in the frame, set this to true.

- has_overlay: The frame contains **artificial overlays**, such as embedded UI elements, pop-up graphics, stickers, video-in-video inserts, menus, or other synthetic elements that are not part of the natural scene. If you see signs of AR/VR interface, streaming UI, or added images, set this to true.

- stitching_or_photographer: The frame shows **stitching artifacts or a large-area view of whoever/whatever carries the camera**. Set this to true ONLY for: (a) visible stitching seams, ghosting, or duplicated/warped bands between stitched views; or (b) the camera carrier occupying a **large portion of the frame** (roughly 15% or more) - e.g. the photographer's head/torso, a motorcycle rider's body and handlebars, a drone body with propellers dominating part of the view. Do NOT flag small or peripheral capture equipment: a tripod leg, selfie-stick, small rig fragment, small circular nadir patch, or a small hand at the frame edge - these are acceptable.

- low_quality: The frame is of **poor visual quality**: blurry, noisy, heavily pixelated, over/under-exposed, or so low-resolution that the scene content is hard to recognize. Judge only texture/sharpness/exposure - do NOT count lens distortion (bent lines, fisheye curvature) as low quality.

2. **POI Categorization**: From the provided list of categories, select **one or more most relevant** labels that best describe the scene. Only use the given categories, copied **exactly** as written below (e.g. "Forest-Mountains", not "Forests") - do not invent new ones.

**Output strictly in JSON format** as follows:

```json
{
  "filter": {
    "has_subtitle_or_watermark": false,
    "black_void": false,
    "has_overlay": false,
    "stitching_or_photographer": false,
    "low_quality": false
  },
  "poi_category": ["Mountains"]
}
```
\end{PromptBox}
\caption{
    Prompt used for VLM-based clip filtering. The enumeration of POI categories is omitted here for brevity.
}\label{fig:filter}
\end{figure*}

\subsection{Dataset Samples}
\begin{figure}[ht!]
    \centering
    \begin{tabular}{c}
        \includegraphics[width=1.0\linewidth]{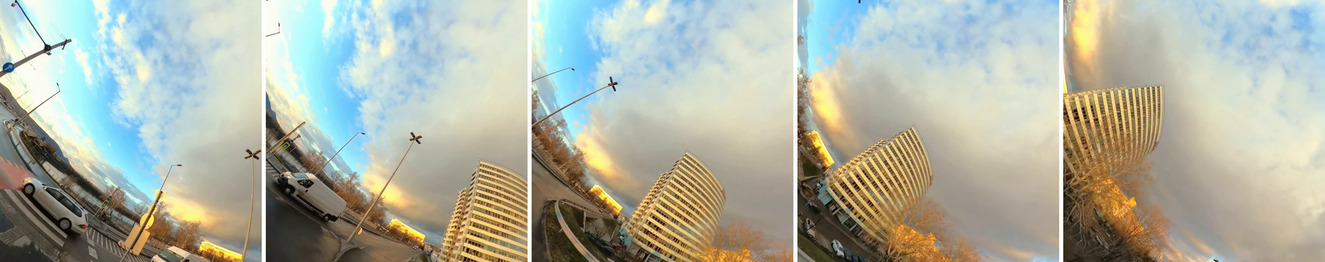} \\
        \includegraphics[width=1.0\linewidth]{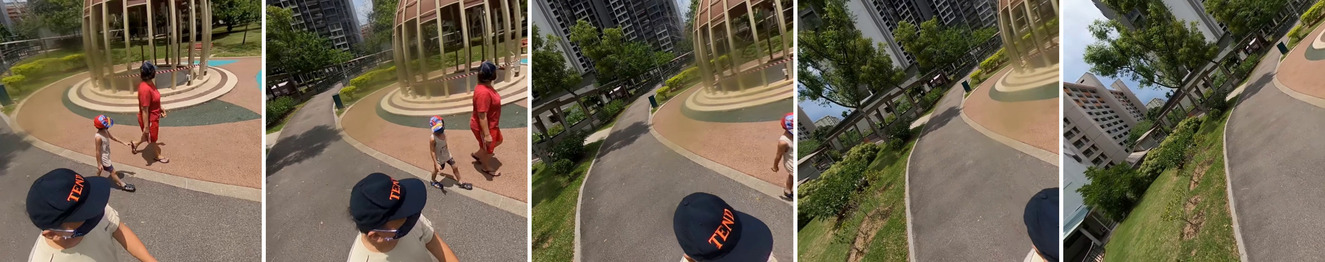} \\
        \includegraphics[width=1.0\linewidth]{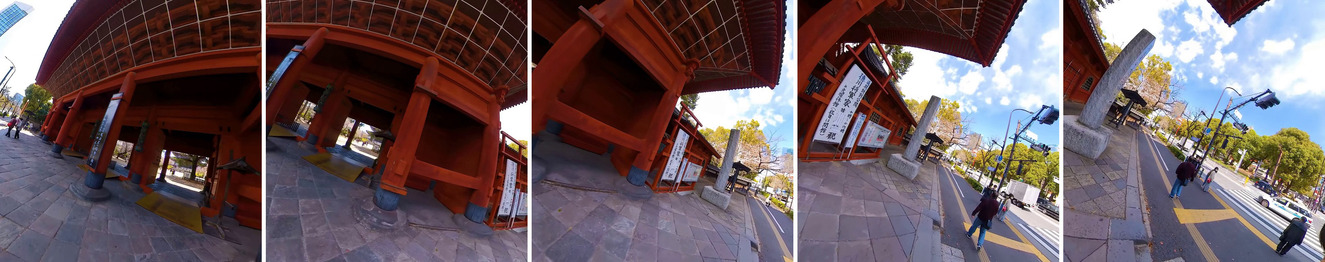} \\
        \includegraphics[width=1.0\linewidth]{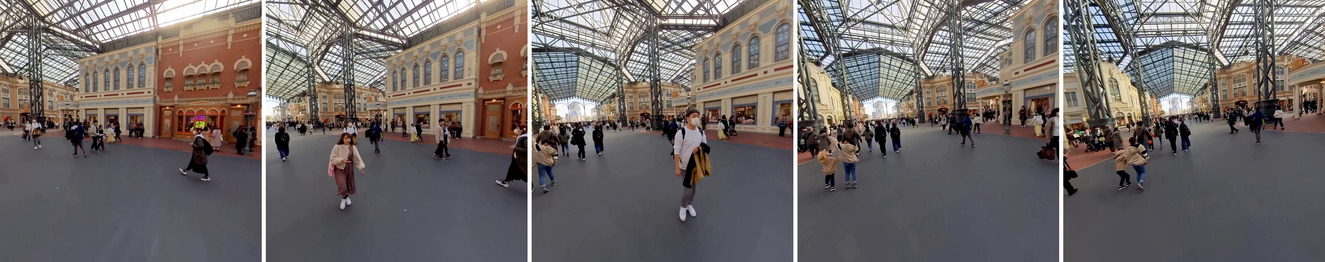} \\
        \includegraphics[width=1.0\linewidth]{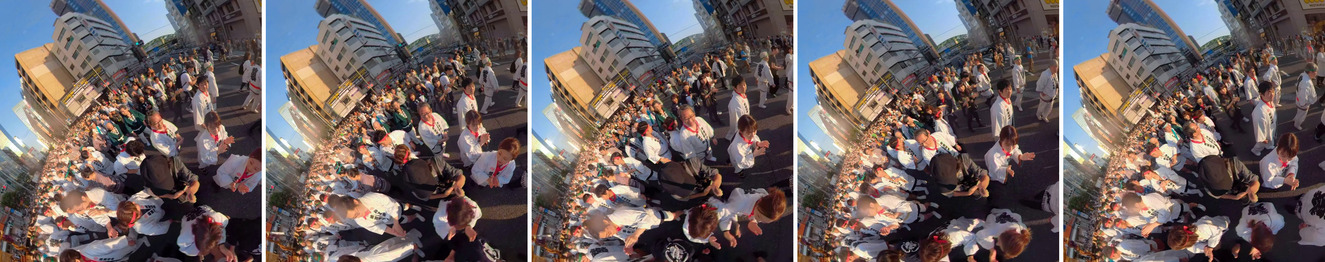} \\
        \includegraphics[width=1.0\linewidth]{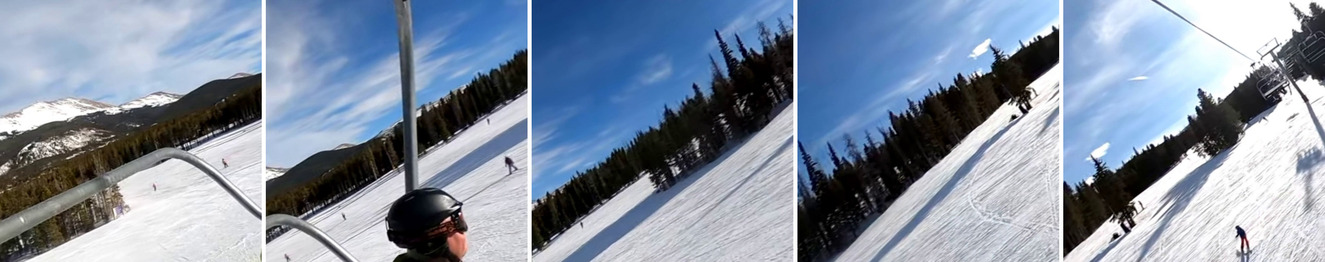}
    \end{tabular}
    \caption{\textbf{Synthesized Projected Video Frame Sequences (Part 1).} We visualize a set of projected video frame sequences synthesized from our gravity-aligned panoramic source. These sequences showcase the diversity of synthesized camera trajectories and lens effects in our dataset.}\label{fig:supp_panshot_samples_1}
\end{figure}

\begin{figure}[ht!]
    \centering
    \begin{tabular}{c}
        \includegraphics[width=1.0\linewidth]{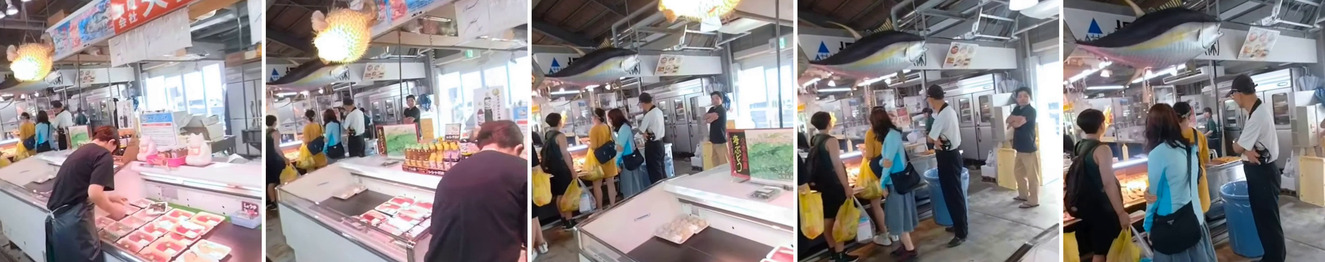} \\
        \includegraphics[width=1.0\linewidth]{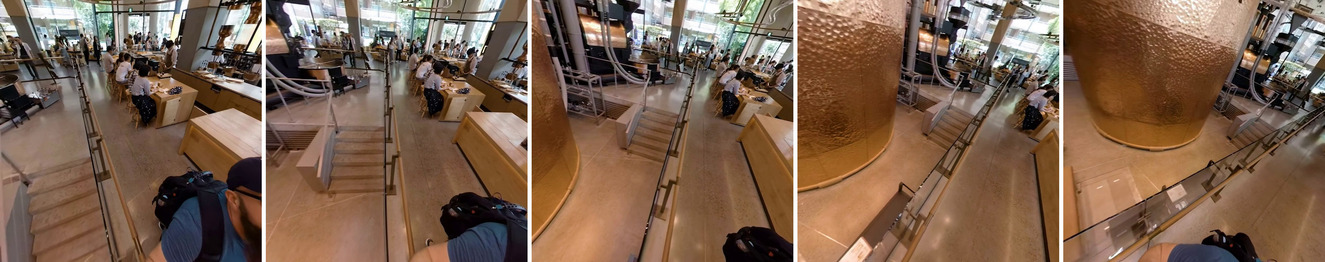} \\
        \includegraphics[width=1.0\linewidth]{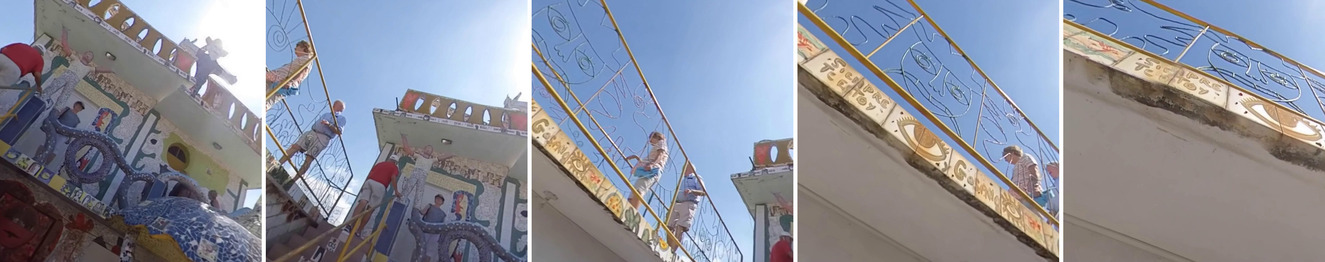} \\
        \includegraphics[width=1.0\linewidth]{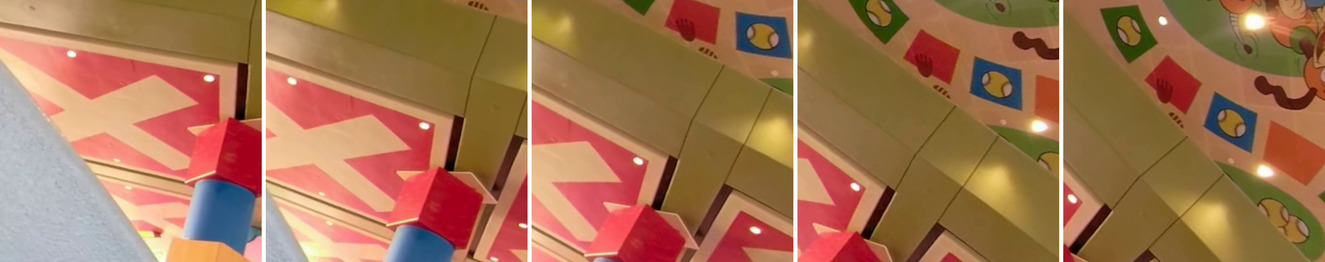} \\
        \includegraphics[width=1.0\linewidth]{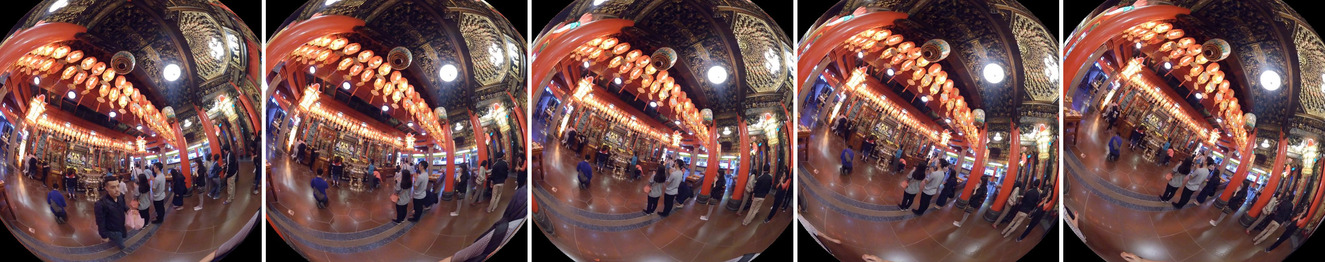} \\
        \includegraphics[width=1.0\linewidth]{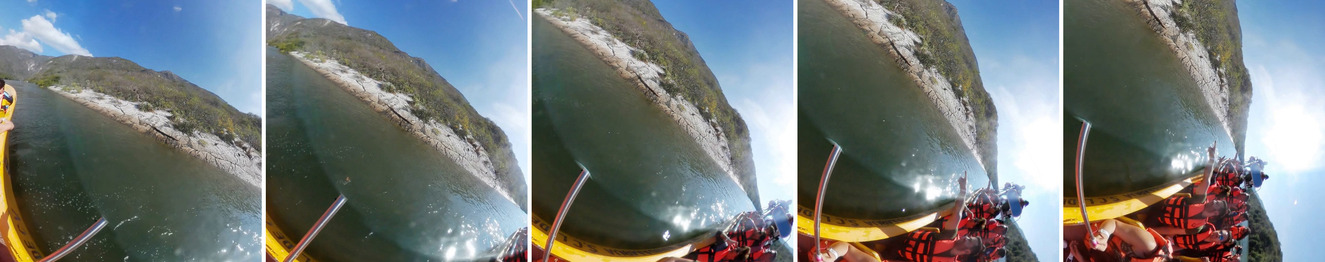}
    \end{tabular}
    \caption{\textbf{Synthesized Projected Video Frame Sequences (Part 2).} Additional representative sequences generated by re-projecting panoramic content onto augmented virtual camera trajectories, demonstrating diverse illumination and complex geometric structures.}\label{fig:supp_panshot_samples_2}
\end{figure}

In \suppref{fig:supp_panshot_samples_1,fig:supp_panshot_samples_2}, we showcase several representative projected video frame sequences synthesized from panoramic videos. These samples demonstrate the diversity of real-world scenarios captured, ranging from indoor environments to various outdoor urban and natural settings. By leveraging gravity-aligned panoramas, our pipeline effectively generates realistic training sequences with diverse intrinsics and distortion patterns by re-projecting the panoramic content onto augmented virtual camera trajectories.

\section{Limitations and Future Work}
\label{sec:supp_limitations}

This section expands the limitations stated in \cref{sec:conclusion} of the main paper.

\textbf{Camera assumptions.} We fix the principal point at the image centre and treat the intrinsics as shared across the sequence. Both assumptions hold for the vast majority of consumer captures, but neither survives an optical zoom during the clip, an asymmetric crop, or a lens whose optical centre is deliberately offset. Relaxing them is a question of parameterisation rather than of architecture. The perspective field constrains the principal point only weakly, so recovering it would be better served by a representation that predicts per-pixel ray directions, in the spirit of AnyCalib~\citep{tirado2025anycalib}; a zoom or an asymmetric crop, in turn, calls for intrinsics estimated independently for each view rather than a single vector shared across the sequence.

\textbf{Sparse-view operating point.} Our gains concentrate where views are few. We train on at most $24$ views and the global cross-frame attention grows quadratically in $N$, so long, densely sampled sequences fall outside our operating point, and that is precisely the regime in which reconstruction-based pipelines keep refining intrinsics through bundle adjustment over hundreds of frames. The two are therefore complementary rather than competing, and the natural direction is to use our estimate as the initialisation of such a pipeline exactly where reconstruction is fragile.
\textbf{Efficiency.} On a single image, the network forward pass takes $70.2 \pm 11.3$\,ms, against $22.3 \pm 9.9$\,ms for GeoCalib~\citep{veicht2024geocalib} and $19.8 \pm 3.3$\,ms for AnyCalib~\citep{tirado2025anycalib}, measured on $1080 \times 1080$ images of MegaDepth-2K with a single NVIDIA L40S and averaged over $99$ images. The gap follows from the backbone: ours is designed to accept a variable number of views and to reason across them, whereas both baselines are single-view models, and AnyCalib returns intrinsics only, so its timing covers strictly less work. Bringing the cost down is largely orthogonal to the calibration itself: distilling the aggregator into a smaller backbone and pruning the tokens that carry no calibration signal are both promising directions.

\end{document}